\documentclass{article}

    \usepackage[final]{neurips_2022}

\RequirePackage{algorithm}
\usepackage{algpseudocode}

\usepackage[utf8]{inputenc} 
\usepackage[T1]{fontenc}    
\usepackage{hyperref}       
\usepackage{url}            
\usepackage{booktabs}       
\usepackage{amsfonts}       
\usepackage{nicefrac}       
\usepackage{microtype}      
\usepackage[table, dvipsnames]{xcolor} 

\usepackage[normalem]{ulem}
\usepackage{caption}
\usepackage{subcaption}

\usepackage{mathtools}
\usepackage{amsthm}
\usepackage{enumitem}
\usepackage{array,multirow}
\newcommand{\STAB}[1]{\begin{tabular}{@{}c@{}}#1\end{tabular}}

\usepackage[capitalize,noabbrev]{cleveref}

\usepackage{wrapfig}

\usepackage{enumitem} 
\setlist{noitemsep} 

\usepackage{pifont}
\usepackage{xpunctuate}
\providecommand{\eg}[0]{\xperiodafter{e.g}}

\providecommand{\ie}[0]{\xperiodafter{i.e}}

\DeclareMathAlphabet{\nicecal}{OMS}{zplm}{m}{n}

\DeclareMathOperator*{\argmin}{arg\,min}

\theoremstyle{plain}

\theoremstyle{definition}

\theoremstyle{remark}

\renewcommand{\cite}[1]{\citep{#1}}

\newcommand{\ignore}[1]{}

\title{On Enforcing Better Conditioned Meta-Learning \\ for Rapid Few-Shot Adaptation}

\author{%
    Markus Hiller\textsuperscript{1} \quad Mehrtash Harandi\textsuperscript{2} \quad Tom Drummond\textsuperscript{1}\\
    \vspace{-2.5mm}\\
    \textsuperscript{1}School of Computing and Information Systems, The University of Melbourne\\
    \textsuperscript{2}Department of Electrical and Computer Systems Engineering,  Monash University\\
    \texttt{markus.hiller@student.unimelb.edu.au} \\
    \texttt{mehrtash.harandi@monash.edu}\\
    \texttt{tom.drummond@unimelb.edu.au}
}

\begin{document}

\maketitle

\begin{abstract}
Inspired by the concept of preconditioning, we propose a novel method to increase adaptation speed for gradient-based meta-learning methods without incurring extra parameters. We demonstrate that recasting the optimisation problem to a non-linear least-squares formulation provides a principled way to actively enforce a \textit{well-conditioned} parameter space for meta-learning models based on the concepts of the condition number and local curvature. Our comprehensive evaluations show that the proposed method significantly outperforms its unconstrained counterpart especially during initial adaptation steps, while achieving comparable or better overall results on several few-shot classification tasks -- creating the possibility of dynamically choosing the number of adaptation steps at inference time.
\end{abstract}

\section{Introduction}
\label{sec:introduction}

Learning a new task from only a very limited number of datapoints in the target domain poses a central challenge in machine learning. 
Gradient-based meta-learning as introduced in MAML~\cite{finn2017_maml} has proved to be a versatile tool to successfully address such \textit{few-shot} problems, and has inspired a significant number of follow-up works tackling a diverse set of applications like regression~\cite{nichol2018_reptile,finn2018_probabilisticMAML}, image classification~\cite{zintgraf2019_cavia,antoniou2018_mamlpp,rajeswaran2019_implicitgrad}, reinforcement~\cite{al2018_reinffsl} and online continual learning~\cite{gupta2018_lamaml}.

Broadly speaking, such approaches define a family of tasks and attempt to solve a bi-level optimization problem. The outer loop aims to learn an effective \emph{meta initialization} of the parameters capturing properties that generalize well across different training tasks. This learned set of parameters is then \emph{adapted} to a novel and previously unseen task in \emph{only a few steps} during the inner loop.

\begin{figure}
\centering
\begin{subfigure}[b]{0.32\textwidth}
    \includegraphics[width=\textwidth]{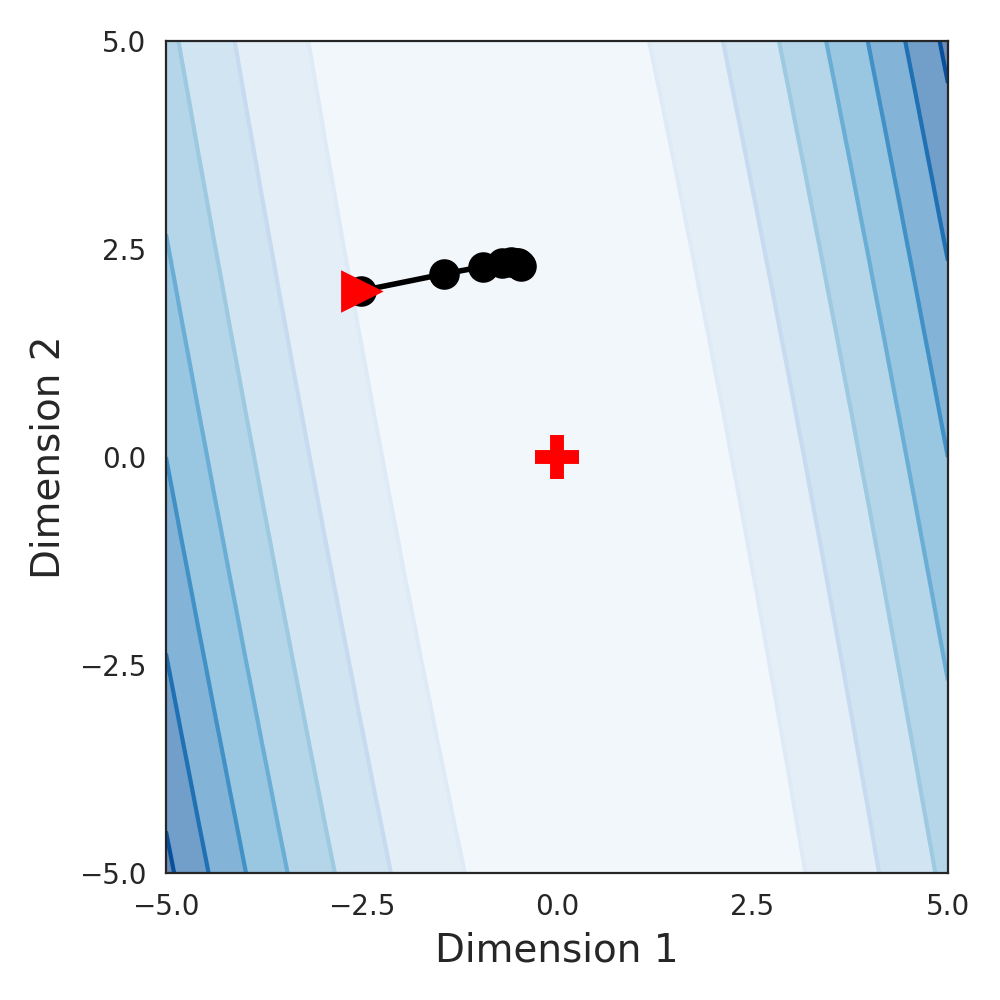}
    \caption{$\kappa_{\textbf{H}}\gg1$}
    \label{subfig:motivation_cond_high}
\end{subfigure}
\begin{subfigure}[b]{0.32\textwidth}
    \includegraphics[width=\textwidth]{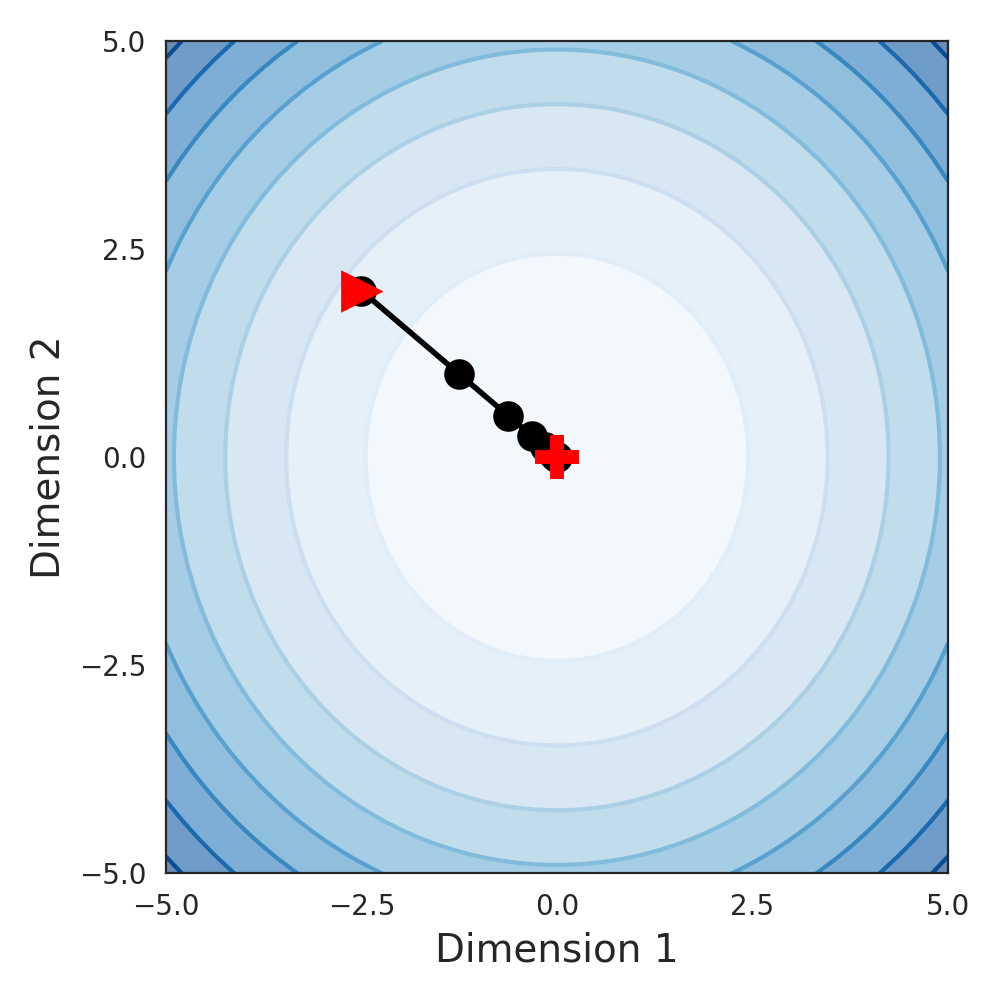}
    \caption{$\kappa_{\textbf{H}}\approx1$}
    \label{subfig:motivation_cond_ideal}
\end{subfigure}
\begin{subfigure}[b]{0.33\textwidth}
    \includegraphics[width=\textwidth]{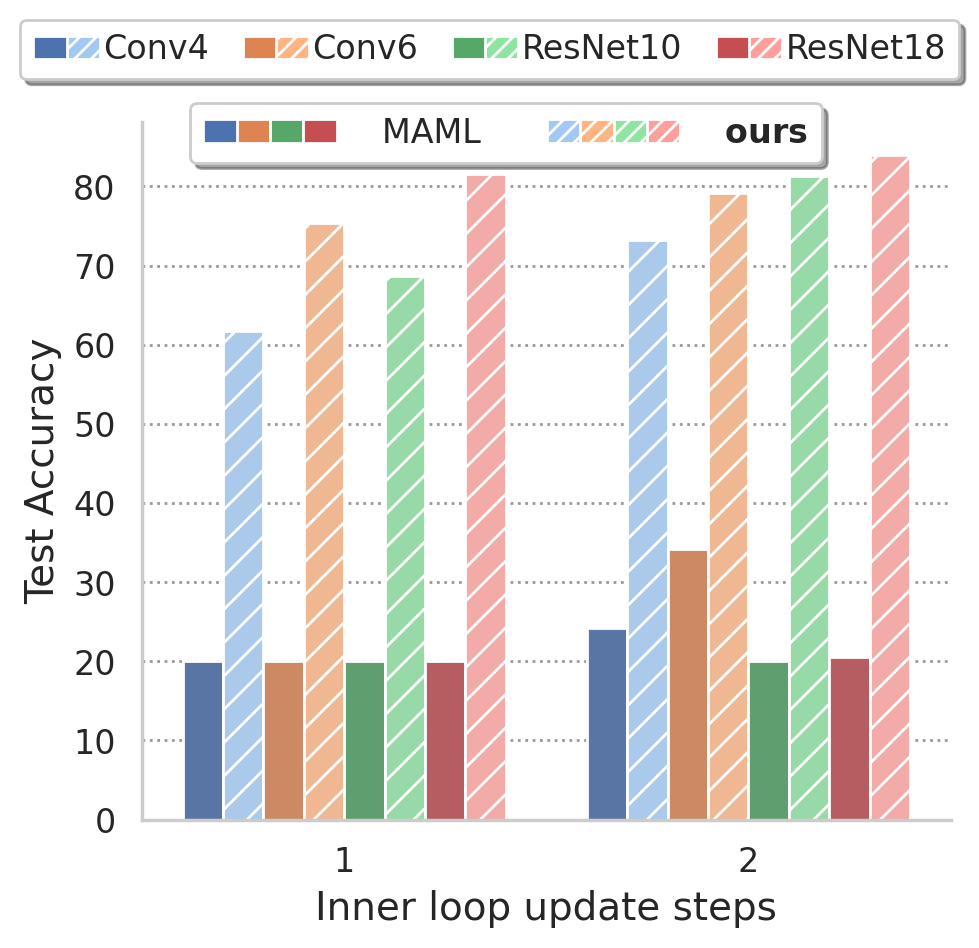}
    \caption{}
    \label{subfig:motivation_acc}
\end{subfigure}
\caption{\textbf{Importance of a well-conditioned parameter space.} (\subref{subfig:motivation_cond_high}) and (\subref{subfig:motivation_cond_ideal}) show the difference in convergence if optimization using SGD is performed in an ill-conditioned parameter space with high condition number~$\kappa_{\mathbf{H}}\gg1$ or well-conditioned one ($\kappa_{\mathbf{H}}\approx1$), respectively. Depicted is an example for two dimensions using $10$ update steps with learning rate $0.5$. (\subref{subfig:motivation_acc}) Average accuracies on the CUB test set for the baseline \textit{without} (MAML) and \textit{with} our proposed conditioning constraint. Displayed are the first two out of five inner-loop update steps for different network architectures.
}
\label{fig:motivation}
\end{figure}

Intuitively, one might expect that each step taken during the inner-loop adaptation \ignore{procedure}ought to be as efficient as possible, enabling the model to quickly converge towards the desired goal (\ie, solving the new task). Evaluating \ignore{currently existing}popular methods like MAML~\cite{finn2017_maml} \ignore{or CAVIA~\cite{zintgraf2019_cavia}}however shows that this is not \ignore{at all}the case (Figures \ref{fig:motivation} and \ref{fig:main_conditioning}). Particularly the initial steps fall short of this expectation. The reason for this behaviour is readily explained: \ignore{(i)}Gradient directions do generally not point towards the actual desired minimum (\cref{fig:motivation}) as a result of the often ill-conditioned parameter space in which the optimization is performed.
\ignore{ , and (ii) there simply exists no incentive for the methods to make each step as efficient as possible, since the only cost term influencing convergence is whether a desired metric is improved after a fixed number of steps. In other words, the imposed constraints are satisfied as long as some step(s) lead(s) to an overall reduction in cost.} 
\ignore{Think about: MAML++ does impose a loss after each step, but we don't compare against it -- Maybe drop (ii) since we don't actually tackle this directly! Alternative: Mention `\textit{while (ii) has been tackled \eg by \citet{antoniou2018_mamlpp},... -- But that might result in the question why there is no comparison!}}

This behaviour is undesirable for a number of reasons. Having high contribution towards reaching the optimum from the start opens up the possibility of dynamically choosing the number of steps to take at inference time, an interesting property for applications with limited or changing access to computational resources, and might further lead to a better overall convergence during the final steps. 

With this goal of rapid adaptation from the very first step in mind, we draw inspiration from the field of optimization -- specifically from the idea of \emph{preconditioning}. 
Instead of performing parameter updates directly, a suitable preconditioning matrix~$\mathbf{P}$ can be chosen and multiplied to the gradient~$\nabla_{\boldsymbol{\theta}}$,
implicitly transforming the optimization problem to a space where updates can be performed more efficiently and resulting in a modulated parameter update
\begin{equation}
	\boldsymbol{\theta}^{(k)} \:\:=\:\: \boldsymbol{\theta}^{(k-1)} - \alpha \, \mathbf{P} \, \nabla_{\boldsymbol{\theta}^{(k-1)}}\nicecal{L}\left(\nicecal{D},f_{\boldsymbol{\theta}^{(k-1)}}\right)\,.
	\label{eq:param_updates_precond}
\end{equation}
The reader might notice that this modified gradient descent update rule directly reduces to the well-known Newton's method if the preconditioning matrix is chosen to be the inverse of the Hessian matrix ($\mathbf{P}=\mathbf{H^{-1}}$ with $\mathbf{H}=\frac{\partial^2}{\partial\boldsymbol{\theta}\partial\boldsymbol{\theta}^\top}$) or an approximation thereof, and would thus enjoy a quadratic convergence rate via appropriate scaling of the gradients based on the local curvature.
The actual choice and computation of a preconditioner is however non-trivial for most optimization problems, especially in the context of deep learning. Variants developed in recent years~\cite{li2017_metasgd,gupta2018_shampoo,park2019_metacurvature,simon2020_gradientPrecon} can all be seen as proposing different ways of estimating suitable preconditioning matrices, often leading to a significant increase in parameters.\par
\noindent\textbf{Our method:}
In contrast to previous works, we propose a new way to achieve preconditioning without incurring any additional parameters. Reconsidering the definition of preconditioned optimization stated in \cref{eq:param_updates_precond}, we raise the following hypothesis:

\textit{If $f_{\boldsymbol{\theta}}$ is solely parameterized by $\boldsymbol{\theta}$ and $\boldsymbol{\theta}$ is learned, there exists a set of parameters~$\Tilde{\boldsymbol{\theta}}$ that achieves comparable performance on unseen tasks but possesses a low condition number~$\kappa$ which allows fast convergence with gradient descent even when choosing the preconditioning matrix to~$\Tilde{\mathbf{P}}=\mathbf{I}$.}

In other words, the preconditioning matrix~$\Tilde{\mathbf{P}}$ can be absorbed into this `new' learnable function~$f_{\Tilde{\boldsymbol{\theta}}}$, enabling the problem to be formulated and optimized directly in a well-conditioned space.

This raises the following questions: (i) Can the conditioning property of a model be actively influenced without negatively affecting its ability to converge and to solve a given task? (ii) If so, does this indeed lead to faster task-adaptation? \ignore{(iii) Does a better-conditioned parameter space provide advantages regarding generalization to other domains?}In this paper, we set out to provide insights and answers to these and additional questions. In particular, our contributions include the following:
\begin{enumerate}
    \item We show that the condition number of a network is a clear indicator for its few-step performance on an unseen task.
    \item We demonstrate a principled way to derive the condition number of a network by reformulating the optimization problem to a non-linear least-squares form over the space of available samples in the support set.
    \item We introduce a novel cost function based around the condition number to actively encourage learning of a well-conditioned parameter space for the model during training, allowing for rapid adaptation at inference time in very few steps.
    \item We conduct in-depth analyses regarding different network architectures, ability to continue adaptation beyond the training horizon, performance regarding an extended set of $N$-way $K$-shot few-shot classification scenarios, and demonstrate the efficacy of our method on all five popular few-shot classification benchmarks.
\end{enumerate}

\section{Preliminaries}
\label{sec:background}
One highly influential algorithm that tackles few-shot learning in a gradient-based fashion has been \emph{Model Agnostic Meta-Learning} (MAML) proposed by~\citet{finn2017_maml}. We start by introducing this algorithm to define the specific problem setting we are aiming to solve and to provide context for the following discussions. This is followed by a brief overview of the \emph{condition number} which constitutes one of the main concepts our method is based on.

\subsection{Problem setting and MAML}
Let $\nicecal{D}^{\mathrm{train}}_\tau$ and $\nicecal{D}^{\mathrm{val}}_\tau$ be the training and validation set of some given task~$\tau \sim p(\nicecal{T})$ (\eg image classification), respectively. Further, assume that~$\nicecal{D} \coloneqq \lbrace\boldsymbol{x}_i,y_i\rbrace_{i=1}^{\lvert \nicecal{D}\rvert}$, with $\boldsymbol{x}_i \in \nicecal{X}, y_i \in \nicecal{Y}$ for some small~$\lvert \nicecal{D}\rvert$. The predictor function of the model is denoted as~$f: \nicecal{X} \times \mathbb{R}^n \rightarrow \nicecal{Y}$ and is parameterized by~$\boldsymbol{\theta}\in \mathbb{R}^n$. MAML~\cite{finn2017_maml} then seeks to determine a universal meta initialization~$\boldsymbol{\theta}^{*}$ by solving
\begin{equation}
	\label{eq:outer_loop_maml}
	\underset{\boldsymbol{\theta}^*}{\argmin} \sum_{\tau \sim p(\nicecal{T})} \nicecal{L}\left(\nicecal{D}^{\mathrm{val}}_{\tau}, \;\; \boldsymbol{\theta}^{(K)}_\tau\!\left(\nicecal{D}^{\mathrm{train}}_{\tau}, \boldsymbol{\theta}^{*}\right)\right),
\end{equation}
where the inner-loop task-specific parameters are given by
\begin{equation}
	\label{eq:inner_loop_maml}
	\boldsymbol{\theta}^{(k)}_{\tau} = \boldsymbol{\theta}^{(k-1)}_{\tau} - \alpha \nabla_{\boldsymbol{\theta}^{(k-1)}}\nicecal{L}\left(\nicecal{D}^{\mathrm{train}}_{\tau},\boldsymbol{\theta}^{(k-1)}_{\tau}\right)  
\end{equation}
for $k\in [1,2,\dots,K]$ with $\boldsymbol{\theta}^{0}_{\tau}=\boldsymbol{\theta}^{*}$ and step size~$\alpha$.
In other words, given a task~$\tau$ MAML starts from the meta-initialisation~$\boldsymbol{\theta^{*}}$ and performs~$K$ gradient update steps (so-called \emph{inner-loop updates}) using the few available training samples of~$\nicecal{D}^{\mathrm{train}}_{\tau}$ for this specific task to obtain the adapted task-specific parameters~$\boldsymbol{\theta}^{(K)}_{\tau}$ (\cref{eq:inner_loop_maml}). It then uses the unseen samples of~$\nicecal{D}^{\mathrm{val}}_{\tau}$ of this task together with the adapted parameters~$\boldsymbol{\theta}^{(K)}_{\tau}$ to improve the meta-initialisation~$\boldsymbol{\theta^{*}}$ in the so-called \emph{outer-loop update} (\cref{eq:outer_loop_maml}). Note that this is possible since~$\boldsymbol{\theta}^{(K)}_{\tau}$ is dependent on its initialisation~$\boldsymbol{\theta}^{0}_{\tau}=\boldsymbol{\theta}^{*}$ as defined in \cref{eq:inner_loop_maml}, and gradients can thus be computed throughout all~$K$ inner-loop updates.

\subsection{Condition number of a matrix}
\label{sec:cond_nmbr_prelim}
One way to quantify local curvature of a parameter space and represent the expected convergence behaviour using a single scalar value is the \emph{condition number}~$\kappa_{\mathbf{H}}$ of the Hessian~$\mathbf{H}_f$ of a function~$f$ that is to be optimized. 
For a square-summable sequence space~$\ell^2$~(\ie, Euclidean), the condition number of~$\mathbf{H}_f$ can be computed via its maximum and minimum singular values~$\sigma\left(\mathbf{H}_f\right)$, or its eigenvalues~$\lambda\left(\mathbf{H}_f\right)$ in case~$\mathbf{H}_f$ is normal, as
\begin{equation}
	\label{eq:condition_nmbr}
	\kappa_{\mathbf{H}} = \frac{\left|\sigma_{\max}(\mathbf{H}_f)\right|}{\left| \sigma_{\min}(\mathbf{H}_f)\right|} \quad \quad\mathrm{or} \quad \quad
	\kappa_{\mathbf{H}} = \frac{\left|\lambda_{\max}(\mathbf{H}_f)\right|}{\left| \lambda_{\min}(\mathbf{H}_f)\right|} \;\;\;\; \mathrm{if} \;\;\;\; \mathbf{H}^*_f\mathbf{H}_f=\mathbf{H}_f\mathbf{H}^{*}_f.
\end{equation}

The high condition number associated with the optimization problem depicted in \cref{fig:motivation}\,(\subref{subfig:motivation_cond_high}) represents a higher difference in local curvature across the parameter space, leading to \ignore{relatively }slow convergence caused by the initial gradients not pointing towards the minimum. In contrast, the better conditioned problem shown in~(\subref{subfig:motivation_cond_ideal}) with a significantly lower condition number~$\kappa_{\mathbf{H}}\approx1$ approaches the desired minimum on a direct way in very few steps.
\section{Learning a better-conditioned parameter space}
\label{sec:method}
We start this section by introducing how the objective used in few-shot learning can be reformulated as a linear least-squares problem, providing the basis to approximate its Hessian. We then introduce our proposed method to enforce a well conditioned parameter space and discuss a variation with constraining only a parameter subset that allows better scalability to models with higher parameter counts. 
\subsection{Reformulated problem setting}
We reconsider the minimization objective of one inner-loop step~$k$ used to perform the task-specific adaptation. The average loss over the support set samples in the inner loop can be reformulated to describe a non-linear least squares problem
\begin{equation}
	\label{eq:net_loss_nls}
	\nicecal{L}\big(\nicecal{D}_{\tau}^{\mathrm{train}},\boldsymbol{\theta}^{(k)}_{\tau}\big) \:\: = \:\:  \sum_{i=1}^{\lvert\nicecal{D}_{\tau}^{\mathrm{train}}\rvert}\sqrt{\frac{1}{\lvert\nicecal{D}_{\tau}^{\mathrm{train}}\rvert}\ell\left(\boldsymbol{x}_i,y_i,\boldsymbol{\theta}^{(k)}_{\tau}\right)}^2
	\:\: = \:\: \sum_{i=1}^{\lvert\nicecal{D}_{\tau}^{\mathrm{train}}\rvert} r_{i}\big(\boldsymbol{\theta}^{(k)}_{\tau}\big)^2,
\end{equation}
with~$ r_{i}\big(\boldsymbol{\theta}^{(k)}_{\tau}\big)$ representing the residual of each individual sample~$i$ and $\lvert\nicecal{D}_{\tau}^{\mathrm{train}}\rvert$ the number of available task-specific samples.

As introduced in the previous section, the Hessian of a function describes its local curvature and can be used to gain insights into the convergence behaviour if the function is to be minimized via some gradient-based optimisation method (\cref{fig:motivation}). Note that the dependence of gradient, Jacobian and Hessian on the dataset~$\nicecal{D}_{\tau}^{\mathrm{train}}$ and parameters~$\boldsymbol{\theta}^{(k)}_{\tau}$ is being partially dropped from the notation for improved readability during the following derivation. The dependence of the residual~$r_i\big(\boldsymbol{\theta}^{(k)}_{\tau}\big)$ on the parameters~$\boldsymbol{\theta}^{(k)}_{\tau}$ is additionally abbreviated via its simplified version~$r_i(\boldsymbol{\theta})$, and $\nicecal{D}_{\tau}^{\mathrm{train}}$ is referred to as $\nicecal{D}$.
Using the residual-based notation of \cref{eq:net_loss_nls} together with the introduced simplifications regarding notation, the gradient of~$\nicecal{L}\big(\nicecal{D}_{\tau}^{\mathrm{train}},\,\boldsymbol{\theta}^{(k)}_{\tau}\big)$ with respect to some parameter~$\theta_j$ is given by
\begin{equation}
	g_j = 2\sum_{i=1}^{\lvert\nicecal{D}\rvert} r_{i}\big(\boldsymbol{\theta}\big) \frac{\partial r_{i}\big(\boldsymbol{\theta}\big)}{\partial \theta_j},
\end{equation}
and the elements of its Hessian, computed by differentiating the gradient elements~$g_j$ with respect to parameter~$\theta_m$, are given by
\begin{equation}
	H_{jm} = 2\sum_{i=1}^{\lvert\nicecal{D}\rvert} \left(
	\frac{\partial r_{i}\big(\boldsymbol{\theta}\big)}{\partial \theta_j}
	\frac{\partial r_{i}\big(\boldsymbol{\theta}\big)}{\partial \theta_m} + 
	r_{i}\big(\boldsymbol{\theta}\big) \frac{\partial^2 r_{i}\big(\boldsymbol{\theta}\big)}{\partial \theta_j\partial \theta_m}\right).
\end{equation}
Ignoring the second-order derivative terms of the Hessian leads to the well-known \emph{Gauss-Newton method} in which the Hessian is approximated via the use of the Jacobian in the form of
\begin{equation}
	H_{jm} \approx 2\sum_{i=1}^{\lvert\nicecal{D}\rvert} \left(
	\frac{\partial r_{i}\big(\boldsymbol{\theta}\big)}{\partial \theta_j}
	\frac{\partial r_{i}\big(\boldsymbol{\theta}\big)}{\partial \theta_m}\right) = 
	2\sum_{i=1}^{\lvert\nicecal{D}\rvert} J_{i\!j}J_{i\!m} \qquad \mathrm{or} \qquad \mathbf{H}\approx 2\,\mathbf{J}^{\top}\mathbf{J}.
\end{equation}

It is to be noted that we did not modify any component of the original problem formulation at any point. Thus, the Jacobian product approximating the Hessian of our objective function directly reflects the convergence behaviour of the optimisation problem. More specifically, computing the condition number~$\kappa_{\mathbf{H}}$ (\cref{sec:cond_nmbr_prelim}) provides a scalar measure how well- or ill-posed the problem is.

\subsection{Conditioning constraint and overall objective}
Computing the condition number by using the maximum and minimum eigenvalues as defined in \cref{eq:condition_nmbr} ignores the distribution of the remaining~$\lvert\nicecal{D}^{\mathrm{train}}_{\tau}\rvert-2$ eigenvalues \ignore{. Since we are aiming to actively enforce learning a well-conditioned parameter space during training, using the actual definition of the condition number in this way}and would unnecessarily weaken the training signal. Instead we propose to use a measure considering the distribution of all eigenvalues~$\boldsymbol{\lambda}(\mathbf{H})$ based on the observation that a well-conditioned problem with~$\kappa_\mathbf{H}\approx 1$ can only be achieved if all eigenvalues are almost identical.
We model this requirement by penalizing high variance of the approximated Hessian's logarithmic eigenvalues \ignore{with basis~$10$ }and define this \textit{conditioning constraint} as a new loss function over the $K$ inner-loop updates as
\begin{equation}
    \nicecal{L}_{\kappa}\left(\boldsymbol{\theta}^{(K)}_\tau\!\left(\nicecal{D}^{\mathrm{train}}_{\tau}, \boldsymbol{\theta}^{*}\right)\right) =  \frac{1}{K} \sum_{k=1}^{K}\mathrm{Var}\left(\log_{10}\left(\boldsymbol{\lambda}\left(\mathbf{J}^{(k)}\mathbf{J}^{(k)^{\top}}\right)\right)\right).
\label{eq:condition_loss}
\end{equation}
This formulation contains two design choices. The first choice to use logarithmic eigenvalues is founded in the fact that the actual magnitude is irrelevant for the convergence property since only its relative magnitude compared to all other eigenvalues is of importance. Using the logarithmic form to compute the variance provides a scale-invariant way of penalising deviation and thus aligns with the ratio-based definition of the condition number, since it returns the same loss penalty for $\lambda_1=0.1,\lambda_2=0.2$ as for $\lambda_1=10,\lambda_2=20$ -- both of which would have an identical condition number~$\kappa=2$ and thus be considered to have similar convergence properties. 
The second choice of equally weighting all $K$ inner-loop update steps is based on simplicity,
and future research might explore whether non-equal weighting is beneficial.

Due to the number of samples being significantly smaller than the number of parameters and the resulting rank-deficiency of~$\mathbf{J}^{\top}\mathbf{J}$, only a subset of size~$\lvert\nicecal{D}^{\mathrm{train}}_{\tau}\rvert$ of the total parameter space is directly influenced by the condition loss~$\nicecal{L}_{\kappa}$ at each task iteration, whereas the remainder \ignore{of the parameter space }is its null space. This allows us to instead compute the smaller product~$\mathbf{J}\mathbf{J}^{\top}$ with equivalent non-zero eigenvalues. It is to be noted that the gradients used for this computation will however backpropagate to all parameters involved in forming~$\mathbf{J}$, thus appropriately constraining all relevant parameters.

The overall objective for the outer loop of the meta-learning process with weighting factor~$\gamma$
\begin{equation}
	\label{eq:cond_new_obj}
	\underset{\boldsymbol{\theta}^*}{\argmin} \sum_{\tau \sim p(\nicecal{T})} \nicecal{L}\left(\nicecal{D}^{\mathrm{val}}_{\tau}, \;\; \boldsymbol{\theta}^{(K)}_\tau\!\left(\nicecal{D}^{\mathrm{train}}_{\tau}, \boldsymbol{\theta}^{*}\right)\right) \:\:  + \:\: \gamma \nicecal{L}_{\kappa}\left(\boldsymbol{\theta}^{(K)}_\tau\!\left(\nicecal{D}^{\mathrm{train}}_{\tau}, \boldsymbol{\theta}^{*}\right)\right)
\end{equation}
encourages the network to learn a set of initialization parameters~$\boldsymbol{\theta}^{*}$ that not only generalizes well across tasks ($\nicecal{L}$) but also possesses a well-posed problem structure with low condition number in the inner loop that allows rapid adaptation in very few steps ($\nicecal{L}_{\kappa}$). This is possible since the condition number representing the convergence quality can computed via the approximated Hessian which in turn is dependent on the network's parameters, allowing the computation of gradients all the way back to the initialization~$\boldsymbol{\theta}^{*}$.

\subsection{Constraining a parameter subset}
To reduce complexity, improve scalability and considering the fact that the network might face some trade-off between accuracy and well-conditioned problem formulation induced by the objective in \cref{eq:cond_new_obj}, we build upon the findings of~\citet{raghu2019_rapidanil} that mainly the last few layers undergo significant adaptation during inner-loop updates. We thus propose and investigate constraining only a specific subset of the network's overall parameters through our proposed conditioning loss, namely the indicated `last few layers'. This significantly reduces computational requirements and makes our approach more scalable to deeper networks.
The only modification required is a slight adaptation of \cref{eq:cond_new_obj} by replacing $\boldsymbol{\theta}^{(K)}_\tau$ in $\nicecal{L}_{\kappa}$ with the new parameter subset~$\boldsymbol{\psi}^{(K)}_\tau \subseteq \boldsymbol{\theta}^{(K)}_\tau$.\par

\section{Experimental evaluation}
\label{sec:experiments}
In our experimental evaluation, we aim to gain empirical insights to answer the following questions: (i) Can the conditioning property of a model be actively influenced without negatively affecting its ability to converge and to solve a task? (ii) If so, does this indeed lead to faster task-adaptation? (iii) How does it impact networks of various depths and trained on different datasets? (iv) Does a better-conditioned parameter space provide advantages across different many-way multi-shot settings? (v) Can we expect continued adaptation beyond the training horizon?
\begin{table}[t]
\caption{\textbf{Validating baseline re-implementation.} Results of our re-implementation 
vs. the original work of~\citet{finn2017_maml}.
Means and 95\% confidence intervals of 600 randomly generated test episodes for 5-way 1-shot and 5-way 5-shot classification on the \textit{mini}ImageNet using a Conv4 model.}
\label{tab:baseline_reimpl}
\vskip -0.2in
    \begin{center}
    \scalebox{0.89}
    {
    \setlength{\tabcolsep}{3pt}
    \setlength{\tabcolsep}{6pt}
    \begin{tabular}{@{}l *4c @{}} 
    \specialrule{.2em}{.1em}{.1em}
     & \multicolumn{2}{c}{\textbf{5-shot}} &  \multicolumn{2}{c}{\textbf{1-shot}}\\
          \textbf{Method} & \textbf{Reported}  & \textbf{Ours} & \textbf{Reported}  & \textbf{Ours} \\    \midrule 
           MAML~\cite{finn2017_maml} & $63.11{\scriptstyle \pm 0.92}$ & $64.50{\scriptstyle \pm 0.69}$ & $48.70{\scriptstyle \pm 1.84}$ & $48.15{\scriptstyle \pm 0.80}$ 
          \\
        \bottomrule
    \end{tabular}
    }
    \end{center}
    \vskip -0.2in
\end{table}
\subsection{Experimental setup} \label{sec:exp_setup}
We choose MAML~\cite{finn2017_maml} as our baseline for comparison due to its vast popularity and the influence regarding follow-up works it inspired since its introduction (see~\cref{sec:relatedwork}). To validate the implementation of our MAML baseline, we report the results achieved by the original and our re-implemented baseline in \cref{tab:baseline_reimpl}, showing results are on-par with the original version.

We evaluate our method on all five popular few-shot classification datasets, namely \textit{mini}ImageNet~\cite{vinyals2016_matchingnet,ravi2017_optfsl}, \textit{tiered}ImageNet~\cite{ren2018_metasemisup}, CIFAR-FS~\cite{bertinetto2019_cifarfsl}, FC100~\cite{oreshkin2018_tadam} and CUB-200-2011~\cite{WahCUB_200_2011}, hereafter referred to as `CUB'. We follow previous work like~\citet{chen2019closerfewshot} and report the averaged test accuracies of 600 randomly sampled experiments using the $k$ samples of the respective $k$-shot setting as support and 16 query samples for each class. Our baselines and methods are trained on a single NVIDIA RTX 3090 for $60{,}000$ episodes for 1-shot and $40{,}000$ for 5-shot tasks, with the best model picked based on highest validation accuracy.  We use 5 inner-loop updates and $\gamma=1$.

\begin{figure}[t]
\centering
\begin{subfigure}[b]{0.32\textwidth}
    \includegraphics[width=\textwidth]{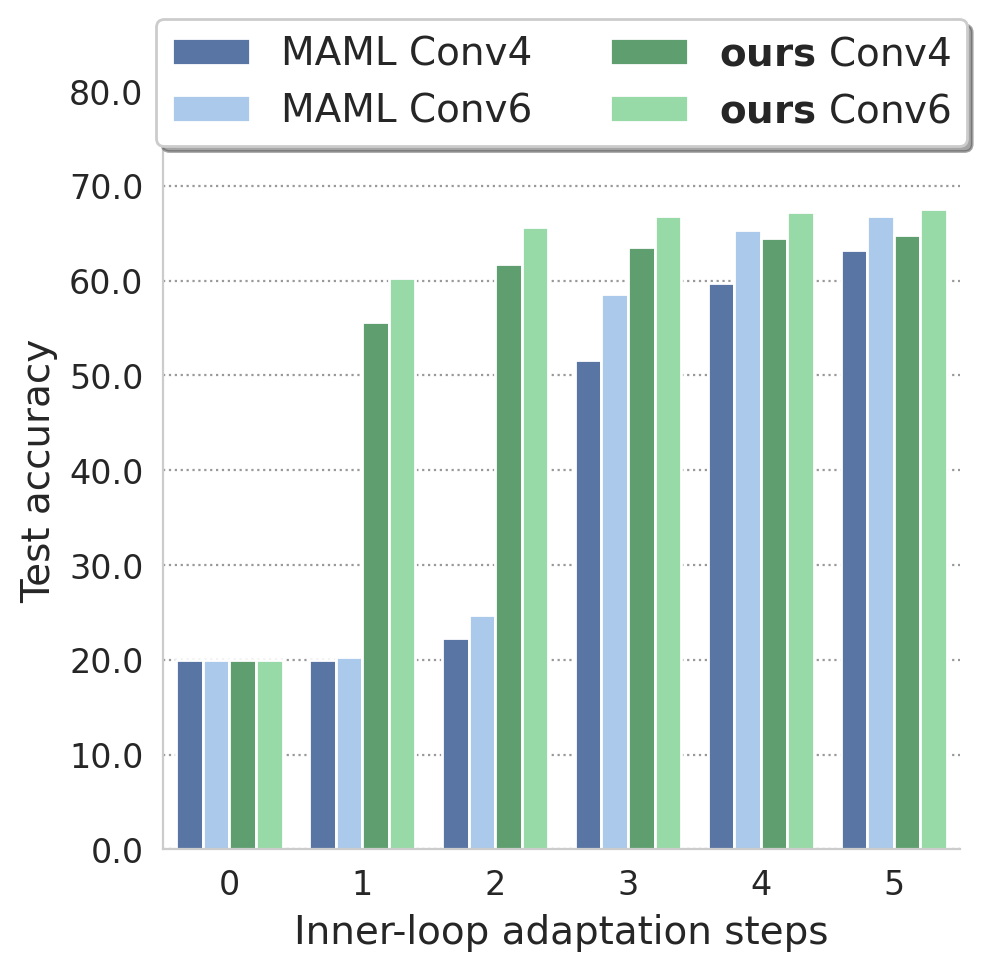}
    \caption{}
    \label{subfig:main_test_acc}
\end{subfigure}
\begin{subfigure}[b]{0.32\textwidth}
    \includegraphics[width=\textwidth]{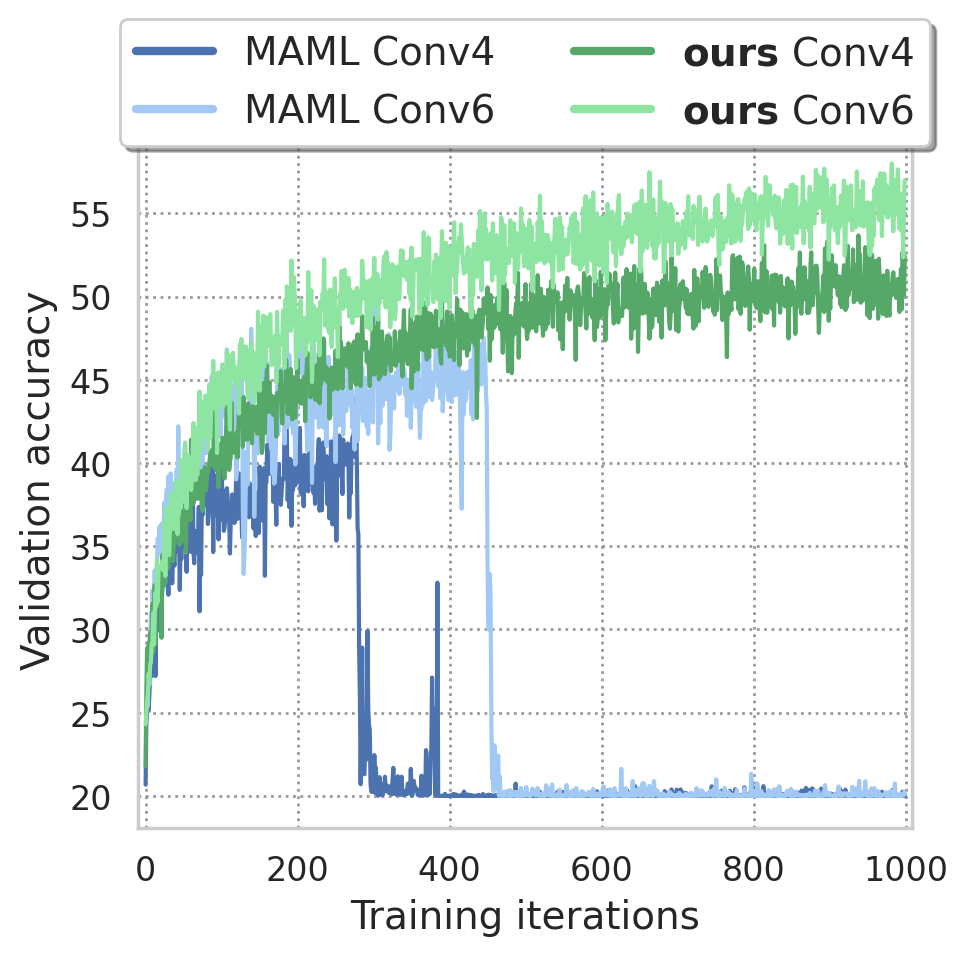}
    \caption{}
    \label{subfig:main_val_acc}
\end{subfigure}
\begin{subfigure}[b]{0.32\textwidth}
    \includegraphics[width=\textwidth]{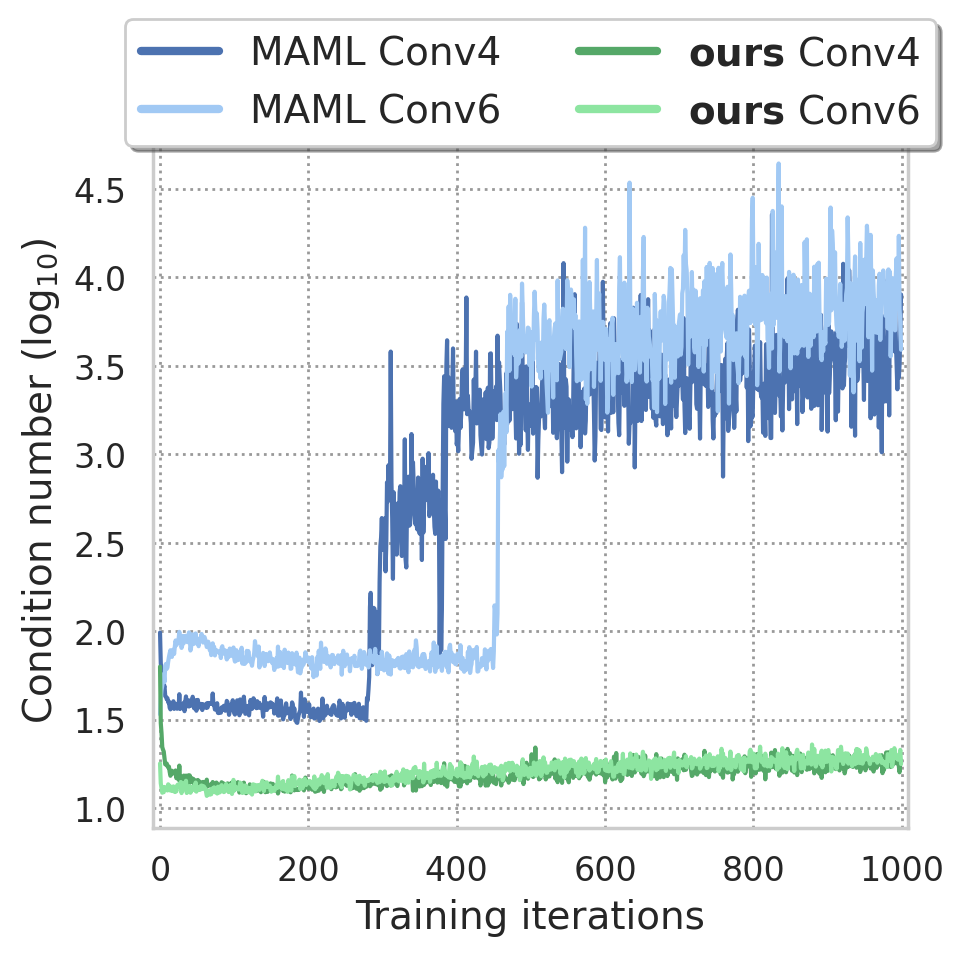}
    \caption{}
    \label{subfig:main_train_cond}
\end{subfigure}
\caption{\textbf{Enforcing well-conditioned adaptation.} Comparison of baseline method MAML~\cite{finn2017_maml} and the same method with our proposed conditioning constraint. Results were obtained on \textit{tiered}Imagenet for two different architectures. (\subref{subfig:main_test_acc}) Test accuracy over inner-loop adaptation steps. (\subref{subfig:main_val_acc}) Validation accuracy of inner-loop adaptation step 1 during training. (\subref{subfig:main_train_cond}) Condition number of parameters at adaptation step 1 during training, depicted in logarithmic scale.  }
\label{fig:main_conditioning}
\vskip -0.1in
\end{figure}
\subsection{Condition number indicates few-step performance}
As very first point, we provide insights into the basis this work is built on: The condition number of a network is closely related to its adaptation capabilities using very few steps. 
To this end, we take a closer look at the behaviour of a network's condition number during training and the associated performance in terms of validation accuracy after the first inner-loop update step (out of five), for methods trained \textit{without} or \textit{with} our proposed conditioning objective (\cref{eq:cond_new_obj}). 
The results depicted in \cref{fig:main_conditioning} demonstrate a clear dependence between the progress of the validation accuracy~(\subref{subfig:main_val_acc})\ignore{ achieved after the first inner-loop update step (out of five)} and its associated condition number~(\subref{subfig:main_train_cond}) in two ways: 1) Low condition number seems to generally align with high validation accuracy, and 2) sudden increases in condition number indicate instability regarding the validation accuracy. We further observe that the lower condition numbers of both models trained with the proposed constraint (`\textit{ours}') seem to improve convergence speed early on during training, leading to a slightly steeper increase in validation accuracy. \par

This dependence between condition number and validation accuracy exists across all inner-loop update steps, with corresponding discussion provided in the supplementary material of this paper. 

\subsection{Per-step adaptation with enforced conditioning}
Having demonstrated the existing dependence between the condition number and performance of a model, we now concentrate on our hypothesis raised in the introduction that actively constraining the network's parameter space to enforce better conditioning is possible without negatively affecting overall performance on unseen tasks. In \cref{fig:main_conditioning}\,(\subref{subfig:main_train_cond}) can be clearly observed that models trained \textit{with} $\mathcal{L}_\kappa$ (`\textit{ours}') exhibit significantly lower condition numbers than their unconstrained counterparts (`\textit{MAML}'), demonstrating that our proposed loss is indeed able to actively encourage learning of a better-conditioned parameter space. 

The performance of these models achieved on novel tasks of the test set is depicted across all five inner-loop update steps in \cref{fig:main_conditioning}\,(\subref{subfig:main_test_acc}) for a 5-way 5-shot scenario. While all networks seem to learn a similarly general meta initialization (step~$0$), the effect of using our proposed~$\mathcal{L}_{\kappa}$ is significant especially during the first steps, increasing the test accuracy by $35.68\%$ and $39.48\%$ (Conv4), and $39.98\%$ and $40.95\%$ (Conv6) for adaptation steps 1 and 2, respectively. While the unconstrained models mostly catch up during the following steps, their adaptation behaviour prevents any use of such trained models with fewer than the maximum number of adaptation steps used in training (here 5) -- in stark contrast to our models which achieve $86\%$ (Conv4) and  $89\%$ (Conv6) of their top accuracy already after one single step. 
We further observed that the faster initial convergence of our method seems to consistently lead to slightly higher overall performance across most experiments we performed, especially for deeper architectures (\cref{sec:deeper_beyond}) with $K\!\geq\!5$.

\subsection{Effect of constrained parameter subsets}
\label{sec:constrain_param_subset}
We now take a closer look at two distinct questions: (i) is the condition number of a parameter subset `representative' of the actual condition number of the network (based on all parameters) and if so, (ii) how do different subsets influence the networks adaptation performance? To this end, we train a Conv4 architecture and enforce conditioning on various parameter subsets of the last few layers. The development of the average condition numbers during training with respect to only the parameters of the classifier (`\textit{cls}') and to all parameters of the network (`\textit{all}') indicate that the condition number of this subset is indeed representative for the entire network (\cref{fig:par_sel}), an observation that applies to all evaluated subsets except the parameters of the embedding layer's batchnorm (`\textit{eBN}'). For further discussion, please see the supplementary material. \cref{tab:par_sel_tab} shows that all methods demonstrate significant improvements of test accuracy in the first few steps compared to the unconstrained baseline, and all except `\textit{eBN}' achieve superior overall performance after 5 steps. 
Despite these results indicating the slight advantage of selecting subset `\textit{emb}', we conduct the following experiments using the `\textit{cls}' subset for the reason that constraining the parameters of the classifier is independent of the backbone and allows for easier transition between different models.

\begin{figure}[t]
\centering
\begin{subfigure}{0.48\textwidth}
    \includegraphics[width=\textwidth]{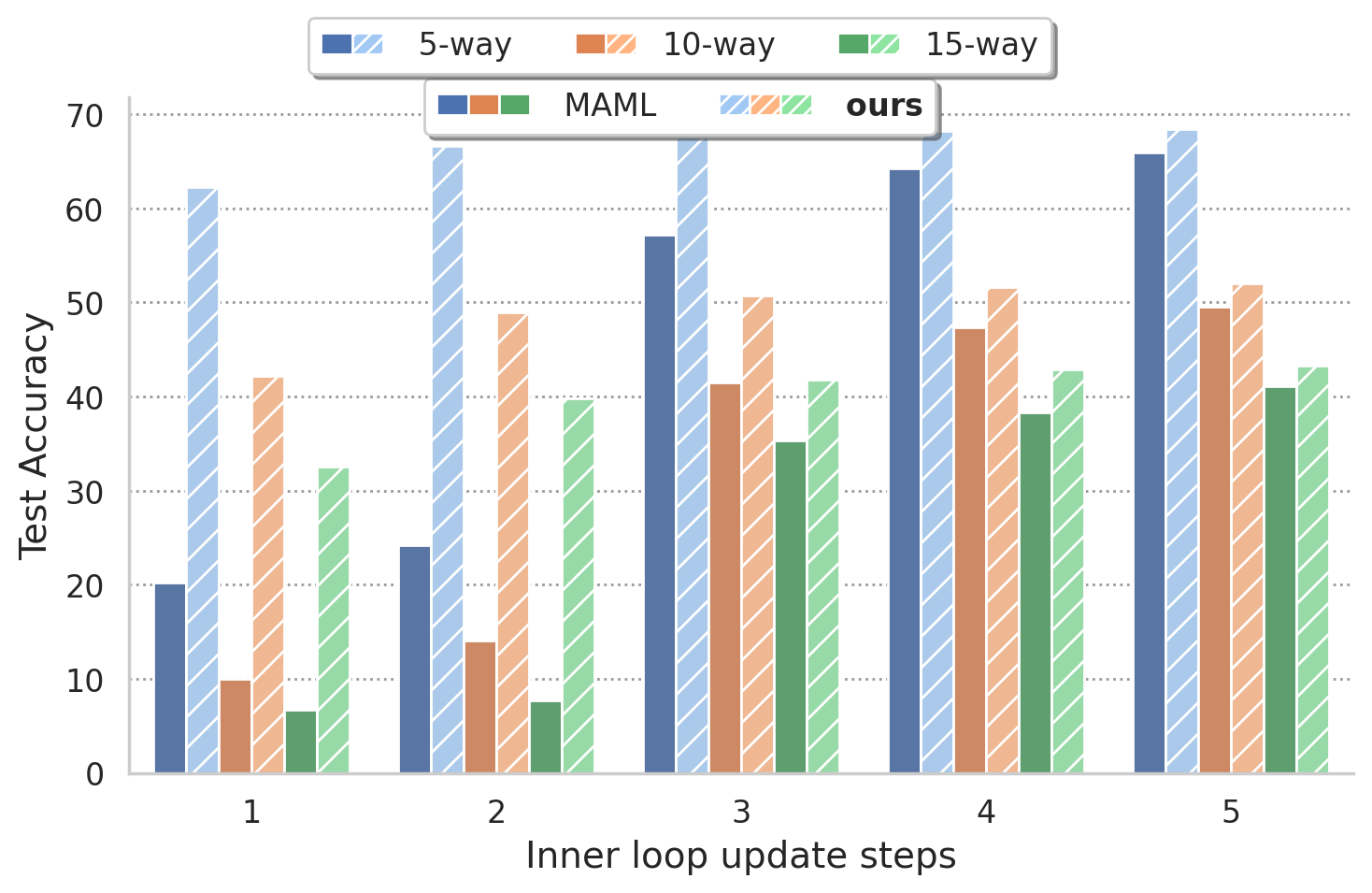}
    \caption{}
    \label{subfig:manyway}
\end{subfigure}
\begin{subfigure}{0.48\textwidth}
    \includegraphics[width=\textwidth]{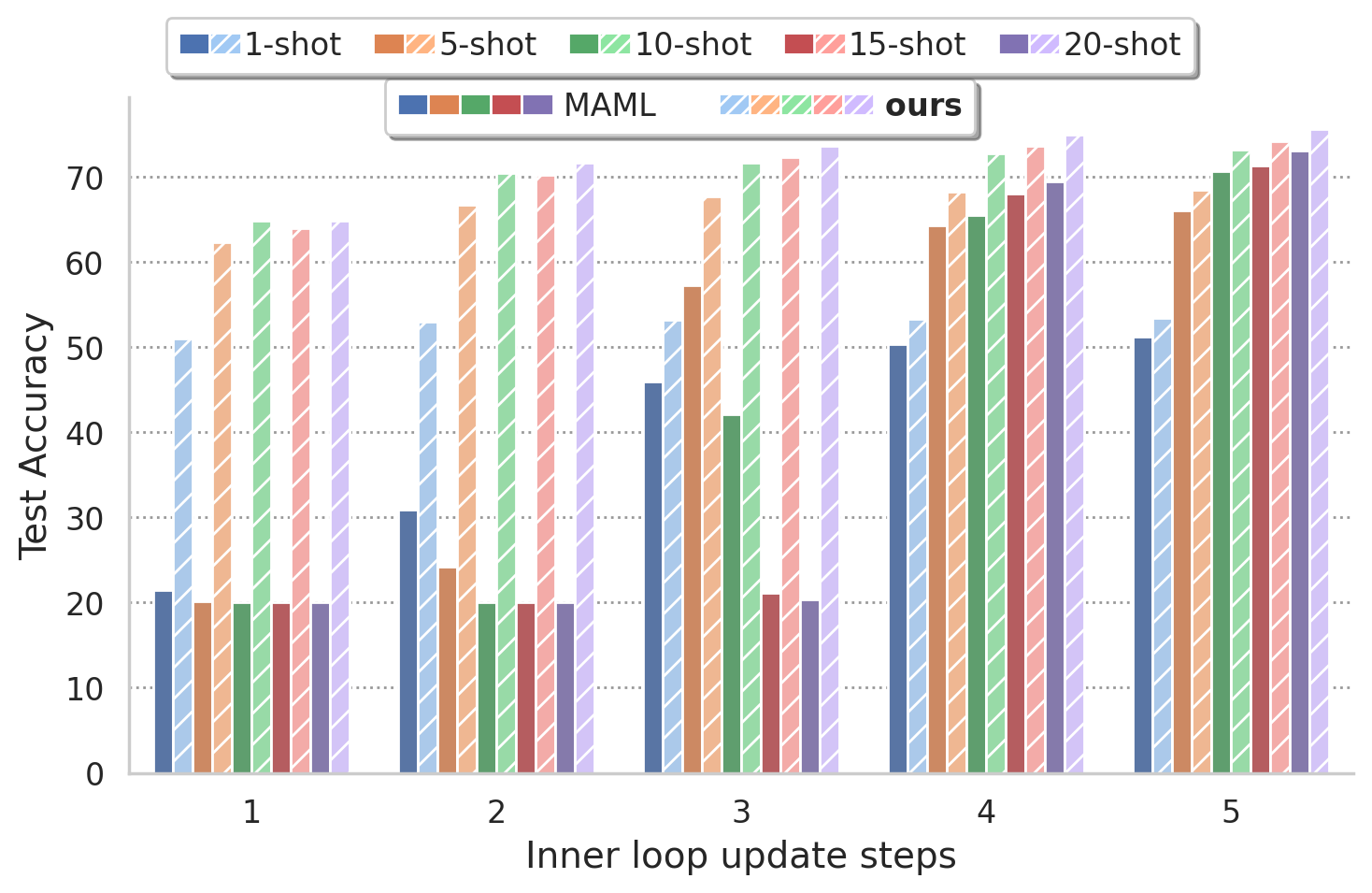}
    \caption{}
    \label{subfig:multishot}
\end{subfigure}
\caption{\textbf{Many-way multi-shot scenarios.} Reported are the average accuracy for the baseline \textit{without} conditioning (MAML) and the version \textit{with} our proposed conditioning loss on the \textit{mini}ImageNet testset. (\subref{subfig:manyway}) depicts the results for various $N$-way $5$-shot, (\subref{subfig:multishot}) for $5$-way $K$-shot scenarios.}
\label{fig:manyway_manyshot}
\vskip -0.18in
\end{figure}

\subsection{Increased network depths and adaptation beyond the training horizon}
\label{sec:deeper_beyond}
We conduct an extensive set of experiments with four different architectures (Conv4, Conv6, ResNet10, ResNet18) comparing the baseline and our method across all 5 datasets to evaluate the performance for changing network depths, both in $1$-shot and $5$-shot settings. The results in \cref{tab:depths_beyond} demonstrate the efficacy of the proposed constraint across all models (please refer to the supplementary material for complete results). We additionally observe that the better-conditioned parameter space seems to also encourage convergence to a better overall optimium especially for deeper architectures, \eg ResNet18 with $73.28\%$ vs. baseline $71.06\%$.  Interestingly, smaller unconstrained baseline models like Conv4 seem to naturally learn a better conditioned parameter space in 1-shot scenarios for most datasets, which might indicate some relationship between the condition number and the combination of task difficulty, gradient noise and network capacity.
Additional results in the lower part of \cref{tab:depths_beyond} show that while our methods are trained using five inner-loop update steps and achieve very fast adaptation from the very first step onwards, they are additionally able to continuously adapt further beyond their training horizon when given the possibility.   

\noindent
\begin{minipage}{\textwidth}
  \begin{minipage}[b]{0.68\textwidth}
  \centering
    \captionof{table}{\textbf{Influence of condition parameter selection} on test accuracy (CUB dataset, Conv4). The condition loss~$\mathcal{L}_{\kappa}$ is enforced w.r.t. one or multiple parameter groups. \textit{cls}: classifier (last layer): \textit{emb}: embedding layer; \textit{eBN}: BN of embedding layer. Highlighted in gray is the `default' setting used throughout this work (simplicity).}
    \label{tab:par_sel_tab}
    \scalebox{0.77}
    {
    \setlength{\tabcolsep}{3pt}
    \begin{tabular}[b]{@{} *3c | *5c@{}} 
    \specialrule{.2em}{.1em}{.1em}
        \multicolumn{3}{c}{\textbf{$\mathcal{L}_\kappa$ w.r.t. }} & \multicolumn{5}{c}{\textbf{5-way 5-shot}}
        \\
        \textbf{cls}  & \textbf{emb} & \textbf{eBN}  &  \textbf{step 1$\uparrow$} & \textbf{step 2$\uparrow$}  & \textbf{step 3$\uparrow$} & \textbf{step 4$\uparrow$} &\textbf{step 5$\uparrow$} \\    \midrule
          &   &   & 
          $20.04{\scriptstyle \pm 0.02}$ & $24.16{\scriptstyle \pm 0.38}$ & $64.30{\scriptstyle \pm 0.84}$ & $74.71{\scriptstyle \pm 0.78}$ & $77.06{\scriptstyle \pm 0.69}$ 
          \\ \midrule\midrule
        \rowcolor{Gray!20!White} \checkmark  &  &  &
          $61.78{\scriptstyle \pm 0.75}$ & $73.28{\scriptstyle \pm 0.71}$ & $75.64{\scriptstyle \pm 0.71}$ & $76.75{\scriptstyle \pm 0.68}$ & $77.24{\scriptstyle \pm 0.67}$ 
          \\
             & \checkmark &  &
          $59.85{\scriptstyle \pm 0.80}$ & $\mathbf{73.54{\scriptstyle \pm 0.75}}$ & $\mathbf{76.61{\scriptstyle \pm 0.69}}$ & $\mathbf{77.73{\scriptstyle \pm 0.67}}$ & $\mathbf{78.14{\scriptstyle \pm 0.67}}$ 
          \\
           &  & \checkmark &
          $62.22{\scriptstyle \pm 0.79}$ & $69.87{\scriptstyle \pm 0.79}$ & $73.04{\scriptstyle \pm 0.77}$ & $74.15{\scriptstyle \pm 0.74}$ & $74.86{\scriptstyle \pm 0.72}$ 
          \\
          \checkmark & \checkmark &  &
          $\mathbf{64.27{\scriptstyle \pm 0.77}}$ & $73.45{\scriptstyle \pm 0.71}$ & $76.02{\scriptstyle \pm 0.69}$ & $77.30{\scriptstyle \pm 0.67}$ & $77.77{\scriptstyle \pm 0.68}$ 
          \\
            & \checkmark & \checkmark &
          $55.54{\scriptstyle \pm 0.83}$ & $71.63{\scriptstyle \pm 0.76}$ & $75.55{\scriptstyle \pm 0.71}$ & $76.78{\scriptstyle \pm 0.70}$ & $77.12{\scriptstyle \pm 0.68}$ 
          \\
             \checkmark &  & \checkmark &
          $63.46{\scriptstyle \pm 0.76}$ & $73.49{\scriptstyle \pm 0.74}$ & $76.31{\scriptstyle \pm 0.69}$ & $77.67{\scriptstyle \pm 0.67}$ & $78.10{\scriptstyle \pm 0.67}$ 
          \\
            \checkmark & \checkmark & \checkmark &
          $60.93{\scriptstyle \pm 0.81}$ & $71.39{\scriptstyle \pm 0.75}$ & $74.83{\scriptstyle \pm 0.71}$ & $76.40{\scriptstyle \pm 0.69}$ & $77.28{\scriptstyle \pm 0.70}$ 
          \\
        \bottomrule
    \end{tabular}
    }
  \end{minipage}
  \hfill
  \begin{minipage}[b]{0.30\textwidth}
    \centering
    \includegraphics[width=\textwidth]{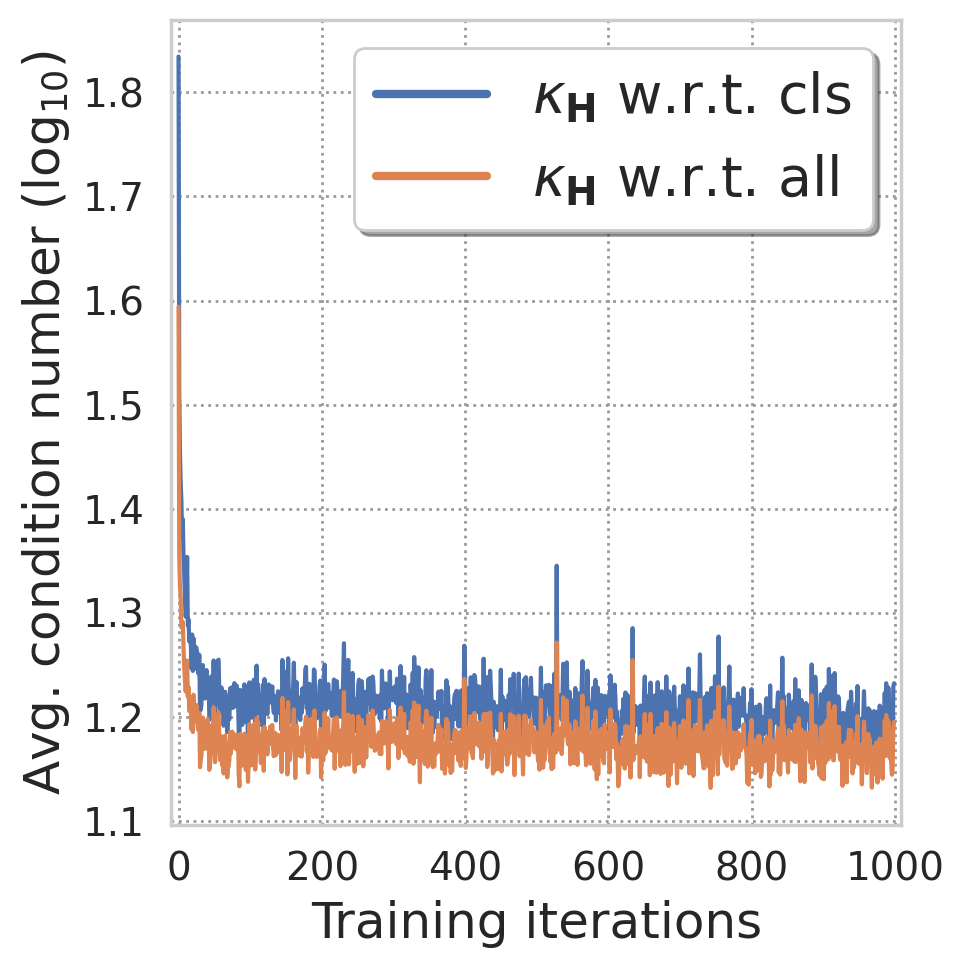}
    \captionof{figure}{Average condition number during training w.r.t \textit{cls} subset and \textit{all} parameters.}
    \label{fig:par_sel}
  \end{minipage}
\end{minipage}
\vspace{-0.7em}

\subsection{Many-way and multi-shot scenarios}
\label{sec:manyway_manyshot}
We finally investigate the influence of a better conditioned parameter space for $N$-way $K$-shot applications with more than the `default' $N\!=\!5$ classes and \ignore{across different numbers of provided support samples, \ie shots }$K\!>\!5$ support samples. \cref{fig:manyway_manyshot}\,(\subref{subfig:manyway}) demonstrates that our method sustains the previously demonstrated fast adaptation from the first step on and seems to handle settings with the increased difficulty of more classes slightly better than the baseline. Scenarios that provide more than $5$~samples per class (\cref{fig:manyway_manyshot}\,(\subref{subfig:multishot})) show that while our method continues to consistently outperform the baseline by significant margins during the first few steps (\eg by $53\%$ for $20$-shot at step $3$), the convergence of the unconstrained baseline seems to get delayed even further with increasing sample numbers -- indicating that the conditioning property can not be improved simply through more available data, but is in contrast rather adversely affected.

\begin{wrapfigure}{R}{6.7cm}
    \vspace{-1.3em}
    \centering
    \hspace{-0.7em}
    \vspace{-0.5em}
    \includegraphics[width=0.48\textwidth]{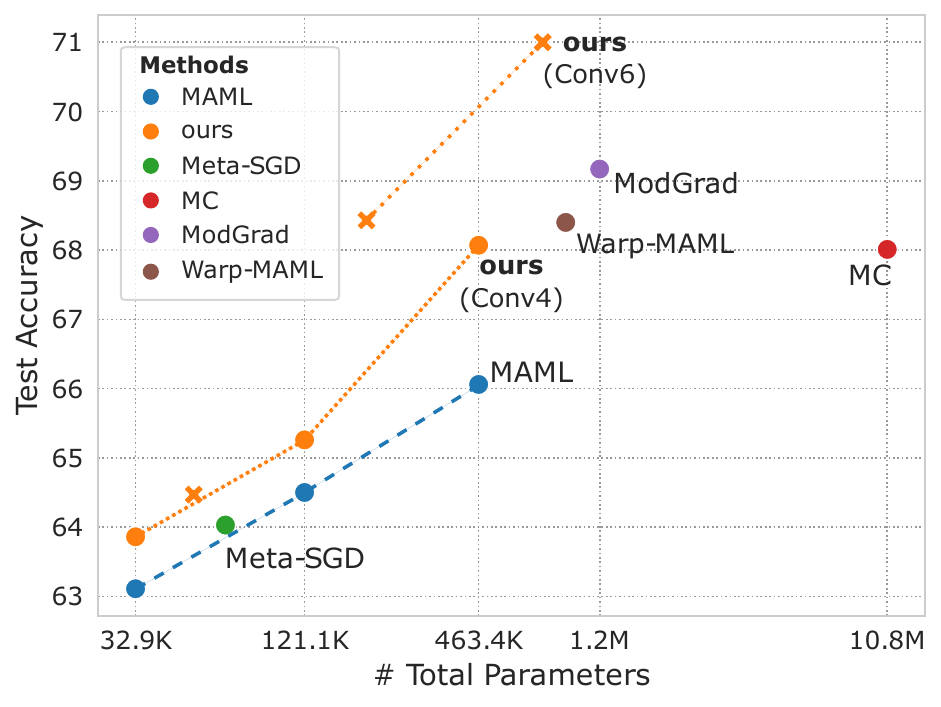}
    \caption{\textbf{Preconditioning methods, number of parameters and accuracies} obtained for $5$-way $5$-shot evaluated on the \textit{mini}ImageNet test set.}\label{fig:precondparams2}
    \vspace{-0.5em}
\end{wrapfigure}

\subsection{Preconditioning methods -- Number of parameters and performance}
We compare our approach to other recently-published preconditioning methods for a 5-way 5-shot classification task evaluated on the \textit{mini}ImageNet test set. 
In this context, we want to draw attention the fact that while most published methods compare based on the used backbone, it is worth considering their introduction of additional parameters that are required at both training and inference time. We therefore outline the gains in accuracy in relation to the total number of parameters in \cref{fig:precondparams2}.
While all methods are able to improve the test accuracy upon their respective baselines, this comes at a cost. 
Meta-SGD~\cite{li2017_metasgd} is able to improve by $1.5\%$ relative to its Conv4\,(32) MAML baseline~\cite{finn2017_maml} but doubles the number of parameters (+$100\%$). ModGrad~\cite{simon2020_gradientPrecon} is able to notably outperform its Conv4\,(64) baseline by $7.2\%$ but requires an increase in parameters of +$873\%$, while Warp-MAML~\cite{flennerhag2019_metawarpedgrad} and MC~\cite{park2019_metacurvature} achieve their relative improvements of $3.5\%$ and $3.0\%$ at the cost of incurring additional parameter costs of +$96\%$ and +$2235\%$, respectively (both Conv4\,(128)). Our approach in contrast improves upon all baselines without adding any additional parameters -- a property that allows us to instead choose larger backbones like Conv6 and thus significantly outperform other methods across the entire parameter-vs.-accuracy spectrum. Please refer to the supplementary material for more detailed information.

\subsection{Taking fewer but bigger steps at inference time}
Our motivational example depicted in \cref{fig:motivation} implies that well-conditioned parameter space should provide a more `direct' path towards the optimum and thus enable us to take fewer but bigger steps. We therefore investigate the behaviour of our method when taking \textit{only one step} but with a \textit{greater step size} at inference time. Note that the network has been trained as in previous experiments with the default setting of using 5 inner-loop adaptation steps with a learning rate `lr' during training.

Our obtained results (\cref{tab:stepchanges}) clearly indicate that a better conditioned parameter space does indeed provide the ability to take fewer but bigger steps towards the minimum to adapt faster at inference time.  While our approach evaluated on the same setting as used during training (5 steps with `lr') still slightly outperforms its 1-step variants and thus hints at a not (yet) ideally-conditioned parameter space, our method taking only 1 step with an appropriately scaled `$5\cdot$lr' step size demonstrates promising task adaptation and even outperforms the respective 5-step MAML baseline model.

\begin{table*}[h]
\caption{\textbf{Varying number of steps and step size at inference time.} Test accuracy for $5$-way $5$-shot on \textit{mini}ImageNet. Model has been trained with default settings of 5 inner-loop update steps with learning rate~lr. Reported are the classification accuracies on the test set, averaged over 600 tasks.}\label{tab:stepchanges}
\vspace{-8pt}
    \begin{center}
    \scalebox{0.8}
    {
    \hspace{-3.39mm}
    {
    \setlength{\tabcolsep}{6pt}
    \begin{tabular}{l cccccc|c}
\specialrule{.2em}{.1em}{.1em}
\textbf{Conv6} & \textbf{0\,steps} & \textbf{1\,step - lr} & \textbf{1\,step - $\boldsymbol{2\cdot}$lr} & \textbf{1\,step - $\boldsymbol{3\cdot}$lr} & \textbf{1\,step - $\boldsymbol{4\cdot}$lr} & \textbf{1\,step - $\boldsymbol{5\cdot}$lr} & \textbf{5\,steps - lr}\\
\toprule\toprule
MAML & $20.00\scriptstyle\pm0.00$ & $20.19\scriptstyle\pm0.07$ & $20.00\scriptstyle\pm0.00$ & $20.14\scriptstyle\pm0.05$ & $23.28\scriptstyle\pm0.37$ & $26.64\scriptstyle\pm0.51$ & $65.96\pm0.71$\\ 
ours & $20.14\scriptstyle\pm0.60$ & $62.31\scriptstyle\pm0.72$ & $65.67\scriptstyle\pm0.71$ & $65.87\scriptstyle\pm0.71$ & $65.72\scriptstyle\pm0.71$ & $66.54\scriptstyle\pm0.70$ & $68.43\pm0.71$\\  
\bottomrule
    \end{tabular}
    }
    }
    \end{center}
    \vskip -1.0em
\end{table*}

\section{Limitations} \label{sec:limits}
Applying our proposed condition loss in its naive form to all parameters of very deep networks might be infeasible due to high parameter counts involved in computing the Jacobians. Our experiments however clearly demonstrate that constraining a representative subset like the classifier is sufficient to improve the overall conditioning property of the model and can significantly improve the convergence behaviour during initial steps (\cref{sec:constrain_param_subset}).

Our method further makes use of second order derivatives (just like the baseline MAML) and thus needs to store the computational graph throughout the inner-loop updates, a property that didn't raise any issues in our experiments but might be a point of concern for applications with extremely limited memory -- a general point worth considering with second-order methods.

\begin{table*}
\caption{\textbf{Increasing the network depth and adaptation steps.} Evaluations are conducted on all five popular FSL datasets: CUB-200-2011~\cite{WahCUB_200_2011}, \textit{mini}ImageNet~\cite{vinyals2016_matchingnet}, \textit{tiered}ImageNet~\cite{ren2018_metasemisup}, CIFAR-FS~\cite{bertinetto2019_cifarfsl} and FC100~\cite{oreshkin2018_tadam}. Reported are the classification accuracies on the unseen test set, averaged over 600 tasks.}
\vspace{-8pt}
    \begin{center}
    \scalebox{0.671}
    {
    \hspace{-3.39mm}
    {
    \setlength{\tabcolsep}{3pt}

    \begin{tabular}{@{}c ll| *5c | *5c@{}} 
    \specialrule{.2em}{.1em}{.1em}
          & & & \multicolumn{5}{c}{\textbf{5-shot}} &  \multicolumn{5}{c}{\textbf{1-shot}}\\
          & \textbf{Network}  & \textbf{Method} & \textbf{step 1$\uparrow$}  & \textbf{step 2$\uparrow$} & \textbf{step 3$\uparrow$}  & \textbf{step 4$\uparrow$} & \textbf{step 5$\uparrow$} & \textbf{step 1$\uparrow$}  & \textbf{step 2$\uparrow$} & \textbf{step 3$\uparrow$}  & \textbf{step 4$\uparrow$} & \textbf{step 5$\uparrow$}\\    \midrule 
         \multirow{8}{*}{\STAB{\rotatebox[origin=c]{90}{\textit{mini}ImageNet}}} 
         & \multirow{2}{*}{Conv4}
          & MAML & $20.00{\scriptstyle \pm 0.01}$ & $20.00{\scriptstyle \pm 0.00}$ & $52.87{\scriptstyle \pm 0.75}$ & $61.12{\scriptstyle \pm 0.75}$ & $64.50{\scriptstyle \pm 0.69}$ 
          & $38.76{\scriptstyle \pm 0.73}$ & $42.33{\scriptstyle \pm 0.74}$ & $45.99{\scriptstyle \pm 0.77}$ & $47.67{\scriptstyle \pm 0.81}$ & $48.15{\scriptstyle \pm 0.80}$ 
          \\
          & & ours & $56.09{\scriptstyle \pm 0.66}$ & $62.28{\scriptstyle \pm 0.70}$ & $64.01{\scriptstyle \pm 0.67}$ & $64.89{\scriptstyle \pm 0.67}$ & $65.26{\scriptstyle \pm 0.67}$ 
          & $43.68{\scriptstyle \pm 0.74}$ & $47.65{\scriptstyle \pm 0.79}$ & $48.50{\scriptstyle \pm 0.79}$ & $48.76{\scriptstyle \pm 0.79}$ & $48.94{\scriptstyle \pm 0.80}$ 
          \\
          \cmidrule{2-13}
         & \multirow{2}{*}{Conv6} 
          & MAML & $20.19{\scriptstyle \pm 0.07}$ & $24.17{\scriptstyle \pm 0.39}$ & $57.25{\scriptstyle \pm 0.72}$ & $64.24{\scriptstyle \pm 0.73}$ & $65.96{\scriptstyle \pm 0.71}$ 
          & $21.40{\scriptstyle \pm 0.23}$ & $30.85{\scriptstyle \pm 0.66}$ & $45.91{\scriptstyle \pm 0.87}$ & $50.30{\scriptstyle \pm 0.88}$ & $51.22{\scriptstyle \pm 0.88}$ 
          \\
          & & ours & $62.31{\scriptstyle \pm 0.72}$ & $66.66{\scriptstyle \pm 0.71}$ & $67.63{\scriptstyle \pm 0.72}$ & $68.21{\scriptstyle \pm 0.71}$ & $68.43{\scriptstyle \pm 0.71}$ 
          & $50.92{\scriptstyle \pm 0.85}$ & $52.98{\scriptstyle \pm 0.89}$ & $53.18{\scriptstyle \pm 0.89}$ & $53.28{\scriptstyle \pm 0.89}$ & $53.34{\scriptstyle \pm 0.89}$ 
          \\
          \cmidrule{2-13}
        & \multirow{2}{*}{ResNet10} 
         & MAML & $20.06{\scriptstyle \pm 0.05}$ & $20.03{\scriptstyle \pm 0.04}$ & $41.74{\scriptstyle \pm 0.74}$ & $63.28{\scriptstyle \pm 0.72}$ & $69.43{\scriptstyle \pm 0.71}$ 
         & $20.02{\scriptstyle \pm 0.04}$ & $22.14{\scriptstyle \pm 0.30}$ & $50.99{\scriptstyle \pm 0.84}$ & $56.47{\scriptstyle \pm 0.82}$ & $57.35{\scriptstyle \pm 0.80}$ 
         \\
          & & ours & $53.93{\scriptstyle \pm 0.75}$ & $68.96{\scriptstyle \pm 0.72}$ & $71.40{\scriptstyle \pm 0.70}$ & $72.18{\scriptstyle \pm 0.69}$ & $72.46{\scriptstyle \pm 0.67}$ 
          & $50.44{\scriptstyle \pm 0.82}$ & $55.86{\scriptstyle \pm 0.84}$ & $57.44{\scriptstyle \pm 0.84}$ & $57.97{\scriptstyle \pm 0.85}$ & $58.20{\scriptstyle \pm 0.85}$ 
          \\
         \cmidrule{2-13}
        & \multirow{2}{*}{ResNet18} 
          & MAML & $20.12{\scriptstyle \pm 0.06}$ & $20.00{\scriptstyle \pm 0.00}$ & $36.64{\scriptstyle \pm 0.69}$ & $65.01{\scriptstyle \pm 0.74}$ & $71.06{\scriptstyle \pm 0.74}$ 
          & $20.02{\scriptstyle \pm 0.03}$ & $20.79{\scriptstyle \pm 0.17}$ & $52.81{\scriptstyle \pm 0.86}$ & $56.42{\scriptstyle \pm 0.90}$ & $56.84{\scriptstyle \pm 0.90}$ 
          \\
          & & ours & $68.60{\scriptstyle \pm 0.71}$ & $72.05{\scriptstyle \pm 0.70}$ & $72.93{\scriptstyle \pm 0.69}$ & $73.10{\scriptstyle \pm 0.69}$ & $73.28{\scriptstyle \pm 0.68}$ 
          & $54.46{\scriptstyle \pm 0.90}$ & $57.01{\scriptstyle \pm 0.91}$ & $57.31{\scriptstyle \pm 0.91}$ & $57.52{\scriptstyle \pm 0.91}$ & $57.64{\scriptstyle \pm 0.91}$ 
          \\
          \bottomrule \\ \toprule
          & \textbf{Dataset}  & \textbf{Method} & \textbf{step 1$\uparrow$}  & \textbf{step 2$\uparrow$} & \textbf{step 3$\uparrow$}  & \textbf{step 4$\uparrow$} & \textbf{step 5$\uparrow$} & \textbf{step 1$\uparrow$}  & \textbf{step 2$\uparrow$} & \textbf{step 3$\uparrow$}  & \textbf{step 4$\uparrow$} & \textbf{step 5$\uparrow$}\\    \midrule 
         \multirow{8}{*}{\STAB{\rotatebox[origin=c]{90}{ResNet18}}} 
         & \multirow{2}{*}{CUB}
          & MAML & $20.00{\scriptstyle \pm 0.00}$ & $20.48{\scriptstyle \pm 0.13}$ & $43.96{\scriptstyle \pm 0.86}$ & $79.56{\scriptstyle \pm 0.74}$ & $83.56{\scriptstyle \pm 0.61}$ 
          & $20.02{\scriptstyle \pm 0.02}$ & $25.57{\scriptstyle \pm 0.54}$ & $72.20{\scriptstyle \pm 0.98}$ & $74.06{\scriptstyle \pm 0.95}$ & $74.52{\scriptstyle \pm 0.94}$ 
          \\
          & & ours & $81.57{\scriptstyle \pm 0.61}$ & $84.00{\scriptstyle \pm 0.55}$ & $84.47{\scriptstyle \pm 0.54}$ & $84.63{\scriptstyle \pm 0.54}$ & $84.66{\scriptstyle \pm 0.54}$ 
          & $66.55{\scriptstyle \pm 1.03}$ & $72.93{\scriptstyle \pm 0.98}$ & $74.38{\scriptstyle \pm 0.97}$ & $74.96{\scriptstyle \pm 0.97}$ & $75.22{\scriptstyle \pm 0.97}$ 
          \\
          \cmidrule{2-13}
         & \multirow{2}{*}{\textit{ti}ImgNet} 
         & MAML & $20.00{\scriptstyle \pm 0.01}$ & $20.10{\scriptstyle \pm 0.05}$ & $38.93{\scriptstyle \pm 0.79}$ & $68.57{\scriptstyle \pm 0.88}$ & $73.90{\scriptstyle \pm 0.79}$ 
         & $20.02{\scriptstyle \pm 0.02}$ & $21.19{\scriptstyle \pm 0.23}$ & $51.30{\scriptstyle \pm 0.95}$ & $56.80{\scriptstyle \pm 1.00}$ & $57.71{\scriptstyle \pm 1.00}$ 
         \\
          & & ours & $70.13{\scriptstyle \pm 0.78}$ & $73.60{\scriptstyle \pm 0.74}$ & $74.31{\scriptstyle \pm 0.74}$ & $74.55{\scriptstyle \pm 0.74}$ & $74.67{\scriptstyle \pm 0.74}$ 
         & $54.79{\scriptstyle \pm 0.96}$ & $57.44{\scriptstyle \pm 0.97}$ & $57.64{\scriptstyle \pm 0.97}$ & $57.71{\scriptstyle \pm 0.98}$ & $57.80{\scriptstyle \pm 0.98}$ 
         \\
          \cmidrule{2-13}
        & \multirow{2}{*}{CIFAR-FS} 
         & MAML & $20.01{\scriptstyle \pm 0.01}$ & $20.04{\scriptstyle \pm 0.03}$ & $45.75{\scriptstyle \pm 0.83}$ & $75.69{\scriptstyle \pm 0.81}$ & $79.59{\scriptstyle \pm 0.70}$ 
         & $20.00{\scriptstyle \pm 0.00}$ & $23.41{\scriptstyle \pm 0.39}$ & $63.34{\scriptstyle \pm 0.99}$ & $67.05{\scriptstyle \pm 0.97}$ & $67.44{\scriptstyle \pm 0.99}$ 
         \\
          & & ours & $77.31{\scriptstyle \pm 0.71}$ & $79.49{\scriptstyle \pm 0.67}$ & $79.96{\scriptstyle \pm 0.66}$ & $80.09{\scriptstyle \pm 0.65}$ & $80.18{\scriptstyle \pm 0.65}$ 
         & $65.73{\scriptstyle \pm 1.03}$ & $68.39{\scriptstyle \pm 1.04}$ & $68.75{\scriptstyle \pm 1.04}$ & $69.04{\scriptstyle \pm 1.03}$ & $69.09{\scriptstyle \pm 1.03}$ 
         \\
         \cmidrule{2-13}
        & \multirow{2}{*}{FC100} 
          & MAML & $20.00{\scriptstyle \pm 0.00}$ & $20.00{\scriptstyle \pm 0.00}$ & $20.00{\scriptstyle \pm 0.00}$ & $48.46{\scriptstyle \pm 0.73}$ & $48.56{\scriptstyle \pm 0.76}$ 
         & $20.00{\scriptstyle \pm 0.00}$ & $20.04{\scriptstyle \pm 0.02}$ & $32.30{\scriptstyle \pm 0.64}$ & $34.37{\scriptstyle \pm 0.69}$ & $35.19{\scriptstyle \pm 0.70}$ 
         \\
         & & ours & $47.11{\scriptstyle \pm 0.73}$ & $50.16{\scriptstyle \pm 0.73}$ & $50.82{\scriptstyle \pm 0.74}$ & $51.08{\scriptstyle \pm 0.74}$ & $51.22{\scriptstyle \pm 0.74}$ 
         & $33.24{\scriptstyle \pm 0.69}$ & $36.05{\scriptstyle \pm 0.71}$ & $36.70{\scriptstyle \pm 0.72}$ & $36.90{\scriptstyle \pm 0.72}$ & $37.02{\scriptstyle \pm 0.72}$ 
         \\
          \bottomrule \\ \toprule
          & \textbf{Network}  & \textbf{Method} & \textbf{step 5$\uparrow$} & \textbf{step 10$\uparrow$}  & \textbf{step 25$\uparrow$} & \textbf{step 50$\uparrow$} & \textbf{step 100$\uparrow$}  &  \textbf{step 5$\uparrow$} & \textbf{step 10$\uparrow$}  & \textbf{step 25$\uparrow$} & \textbf{step 50$\uparrow$} &\textbf{step 100$\uparrow$} \\  \midrule
         \multirow{4}{*}{\STAB{\rotatebox[origin=c]{90}{CUB}}} 
         & Conv4 & ours & $77.24{\scriptstyle \pm 0.67}$ & $77.48{\scriptstyle \pm 0.68}$ & $77.74{\scriptstyle \pm 0.67}$ & $77.88{\scriptstyle \pm 0.68}$ & $78.06{\scriptstyle \pm 0.68}$ 
          & $63.26{\scriptstyle \pm 0.98}$ & $63.42{\scriptstyle \pm 0.99}$ & $63.44{\scriptstyle \pm 1.00}$ & $63.54{\scriptstyle \pm 1.00}$ & $63.47{\scriptstyle \pm 1.00}$ 
          \\
          & Conv6 & ours & $80.65{\scriptstyle \pm 0.63}$ & $80.90{\scriptstyle \pm 0.63}$ & $81.15{\scriptstyle \pm 0.62}$ & $81.25{\scriptstyle \pm 0.62}$ & $81.35{\scriptstyle \pm 0.62}$ 
          & $68.87{\scriptstyle \pm 1.06}$ & $69.12{\scriptstyle \pm 1.06}$ & $69.19{\scriptstyle \pm 1.06}$ & $69.16{\scriptstyle \pm 1.06}$ & $69.15{\scriptstyle \pm 1.06}$ 
          \\
          & ResNet10 & ours & $83.82{\scriptstyle \pm 0.59}$ & $84.09{\scriptstyle \pm 0.58}$ & $84.11{\scriptstyle \pm 0.59}$ & $84.26{\scriptstyle \pm 0.57}$ & $84.28{\scriptstyle \pm 0.57}$ 
          & $74.99{\scriptstyle \pm 0.96}$ & $74.99{\scriptstyle \pm 0.97}$ & $75.19{\scriptstyle \pm 0.97}$ & $75.24{\scriptstyle \pm 0.97}$ & $75.33{\scriptstyle \pm 0.97}$ 
          \\
          &ResNet18 & ours & $84.66{\scriptstyle \pm 0.54}$ & $84.83{\scriptstyle \pm 0.53}$ & $84.94{\scriptstyle \pm 0.53}$ & $85.01{\scriptstyle \pm 0.53}$ & $85.01{\scriptstyle \pm 0.53}$ 
          & $75.22{\scriptstyle \pm 0.97}$ & $75.52{\scriptstyle \pm 0.95}$ & $75.66{\scriptstyle \pm 0.95}$ & $75.74{\scriptstyle \pm 0.95}$ & $75.81{\scriptstyle \pm 0.95}$ 
          \\
        \bottomrule
    \end{tabular}
    }
    }
    \end{center}
    \label{tab:depths_beyond}
    \vskip -0.2in
\end{table*}

\section{Related work}
\label{sec:relatedwork}
\textbf{Gradient-based meta-learning}. \ignore{While several different approaches to tackle few-shot problems have been developed over the last years, methods most related to our approach are `gradient-' \textit{aka} `optimization-based' and are sometimes described as the \textit{most general} form of algorithms to tackle meta-learning scenarios~\cite{yang2021_freelunch}.} Methods most related to our approach generally aim to find a set of model parameters that can adapt to novel tasks within a few steps of gradient descent. 
Early seminal works~\cite{andrychowicz2016_learningtolearn,finn2017_maml} have spiked significant follow-up research like first order variants~\cite{nichol2018_reptile}, probabilistic reformulations~\cite{finn2018_probabilisticMAML} and generative approaches~\cite{rusu2018_leo}, among many more~\cite{ravi2017_optfsl,rajeswaran2019_implicitgrad,zintgraf2019_cavia}. While some have \ignore{focused on in-depth analysis and }derived theoretical guarantees~\cite{balcan2019_provable,grant2018_recasting}, others improved training stability~\cite{antoniou2018_mamlpp}.
Two main challenges have to be overcome to solve a previously unseen task with a highly limited number of data samples in this way: overfitting~\cite{sun2019_metatransfer} and rapid adaptation. 
While several works have attempted the first~\cite{rusu2018_leo,zhou2018_conceptspace,zintgraf2019_cavia}, learning an efficient way of updating the model's parameters that enables fast adaptation remains an open problem~\cite{flennerhag2019_metawarpedgrad}\ignore{and has received far less attention}. 
On common drawback is the unknown adaptation quality, \ie how many updates have to be performed to achieve satisfactory results on a new task. This is commonly solved by either simply choosing the de-facto default value of five steps, or the highest number of steps that seems feasible for a given task based on empirical results. In contrast, our proposed approach of constraining methods to learn a well-conditioned parameter space achieves fast adaptation from the very first step onwards and gradually increases, allowing users to dynamically choose the number at inference time. \par
\textbf{Gradient modulation}.
Few works have attempted to tackle the challenge of finding a more suitable way of updating the parameters. Early methods use LSTMs to update models iteratively by learning both the initialisation and update strategy\ignore{ via meta-learning}~\cite{andrychowicz2016_learningtolearn,ravi2017_optfsl} but are difficult to train~\cite{li2017_metasgd}. This idea has been further developed by~\citet{li2017_metasgd} towards \ignore{an SGD-like method called }Meta-SGD, where the initialisation is meta-learned together with update direction and parameter-specific learning rates. While this can be seen as a form of preconditioned optimisation, more direct attempts \ignore{towards manipulating the update process via multiplying estimated preconditioning matrices to transform or modulate the gradients }have recently been proposed.
An interleaving-based method is used to improve the convergence speed \ignore{of gradient descent }via meta-learning a parameterized preconditioning matrix by~\citet{flennerhag2019_metawarpedgrad}, while~\citet{park2019_metacurvature} instead learn curvature information to transform the gradients driving the task-specific adaptation \ignore{to yield better generalisation} and~\citet{simon2020_gradientPrecon} \ignore{introduce a way to }modulate the gradients using an additional module and
low-rank approximation. 
However, all these approaches aim to estimate some form of additional preconditioning matrix to transform the gradients, incurring an extensive number of additional parameters and overhead at both training and inference time.
In contrast, our method imposes a constraint directly onto the parameter space of our network and thus targets the same idea of a well-conditioned optimisation space but within the already existing capacity of the network. \par
\textbf{Condition number.} The relevance of the condition number has been discussed in the context of deep learning by~\citet{goodfellow2016_deepbook}, and few works have made use of it as a `passive metric', \eg to monitor and assess quality of convergence in the context of Generative Adversarial Networks~\cite{zuo2021_improved}.
We take a different approach and show that in the bi-level nature of gradient-based few-shot meta-learning, we can actively manipulate the learning process of the function to learn a well-posed optimisation problem that demonstrates superior convergence properties. 

\section{Conclusion}
In this paper, we demonstrated how to actively encourage a deep network to learn a better-conditioned parameter space that allows significantly faster adaptation to novel tasks in few-shot settings. To this end, we introduced a new constraint that is easily integrated into optimisation-based few-shot methods via an additive loss term. Our constraint is based around the condition number of local curvatures and does not incur any additional parameters. We showed the efficacy of our approach through comprehensive experiments on five datasets with various different architectures and few-show scenarios, significantly outperforming unconstrained methods during initial adaptation.\par
\noindent \textbf{Acknowledgements.} The authors would like to thank \citet{finn2017_maml} and \citet{chen2019closerfewshot} for sharing their insights, experimental details and code.\par
\noindent Part of this research was conducted while MH was with Monash University.
This research was partially undertaken using the LIEF HPC-GPGPU Facility hosted at the University of Melbourne. This Facility was established with the assistance of LIEF Grant LE170100200.

{
\small
\bibliography{biblio_conditioning}
\bibliographystyle{icml2022}
}



\end{document}


\maketitle 
\appendix


\section{Datasets used for experiments}
This section provides additional background information about the five datasets we use to train and evaluate the baseline and our approach in the main paper. \par
\textbf{CUB-200-2011}. The Caltech-UCSD Birds-200-2011, \textit{aka} CUB-200-2011 or simply `CUB', is focused on fine-grained image classification tasks. It was proposed by~\citet{WahCUB_200_2011} and contains $11{,}788$ images of 200 subcategories. We follow previous works like~\citet{chen2019closerfewshot} and use the evaluation protocol introduced by \citet{hilliard2018_fewcubsplit}, splitting the dataset into 100 classes for training, 50 for validation and 50 for testing.\par
\textbf{\textit{mini}ImageNet}. The \textit{mini}ImageNet dataset has been initially proposed by~\cite{vinyals2016_matchingnet} and the specific few-shot settings have been further refined in later work by~\cite{ravi2017_optfsl}. It consists of a 100 class subset selected from the ImageNet dataset~\cite{russakovsky_2015imagenet} with 600 images for each class. The dataset is split into 64 training, 16 validation and 20 test classes. \par
\textbf{\textit{tiered}ImageNet}. The \textit{tiered}ImageNet~\cite{ren2018_metasemisup} dataset is equally a subset of classes selected from the bigger ImageNet~\cite{russakovsky_2015imagenet} dataset, however with a different structure and substantially larger set of classes. It is composed of 34 super-classes with a total of 608 categories that are split into 20, 6 and 8 super-classes totalling in $779{,}165$ images. This unique split aims at achieving better separation between training, validation and testing, respectively.\par
\textbf{CIFAR-FS}. The CIFAR-FS dataset~\cite{bertinetto2019_cifarfsl} contains essentially the data from the CIFAR100~\cite{krizhevsky2009_cifar100} dataset and splits the 100 categories of 600 images each into 64 training, 16 validation and 20 test classes.\par
\textbf{FC-100}. The FC-100 dataset~\cite{oreshkin2018_tadam} is similarly derived from CIFAR100~\cite{krizhevsky2009_cifar100} and split into 60 training, 20 validation and 20 test classes, but follows a splitting approach more similar to \textit{tiered}ImageNet to increase separation between classes and difficulty.\par

\section{Effect of increased network depth}
All experiments are conducted with equal contribution of all adaptation steps to the conditioning loss (as defined in Equation~(9) of the main paper), with the conditioning constraint enforced with respect to the parameters of the classifier. To provide insights into the effect of increasing the depth and number of parameters onto the conditioning performance, we evaluate the following architectures on all five popular few-shot classification benchmarks for 5-shot and 1-shot settings: Convolutional networks with 4 layers (Conv4) and 6 layers (Conv6), as well as the two residual networks ResNet10 and ResNet18. While a selection of the results has been discussed in the main paper, the test accuracies on all datasets across all architectures are presented in \cref{tab:depths_sup}.
\begin{table*}[h]
\caption{\textbf{Increasing the network's depth and number of parameters.} Evaluations are conducted on all five popular FSL datasets: CUB-200-2011~\cite{WahCUB_200_2011}, \textit{mini}ImageNet~\cite{vinyals2016_matchingnet}, \textit{tiered}ImageNet~\cite{ren2018_metasemisup}, CIFAR-FS~\cite{bertinetto2019_cifarfsl} and FC100~\cite{oreshkin2018_tadam}. Reported are the classification accuracies on the unseen test set, averaged over 600 tasks following previous works like~\citet{chen2019closerfewshot}.}
    \begin{center}
    \scalebox{0.675}
    {
    \hspace{-3.39mm}
    {
    \setlength{\tabcolsep}{3pt}
    \begin{tabular}{@{}c ll| *5c | *5c@{}} 
    \specialrule{.2em}{.1em}{.1em}
          & & & \multicolumn{5}{c}{\textbf{5-shot}} &  \multicolumn{5}{c}{\textbf{1-shot}}\\
          & \textbf{Network}  & \textbf{Method} & \textbf{step 1$\uparrow$}  & \textbf{step 2$\uparrow$} & \textbf{step 3$\uparrow$}  & \textbf{step 4$\uparrow$} & \textbf{step 5$\uparrow$} & \textbf{step 1$\uparrow$}  & \textbf{step 2$\uparrow$} & \textbf{step 3$\uparrow$}  & \textbf{step 4$\uparrow$} & \textbf{step 5$\uparrow$}\\    \midrule 
         \multirow{8}{*}{\STAB{\rotatebox[origin=c]{90}{CUB}}} 
         & \multirow{2}{*}{{Conv4}} 
          & MAML & $20.04{\scriptstyle \pm 0.02}$ & $24.16{\scriptstyle \pm 0.38}$ & $64.30{\scriptstyle \pm 0.84}$ & $74.71{\scriptstyle \pm 0.78}$ & $77.06{\scriptstyle \pm 0.69}$ 
          &$50.53{\scriptstyle \pm 0.92}$ & $57.94{\scriptstyle \pm 0.97}$ & $60.40{\scriptstyle \pm 0.98}$ & $60.90{\scriptstyle \pm 0.98}$ & $61.09{\scriptstyle \pm 0.98}$ 
          \\
          & & ours  & $61.78{\scriptstyle \pm 0.75}$ & $73.28{\scriptstyle \pm 0.71}$ & $75.64{\scriptstyle \pm 0.71}$ & $76.75{\scriptstyle \pm 0.68}$ & $77.24{\scriptstyle \pm 0.67}$ 
          &$55.93{\scriptstyle \pm 0.92}$ & $61.62{\scriptstyle \pm 0.96}$ & $62.82{\scriptstyle \pm 0.98}$ & $63.20{\scriptstyle \pm 0.99}$ & $63.26{\scriptstyle \pm 0.98}$ 
          \\
          \cmidrule{2-13}
         & \multirow{2}{*}{Conv6} 
          & MAML & $20.00{\scriptstyle \pm 0.00}$ & $34.14{\scriptstyle \pm 0.62}$ & $73.58{\scriptstyle \pm 0.82}$ & $79.40{\scriptstyle \pm 0.70}$ & $80.52{\scriptstyle \pm 0.65}$ 
          & $20.06{\scriptstyle \pm 0.04}$ & $25.85{\scriptstyle \pm 0.48}$ & $62.54{\scriptstyle \pm 1.02}$ & $66.78{\scriptstyle \pm 0.98}$ &  $67.67{\scriptstyle \pm 0.98}$ 
          \\
          & & ours & $75.42{\scriptstyle \pm 0.72}$ & $79.21{\scriptstyle \pm 0.65}$ & $80.01{\scriptstyle \pm 0.65}$ & $80.44{\scriptstyle \pm 0.63}$ & $80.65{\scriptstyle \pm 0.63}$  
          & $65.15{\scriptstyle \pm 1.06}$ & $68.28{\scriptstyle \pm 1.06}$ & $68.54{\scriptstyle \pm 1.07}$ & $68.76{\scriptstyle \pm 1.06}$ & $68.87{\scriptstyle \pm 1.06}$ 
          \\
          \cmidrule{2-13}
        & \multirow{2}{*}{ResNet10} 
         & MAML & $20.00{\scriptstyle \pm 0.00}$ & $20.00{\scriptstyle \pm 0.00}$ & $20.60{\scriptstyle \pm 0.13}$ & $76.65{\scriptstyle \pm 0.79}$ & $82.13{\scriptstyle \pm 0.64}$  
         & $20.00{\scriptstyle \pm 0.00}$ & $33.55{\scriptstyle \pm 0.79}$ & $67.41{\scriptstyle \pm 1.08}$ & $72.20{\scriptstyle \pm 0.97}$ & $73.04{\scriptstyle \pm 0.96}$\\ 
         & & ours & $68.71{\scriptstyle \pm 0.79}$ & $81.36{\scriptstyle \pm 0.64}$ & $82.98{\scriptstyle \pm 0.61}$ & $83.53{\scriptstyle \pm 0.60}$ & $83.82{\scriptstyle \pm 0.59}$ 
         & $63.53{\scriptstyle \pm 1.05}$ & $71.56{\scriptstyle \pm 1.02}$ & $73.87{\scriptstyle \pm 1.08}$ & $74.72{\scriptstyle \pm 0.97}$ & $74.99{\scriptstyle \pm 0.96}$ 
         \\
         \cmidrule{2-13}
        & \multirow{2}{*}{ResNet18} 
          & MAML & $20.00{\scriptstyle \pm 0.00}$ & $20.48{\scriptstyle \pm 0.13}$ & $43.96{\scriptstyle \pm 0.86}$ & $79.56{\scriptstyle \pm 0.74}$ & $83.56{\scriptstyle \pm 0.61}$ 
          & $20.02{\scriptstyle \pm 0.02}$ & $25.57{\scriptstyle \pm 0.54}$ & $72.20{\scriptstyle \pm 0.98}$ & $74.06{\scriptstyle \pm 0.95}$ & $74.52{\scriptstyle \pm 0.94}$ 
          \\
          & & ours & $81.57{\scriptstyle \pm 0.61}$ & $84.00{\scriptstyle \pm 0.55}$ & $84.47{\scriptstyle \pm 0.54}$ & $84.63{\scriptstyle \pm 0.54}$ & $84.66{\scriptstyle \pm 0.54}$ 
          & $66.55{\scriptstyle \pm 1.03}$ & $72.93{\scriptstyle \pm 0.98}$ & $74.38{\scriptstyle \pm 0.97}$ & $74.96{\scriptstyle \pm 0.97}$ & $75.22{\scriptstyle \pm 0.97}$ 
          \\
          \midrule \midrule
          \multirow{8}{*}{\STAB{\rotatebox[origin=c]{90}{\textit{mini}ImageNet}}} 
         & \multirow{2}{*}{Conv4}
          & MAML & $20.00{\scriptstyle \pm 0.01}$ & $20.00{\scriptstyle \pm 0.00}$ & $52.87{\scriptstyle \pm 0.75}$ & $61.12{\scriptstyle \pm 0.75}$ & $64.50{\scriptstyle \pm 0.69}$ 
          & $38.76{\scriptstyle \pm 0.73}$ & $42.33{\scriptstyle \pm 0.74}$ & $45.99{\scriptstyle \pm 0.77}$ & $47.67{\scriptstyle \pm 0.81}$ & $48.15{\scriptstyle \pm 0.80}$ 
          \\
          & & ours & $56.09{\scriptstyle \pm 0.66}$ & $62.28{\scriptstyle \pm 0.70}$ & $64.01{\scriptstyle \pm 0.67}$ & $64.89{\scriptstyle \pm 0.67}$ & $65.26{\scriptstyle \pm 0.67}$ 
          & $43.68{\scriptstyle \pm 0.74}$ & $47.65{\scriptstyle \pm 0.79}$ & $48.50{\scriptstyle \pm 0.79}$ & $48.76{\scriptstyle \pm 0.79}$ & $48.94{\scriptstyle \pm 0.80}$ 
          \\
          \cmidrule{2-13}
         & \multirow{2}{*}{Conv6} 
          & MAML & $20.19{\scriptstyle \pm 0.07}$ & $24.17{\scriptstyle \pm 0.39}$ & $57.25{\scriptstyle \pm 0.72}$ & $64.24{\scriptstyle \pm 0.73}$ & $65.96{\scriptstyle \pm 0.71}$ 
          & $21.40{\scriptstyle \pm 0.23}$ & $30.85{\scriptstyle \pm 0.66}$ & $45.91{\scriptstyle \pm 0.87}$ & $50.30{\scriptstyle \pm 0.88}$ & $51.22{\scriptstyle \pm 0.88}$ 
          \\
          & & ours & $62.31{\scriptstyle \pm 0.72}$ & $66.66{\scriptstyle \pm 0.71}$ & $67.63{\scriptstyle \pm 0.72}$ & $68.21{\scriptstyle \pm 0.71}$ & $68.43{\scriptstyle \pm 0.71}$ 
          & $50.92{\scriptstyle \pm 0.85}$ & $52.98{\scriptstyle \pm 0.89}$ & $53.18{\scriptstyle \pm 0.89}$ & $53.28{\scriptstyle \pm 0.89}$ & $53.34{\scriptstyle \pm 0.89}$ 
          \\
          \cmidrule{2-13}
        & \multirow{2}{*}{ResNet10} 
         & MAML & $20.06{\scriptstyle \pm 0.05}$ & $20.03{\scriptstyle \pm 0.04}$ & $41.74{\scriptstyle \pm 0.74}$ & $63.28{\scriptstyle \pm 0.72}$ & $69.43{\scriptstyle \pm 0.71}$ 
         & $20.02{\scriptstyle \pm 0.04}$ & $22.14{\scriptstyle \pm 0.30}$ & $50.99{\scriptstyle \pm 0.84}$ & $56.47{\scriptstyle \pm 0.82}$ & $57.35{\scriptstyle \pm 0.80}$ 
         \\
          & & ours & $53.93{\scriptstyle \pm 0.75}$ & $68.96{\scriptstyle \pm 0.72}$ & $71.40{\scriptstyle \pm 0.70}$ & $72.18{\scriptstyle \pm 0.69}$ & $72.46{\scriptstyle \pm 0.67}$ 
          & $50.44{\scriptstyle \pm 0.82}$ & $55.86{\scriptstyle \pm 0.84}$ & $57.44{\scriptstyle \pm 0.84}$ & $57.97{\scriptstyle \pm 0.85}$ & $58.20{\scriptstyle \pm 0.85}$ 
          \\
         \cmidrule{2-13}
        & \multirow{2}{*}{ResNet18} 
          & MAML & $20.12{\scriptstyle \pm 0.06}$ & $20.00{\scriptstyle \pm 0.00}$ & $36.64{\scriptstyle \pm 0.69}$ & $65.01{\scriptstyle \pm 0.74}$ & $71.06{\scriptstyle \pm 0.74}$ 
          & $20.02{\scriptstyle \pm 0.03}$ & $20.79{\scriptstyle \pm 0.17}$ & $52.81{\scriptstyle \pm 0.86}$ & $56.42{\scriptstyle \pm 0.90}$ & $56.84{\scriptstyle \pm 0.90}$ 
          \\
          & & ours & $68.60{\scriptstyle \pm 0.71}$ & $72.05{\scriptstyle \pm 0.70}$ & $72.93{\scriptstyle \pm 0.69}$ & $73.10{\scriptstyle \pm 0.69}$ & $73.28{\scriptstyle \pm 0.68}$ 
          & $54.46{\scriptstyle \pm 0.90}$ & $57.01{\scriptstyle \pm 0.91}$ & $57.31{\scriptstyle \pm 0.91}$ & $57.52{\scriptstyle \pm 0.91}$ & $57.64{\scriptstyle \pm 0.91}$ 
          \\
          \midrule \midrule
          \multirow{8}{*}{\STAB{\rotatebox[origin=c]{90}{\textit{tiered}ImageNet}}} 
         & \multirow{2}{*}{Conv4}
          & MAML & $20.00{\scriptstyle \pm 0.00}$ & $22.30{\scriptstyle \pm 0.28}$ & $51.64{\scriptstyle \pm 0.77}$ & $59.76{\scriptstyle \pm 0.83}$ & $63.26{\scriptstyle \pm 0.77}$ 
          & $40.25{\scriptstyle \pm 0.80}$ & $45.17{\scriptstyle \pm 0.85}$ & $46.49{\scriptstyle \pm 0.89}$ & $47.03{\scriptstyle \pm 0.89}$ & $47.33{\scriptstyle \pm 0.90}$ 
          \\
          & & ours & $55.68{\scriptstyle \pm 0.75}$ & $61.78{\scriptstyle \pm 0.77}$ & $63.49{\scriptstyle \pm 0.77}$ & $64.49{\scriptstyle \pm 0.75}$ & $64.77{\scriptstyle \pm 0.75}$ 
          & $42.23{\scriptstyle \pm 0.83}$ & $45.97{\scriptstyle \pm 0.88}$ & $46.90{\scriptstyle \pm 0.88}$ & $47.22{\scriptstyle \pm 0.88}$ & $47.34{\scriptstyle \pm 0.88}$ 
          \\
          \cmidrule{2-13}
         & \multirow{2}{*}{Conv6} 
         & MAML & $20.29{\scriptstyle \pm 0.10}$ & $24.73{\scriptstyle \pm 0.42}$ & $58.56{\scriptstyle \pm 0.79}$ & $65.33{\scriptstyle \pm 0.82}$ & $66.78{\scriptstyle \pm 0.79}$ 
         & $43.29{\scriptstyle \pm 0.91}$ & $48.13{\scriptstyle \pm 0.92}$ & $49.59{\scriptstyle \pm 0.95}$ & $50.20{\scriptstyle \pm 0.94}$ & $50.40{\scriptstyle \pm 0.96}$ 
         \\
         & & ours & $60.27{\scriptstyle \pm 0.78}$ & $65.68{\scriptstyle \pm 0.79}$ & $66.76{\scriptstyle \pm 0.79}$ & $67.24{\scriptstyle \pm 0.79}$ & $67.60{\scriptstyle \pm 0.79}$ 
         & $46.10{\scriptstyle \pm 0.88}$ & $49.59{\scriptstyle \pm 0.93}$ & $50.44{\scriptstyle \pm 0.95}$ & $50.72{\scriptstyle \pm 0.95}$ & $50.87{\scriptstyle \pm 0.95}$ 
         \\
          \cmidrule{2-13}
        & \multirow{2}{*}{ResNet10} 
         & MAML & $20.01{\scriptstyle \pm 0.01}$ & $20.27{\scriptstyle \pm 0.10}$ & $36.18{\scriptstyle \pm 0.76}$ & $67.61{\scriptstyle \pm 0.90}$ & $73.03{\scriptstyle \pm 0.85}$ 
         & $20.00{\scriptstyle \pm 0.00}$ & $24.68{\scriptstyle \pm 0.50}$ & $52.25{\scriptstyle \pm 0.92}$ & $57.34{\scriptstyle \pm 0.94}$ & $57.80{\scriptstyle \pm 0.94}$ 
         \\
          & & ours & $58.15{\scriptstyle \pm 0.83}$ & $71.83{\scriptstyle \pm 0.75}$ & $74.40{\scriptstyle \pm 0.74}$ & $75.40{\scriptstyle \pm 0.72}$ & $75.77{\scriptstyle \pm 0.71}$ 
         & $52.88{\scriptstyle \pm 0.97}$ & $58.09{\scriptstyle \pm 0.96}$ & $59.19{\scriptstyle \pm 0.96}$ & $59.54{\scriptstyle \pm 0.96}$ & $59.65{\scriptstyle \pm 0.96}$ 
         \\
         \cmidrule{2-13}
        & \multirow{2}{*}{ResNet18} 
          & MAML & $20.00{\scriptstyle \pm 0.01}$ & $20.10{\scriptstyle \pm 0.05}$ & $38.93{\scriptstyle \pm 0.79}$ & $68.57{\scriptstyle \pm 0.88}$ & $73.90{\scriptstyle \pm 0.79}$ 
         & $20.02{\scriptstyle \pm 0.02}$ & $21.19{\scriptstyle \pm 0.23}$ & $51.30{\scriptstyle \pm 0.95}$ & $56.80{\scriptstyle \pm 1.00}$ & $57.71{\scriptstyle \pm 1.00}$ 
         \\
          & & ours & $70.13{\scriptstyle \pm 0.78}$ & $73.60{\scriptstyle \pm 0.74}$ & $74.31{\scriptstyle \pm 0.74}$ & $74.55{\scriptstyle \pm 0.74}$ & $74.67{\scriptstyle \pm 0.74}$ 
         & $54.79{\scriptstyle \pm 0.96}$ & $57.44{\scriptstyle \pm 0.97}$ & $57.64{\scriptstyle \pm 0.97}$ & $57.71{\scriptstyle \pm 0.98}$ & $57.80{\scriptstyle \pm 0.98}$ 
         \\
          \midrule \midrule
          \multirow{8}{*}{\STAB{\rotatebox[origin=c]{90}{CIFAR-FS}}} 
         & \multirow{2}{*}{Conv4}
         & MAML & $20.00{\scriptstyle \pm 0.01}$ & $20.00{\scriptstyle \pm 0.00}$ & $56.12{\scriptstyle \pm 0.85}$ & $67.08{\scriptstyle \pm 0.81}$ & $69.97{\scriptstyle \pm 0.75}$ 
         & $34.71{\scriptstyle \pm 0.75}$ & $42.34{\scriptstyle \pm 0.91}$ & $48.57{\scriptstyle \pm 0.96}$ & $51.10{\scriptstyle \pm 0.95}$ & $51.84{\scriptstyle \pm 0.93}$ 
         \\
         & & ours & $60.77{\scriptstyle \pm 0.82}$ & $67.55{\scriptstyle \pm 0.79}$ & $69.04{\scriptstyle \pm 0.77}$ & $69.90{\scriptstyle \pm 0.77}$ & $70.32{\scriptstyle \pm 0.76}$ 
         & $46.59{\scriptstyle \pm 0.92}$ & $50.84{\scriptstyle \pm 0.98}$ & $51.59{\scriptstyle \pm 0.98}$ & $51.92{\scriptstyle \pm 0.98}$ & $52.01{\scriptstyle \pm 0.99}$ 
         \\
          \cmidrule{2-13}
         & \multirow{2}{*}{Conv6} 
         & MAML & $20.22{\scriptstyle \pm 0.08}$ & $24.66{\scriptstyle \pm 0.40}$ & $65.34{\scriptstyle \pm 0.85}$ & $72.33{\scriptstyle \pm 0.82}$ & $74.00{\scriptstyle \pm 0.79}$ 
         & $24.83{\scriptstyle \pm 0.47}$ & $43.05{\scriptstyle \pm 0.82}$ & $54.90{\scriptstyle \pm 0.95}$ & $57.11{\scriptstyle \pm 0.95}$ & $57.83{\scriptstyle \pm 0.95}$ 
         \\
          & & ours & $67.12{\scriptstyle \pm 0.79}$ & $72.63{\scriptstyle \pm 0.79}$ & $73.75{\scriptstyle \pm 0.77}$ & $74.24{\scriptstyle \pm 0.77}$ & $74.47{\scriptstyle \pm 0.78}$ 
         & $54.07{\scriptstyle \pm 0.92}$ & $57.11{\scriptstyle \pm 0.94}$ & $57.68{\scriptstyle \pm 0.96}$ & $57.92{\scriptstyle \pm 0.96}$ & $58.12{\scriptstyle \pm 0.96}$ 
         \\
          \cmidrule{2-13}
        & \multirow{2}{*}{ResNet10} 
         & MAML & $20.04{\scriptstyle \pm 0.03}$ & $20.00{\scriptstyle \pm 0.00}$ & $40.23{\scriptstyle \pm 0.87}$ & $71.29{\scriptstyle \pm 0.87}$ & $77.12{\scriptstyle \pm 0.77}$ 
         & $20.00{\scriptstyle \pm 0.00}$ & $57.47{\scriptstyle \pm 0.97}$ & $63.57{\scriptstyle \pm 1.01}$ & $64.93{\scriptstyle \pm 1.00}$ & $65.42{\scriptstyle \pm 0.98}$ 
         \\
         & & ours & $60.93{\scriptstyle \pm 0.87}$ & $75.65{\scriptstyle \pm 0.76}$ & $77.67{\scriptstyle \pm 0.72}$ & $78.45{\scriptstyle \pm 0.72}$ & $78.90{\scriptstyle \pm 0.70}$ 
         & $57.58{\scriptstyle \pm 0.96}$ & $63.97{\scriptstyle \pm 0.94}$ & $65.63{\scriptstyle \pm 0.94}$ & $66.07{\scriptstyle \pm 0.94}$ & $66.17{\scriptstyle \pm 0.95}$ 
         \\
         \cmidrule{2-13}
        & \multirow{2}{*}{ResNet18} 
          & MAML & $20.01{\scriptstyle \pm 0.01}$ & $20.04{\scriptstyle \pm 0.03}$ & $45.75{\scriptstyle \pm 0.83}$ & $75.69{\scriptstyle \pm 0.81}$ & $79.59{\scriptstyle \pm 0.70}$ 
         & $20.00{\scriptstyle \pm 0.00}$ & $23.41{\scriptstyle \pm 0.39}$ & $63.34{\scriptstyle \pm 0.99}$ & $67.05{\scriptstyle \pm 0.97}$ & $67.44{\scriptstyle \pm 0.99}$ 
         \\
          & & ours & $77.31{\scriptstyle \pm 0.71}$ & $79.49{\scriptstyle \pm 0.67}$ & $79.96{\scriptstyle \pm 0.66}$ & $80.09{\scriptstyle \pm 0.65}$ & $80.18{\scriptstyle \pm 0.65}$ 
         & $65.73{\scriptstyle \pm 1.03}$ & $68.39{\scriptstyle \pm 1.04}$ & $68.75{\scriptstyle \pm 1.04}$ & $69.04{\scriptstyle \pm 1.03}$ & $69.09{\scriptstyle \pm 1.03}$ 
         \\
          \midrule \midrule
          \multirow{8}{*}{\STAB{\rotatebox[origin=c]{90}{FC100}}} 
         & \multirow{2}{*}{Conv4}
          & MAML & $20.00{\scriptstyle \pm 0.00}$ & $20.00{\scriptstyle \pm 0.00}$ & $38.03{\scriptstyle \pm 0.66}$ & $42.96{\scriptstyle \pm 0.74}$ & $47.61{\scriptstyle \pm 0.72}$ 
         & $20.04{\scriptstyle \pm 0.03}$ & $27.98{\scriptstyle \pm 0.57}$ & $31.12{\scriptstyle \pm 0.68}$ & $35.17{\scriptstyle \pm 0.76}$ & $36.08{\scriptstyle \pm 0.76}$ 
         \\
          & & ours & $40.46{\scriptstyle \pm 0.68}$ & $44.43{\scriptstyle \pm 0.72}$ & $46.90{\scriptstyle \pm 0.72}$ & $47.51{\scriptstyle \pm 0.73}$ & $48.01{\scriptstyle \pm 0.73}$ 
         & $33.84{\scriptstyle \pm 0.72}$ & $36.16{\scriptstyle \pm 0.75}$ & $36.68{\scriptstyle \pm 0.76}$ & $36.77{\scriptstyle \pm 0.76}$ & $36.82{\scriptstyle \pm 0.76}$ 
         \\
          \cmidrule{2-13}
          & \multirow{2}{*}{Conv6}
         & MAML & $20.00{\scriptstyle \pm 0.00}$ & $20.70{\scriptstyle \pm 0.14}$ & $39.62{\scriptstyle \pm 0.59}$ & $42.72{\scriptstyle \pm 0.71}$ & $46.48{\scriptstyle \pm 0.70}$ 
         & $30.31{\scriptstyle \pm 0.66}$ & $32.49{\scriptstyle \pm 0.69}$ & $34.67{\scriptstyle \pm 0.74}$ & $35.23{\scriptstyle \pm 0.76}$ & $35.64{\scriptstyle \pm 0.76}$ 
         \\
          & & ours & $41.16{\scriptstyle \pm 0.62}$ & $45.25{\scriptstyle \pm 0.68}$ & $46.57{\scriptstyle \pm 0.70}$ & $47.13{\scriptstyle \pm 0.71}$ & $47.45{\scriptstyle \pm 0.72}$ 
         & $33.55{\scriptstyle \pm 0.64}$ & $35.55{\scriptstyle \pm 0.65}$ & $36.15{\scriptstyle \pm 0.67}$ & $36.36{\scriptstyle \pm 0.66}$ & $36.40{\scriptstyle \pm 0.67}$ 
         \\
          \cmidrule{2-13}
        & \multirow{2}{*}{ResNet10} 
         & MAML & $20.00{\scriptstyle \pm 0.00}$ & $20.00{\scriptstyle \pm 0.01}$ & $30.54{\scriptstyle \pm 0.58}$ & $42.68{\scriptstyle \pm 0.71}$ & $47.03{\scriptstyle \pm 0.73}$ 
         & $20.00{\scriptstyle \pm 0.00}$ & $20.74{\scriptstyle \pm 0.16}$ & $33.19{\scriptstyle \pm 0.66}$ & $34.99{\scriptstyle \pm 0.73}$ & $36.33{\scriptstyle \pm 0.73}$ 
         \\
          & & ours & $42.66{\scriptstyle \pm 0.68}$ & $48.58{\scriptstyle \pm 0.70}$ & $49.94{\scriptstyle \pm 0.69}$ & $50.65{\scriptstyle \pm 0.69}$ & $50.86{\scriptstyle \pm 0.70}$ 
         & $32.17{\scriptstyle \pm 0.63}$ & $35.63{\scriptstyle \pm 0.71}$ & $36.78{\scriptstyle \pm 0.72}$ & $37.31{\scriptstyle \pm 0.73}$ & $37.50{\scriptstyle \pm 0.73}$ 
         \\
         \cmidrule{2-13}
        & \multirow{2}{*}{ResNet18} 
         & MAML & $20.00{\scriptstyle \pm 0.00}$ & $20.00{\scriptstyle \pm 0.00}$ & $20.00{\scriptstyle \pm 0.00}$ & $48.46{\scriptstyle \pm 0.73}$ & $48.56{\scriptstyle \pm 0.76}$ 
         & $20.00{\scriptstyle \pm 0.00}$ & $20.04{\scriptstyle \pm 0.02}$ & $32.30{\scriptstyle \pm 0.64}$ & $34.37{\scriptstyle \pm 0.69}$ & $35.19{\scriptstyle \pm 0.70}$ 
         \\
         & & ours & $47.11{\scriptstyle \pm 0.73}$ & $50.16{\scriptstyle \pm 0.73}$ & $50.82{\scriptstyle \pm 0.74}$ & $51.08{\scriptstyle \pm 0.74}$ & $51.22{\scriptstyle \pm 0.74}$ 
         & $33.24{\scriptstyle \pm 0.69}$ & $36.05{\scriptstyle \pm 0.71}$ & $36.70{\scriptstyle \pm 0.72}$ & $36.90{\scriptstyle \pm 0.72}$ & $37.02{\scriptstyle \pm 0.72}$ 
         \\
        \bottomrule
    \end{tabular}
    }
    }
    \end{center}
    \vskip -0.1in
    \label{tab:depths_sup}
\end{table*}

\section{Adaptation beyond the training horizon}
In this section, we provide further details regarding the behaviour of the baseline trained \textit{without} (`MAML') and \textit{with} the proposed condition loss~$\nicecal{L}_\kappa$ (`ours') when the models are provided with the possibility to perform an increased number of adaptation steps beyond the training horizon \textbf{at test time only} -- in our evaluated case up to 100 update steps. The training was in contrast performed with five adaptation steps. Results obtained on the test datasets for 5-way 5-shot and 5-way 1-shot scenarios are presented in~\cref{tab:adapt_beyond_all_sup} and follow the trends that have been discussed in the main paper.
\begin{table*}[h]
    \begin{center}
    \caption{\textbf{Adaptation beyond the training horizon.} Evaluations are conducted on all five popular FSL datasets: CUB-200-2011~\cite{WahCUB_200_2011}, \textit{mini}ImageNet~\cite{vinyals2016_matchingnet}, \textit{tiered}ImageNet~\cite{ren2018_metasemisup}, CIFAR-FS~\cite{bertinetto2019_cifarfsl} and FC100~\cite{oreshkin2018_tadam}. Models have been trained with 5 inner-loop update steps, but are evaluated using additional update steps at inference time.}
    \scalebox{0.60}
    {
    \hspace{-3.39mm}
    {
    \setlength{\tabcolsep}{3pt}
    \setlength{\tabcolsep}{6pt}
    \begin{tabular}{@{}c ll| *5c | *5c@{}} 
    \specialrule{.2em}{.1em}{.1em}
          & & & \multicolumn{5}{c}{\textbf{5-shot}} &  \multicolumn{5}{c}{\textbf{1-shot}}\\
          & \textbf{Network}  & \textbf{Method} & \textbf{step 5$\uparrow$} & \textbf{step 10$\uparrow$}  & \textbf{step 25$\uparrow$} & \textbf{step 50$\uparrow$} & \textbf{step 100$\uparrow$}  &  \textbf{step 5$\uparrow$} & \textbf{step 10$\uparrow$}  & \textbf{step 25$\uparrow$} & \textbf{step 50$\uparrow$} &\textbf{step 100$\uparrow$} \\    
          \midrule 
         
          \multirow{8}{*}{\STAB{\rotatebox[origin=c]{90}{CUB}}} 
         & \multirow{2}{*}{Conv4} 
          & MAML & $77.06{\scriptstyle \pm 0.69}$ & $77.84{\scriptstyle \pm 0.66}$ & $77.87{\scriptstyle \pm 0.65}$ & $77.78{\scriptstyle \pm 0.66}$ & $77.83{\scriptstyle \pm 0.65}$ 
          & $61.09{\scriptstyle \pm 0.98}$ & $61.33{\scriptstyle \pm 0.99}$ & $61.50{\scriptstyle \pm 0.99}$ & $61.61{\scriptstyle \pm 0.98}$ & $61.65{\scriptstyle \pm 0.98}$ 
          \\
          & & ours & $77.24{\scriptstyle \pm 0.67}$ & $77.48{\scriptstyle \pm 0.68}$ & $77.74{\scriptstyle \pm 0.67}$ & $77.88{\scriptstyle \pm 0.68}$ & $78.06{\scriptstyle \pm 0.68}$ 
          & $63.26{\scriptstyle \pm 0.98}$ & $63.42{\scriptstyle \pm 0.99}$ & $63.44{\scriptstyle \pm 1.00}$ & $63.54{\scriptstyle \pm 1.00}$ & $63.47{\scriptstyle \pm 1.00}$ 
          \\
          \cmidrule{2-13}
         & \multirow{2}{*}{Conv6} 
          & MAML & $80.52{\scriptstyle \pm 0.65}$ & $80.89{\scriptstyle \pm 0.63}$ & $81.00{\scriptstyle \pm 0.63}$ & $81.00{\scriptstyle \pm 0.63}$ & $81.01{\scriptstyle \pm 0.64}$ 
          & $67.67{\scriptstyle \pm 0.98}$ & $68.03{\scriptstyle \pm 0.98}$ & $68.39{\scriptstyle \pm 0.98}$ & $68.48{\scriptstyle \pm 0.98}$ & $68.56{\scriptstyle \pm 0.98}$ 
          \\
          & & ours & $80.65{\scriptstyle \pm 0.63}$ & $80.90{\scriptstyle \pm 0.63}$ & $81.15{\scriptstyle \pm 0.62}$ & $81.25{\scriptstyle \pm 0.62}$ & $81.35{\scriptstyle \pm 0.62}$ 
          & $68.87{\scriptstyle \pm 1.06}$ & $69.12{\scriptstyle \pm 1.06}$ & $69.19{\scriptstyle \pm 1.06}$ & $69.16{\scriptstyle \pm 1.06}$ & $69.15{\scriptstyle \pm 1.06}$ 
          \\
          \cmidrule{2-13}
        & \multirow{2}{*}{ResNet10} 
         & MAML & $82.13{\scriptstyle \pm 0.64}$ & $83.90{\scriptstyle \pm 0.59}$ & $83.99{\scriptstyle \pm 0.59}$ & $84.09{\scriptstyle \pm 0.59}$ & $84.08{\scriptstyle \pm 0.59}$ 
         & $73.04{\scriptstyle \pm 0.96}$ & $73.56{\scriptstyle \pm 0.96}$ & $73.80{\scriptstyle \pm 0.96}$ & $73.97{\scriptstyle \pm 0.96}$ & $74.01{\scriptstyle \pm 0.96}$ 
         \\
          & & ours & $83.82{\scriptstyle \pm 0.59}$ & $84.09{\scriptstyle \pm 0.58}$ & $84.11{\scriptstyle \pm 0.59}$ & $84.26{\scriptstyle \pm 0.57}$ & $84.28{\scriptstyle \pm 0.57}$ 
          & $74.99{\scriptstyle \pm 0.96}$ & $74.99{\scriptstyle \pm 0.97}$ & $75.19{\scriptstyle \pm 0.97}$ & $75.24{\scriptstyle \pm 0.97}$ & $75.33{\scriptstyle \pm 0.97}$ 
          \\
         \cmidrule{2-13}
        & \multirow{2}{*}{ResNet18} 
          & MAML & $83.56{\scriptstyle \pm 0.61}$ & $84.52{\scriptstyle \pm 0.55}$ & $84.56{\scriptstyle \pm 0.55}$ & $84.65{\scriptstyle \pm 0.54}$ & $84.59{\scriptstyle \pm 0.56}$ 
          & $74.52{\scriptstyle \pm 0.94}$ & $74.97{\scriptstyle \pm 0.95}$ & $75.23{\scriptstyle \pm 0.95}$ & $75.30{\scriptstyle \pm 0.95}$ & $75.34{\scriptstyle \pm 0.94}$ 
          \\
          & & ours & $84.66{\scriptstyle \pm 0.54}$ & $84.83{\scriptstyle \pm 0.53}$ & $84.94{\scriptstyle \pm 0.53}$ & $85.01{\scriptstyle \pm 0.53}$ & $85.01{\scriptstyle \pm 0.53}$ 
          & $75.22{\scriptstyle \pm 0.97}$ & $75.52{\scriptstyle \pm 0.95}$ & $75.66{\scriptstyle \pm 0.95}$ & $75.74{\scriptstyle \pm 0.95}$ & $75.81{\scriptstyle \pm 0.95}$ 
          \\
          \midrule \midrule
          \multirow{8}{*}{\STAB{\rotatebox[origin=c]{90}{\textit{mini}-ImageNet}}} 
         & \multirow{2}{*}{Conv4} 
          & MAML & $64.50{\scriptstyle \pm 0.69}$ & $65.35{\scriptstyle \pm 0.70}$ & $65.71{\scriptstyle \pm 0.70}$ & $65.90{\scriptstyle \pm 0.70}$ & $66.07{\scriptstyle \pm 0.70}$ 
          & $48.15{\scriptstyle \pm 0.80}$ & $48.52{\scriptstyle \pm 0.80}$ & $48.75{\scriptstyle \pm 0.79}$ & $48.86{\scriptstyle \pm 0.79}$ & $48.90{\scriptstyle \pm 0.79}$ 
          \\
          & & ours & $65.26{\scriptstyle \pm 0.67}$ & $65.84{\scriptstyle \pm 0.67}$ & $66.26{\scriptstyle \pm 0.67}$ & $66.46{\scriptstyle \pm 0.67}$ & $66.58{\scriptstyle \pm 0.66}$ 
          & $48.94{\scriptstyle \pm 0.80}$ & $49.16{\scriptstyle \pm 0.80}$ & $49.31{\scriptstyle \pm 0.81}$ & $49.45{\scriptstyle \pm 0.80}$ & $49.53{\scriptstyle \pm 0.80}$ 
          \\
          \cmidrule{2-13}
         & \multirow{2}{*}{Conv6} 
          & MAML & $65.96{\scriptstyle \pm 0.71}$ & $66.55{\scriptstyle \pm 0.71}$ & $66.63{\scriptstyle \pm 0.72}$ & $66.72{\scriptstyle \pm 0.71}$ & $66.80{\scriptstyle \pm 0.71}$ 
          & $51.22{\scriptstyle \pm 0.88}$ & $51.42{\scriptstyle \pm 0.87}$ & $51.48{\scriptstyle \pm 0.87}$ & $51.52{\scriptstyle \pm 0.87}$ & $51.53{\scriptstyle \pm 0.87}$ 
          \\
          & & ours & $68.43{\scriptstyle \pm 0.71}$ & $68.84{\scriptstyle \pm 0.71}$ & $69.11{\scriptstyle \pm 0.70}$ & $69.28{\scriptstyle \pm 0.70}$ & $69.39{\scriptstyle \pm 0.70}$ 
          & $53.34{\scriptstyle \pm 0.89}$ & $53.42{\scriptstyle \pm 0.89}$ & $53.51{\scriptstyle \pm 0.89}$ & $53.55{\scriptstyle \pm 0.89}$ & $53.55{\scriptstyle \pm 0.89}$ 
          \\
          \cmidrule{2-13}
        & \multirow{2}{*}{ResNet10} 
          & MAML & $69.43{\scriptstyle \pm 0.71}$ & $71.90{\scriptstyle \pm 0.68}$ & $72.29{\scriptstyle \pm 0.66}$ & $72.28{\scriptstyle \pm 0.67}$ & $72.21{\scriptstyle \pm 0.67}$ 
          & $57.35{\scriptstyle \pm 0.80}$ & $57.80{\scriptstyle \pm 0.83}$ & $57.86{\scriptstyle \pm 0.84}$ & $57.91{\scriptstyle \pm 0.84}$ & $58.17{\scriptstyle \pm 0.85}$ 
         \\
          & & ours & $72.46{\scriptstyle \pm 0.71}$ & $73.10{\scriptstyle \pm 0.67}$ & $73.33{\scriptstyle \pm 0.68}$ & $73.28{\scriptstyle \pm 0.68}$ & $73.35{\scriptstyle \pm 0.68}$ 
          & $58.20{\scriptstyle \pm 0.85}$ & $58.28{\scriptstyle \pm 0.87}$ & $58.27{\scriptstyle \pm 0.87}$ & $58.29{\scriptstyle \pm 0.87}$ & $58.38{\scriptstyle \pm 0.88}$ 
          \\
         \cmidrule{2-13}
        & \multirow{2}{*}{ResNet18} 
          & MAML & $71.06{\scriptstyle \pm 0.74}$ & $73.37{\scriptstyle \pm 0.68}$ & $73.57{\scriptstyle \pm 0.69}$ & $73.56{\scriptstyle \pm 0.69}$ & $73.53{\scriptstyle \pm 0.69}$ 
          & $56.84{\scriptstyle \pm 0.90}$ & $57.11{\scriptstyle \pm 0.91}$ & $56.97{\scriptstyle \pm 0.91}$ & $56.96{\scriptstyle \pm 0.91}$ & $57.00{\scriptstyle \pm 0.91}$ 
          \\
          & & ours & $73.28{\scriptstyle \pm 0.68}$ & $73.46{\scriptstyle \pm 0.68}$ & $73.57{\scriptstyle \pm 0.68}$ & $73.57{\scriptstyle \pm 0.68}$ & $73.62{\scriptstyle \pm 0.68}$ 
          & $57.64{\scriptstyle \pm 0.91}$ & $57.82{\scriptstyle \pm 0.90}$ & $57.92{\scriptstyle \pm 0.91}$ & $58.02{\scriptstyle \pm 0.91}$ & $58.05{\scriptstyle \pm 0.91}$ 
          \\
          \midrule \midrule
        \multirow{8}{*}{\STAB{\rotatebox[origin=c]{90}{\textit{tiered}ImageNet}}} 
         & \multirow{2}{*}{Conv4}
          & MAML & $63.26{\scriptstyle \pm 0.77}$ & $63.87{\scriptstyle \pm 0.77}$ & $64.26{\scriptstyle \pm 0.77}$ & $64.52{\scriptstyle \pm 0.76}$ & $64.65{\scriptstyle \pm 0.76}$ 
          & $47.33{\scriptstyle \pm 0.90}$ & $47.74{\scriptstyle \pm 0.90}$ & $47.93{\scriptstyle \pm 0.91}$ & $48.04{\scriptstyle \pm 0.91}$ & $48.08{\scriptstyle \pm 0.91}$ 
          \\
          & & ours & $64.77{\scriptstyle \pm 0.75}$ & $65.56{\scriptstyle \pm 0.74}$ & $65.91{\scriptstyle \pm 0.74}$ & $66.01{\scriptstyle \pm 0.74}$ & $66.05{\scriptstyle \pm 0.74}$ 
          & $47.34{\scriptstyle \pm 0.88}$ & $47.74{\scriptstyle \pm 0.90}$ & $48.07{\scriptstyle \pm 0.90}$ & $48.21{\scriptstyle \pm 0.90}$ & $48.29{\scriptstyle \pm 0.90}$ 
          \\
          \cmidrule{2-13}
         & \multirow{2}{*}{Conv6} 
         & MAML & $66.78{\scriptstyle \pm 0.79}$ & $67.26{\scriptstyle \pm 0.79}$ & $67.40{\scriptstyle \pm 0.79}$ & $67.42{\scriptstyle \pm 0.79}$ & $67.51{\scriptstyle \pm 0.79}$ 
         & $50.40{\scriptstyle \pm 0.96}$ & $50.74{\scriptstyle \pm 0.95}$ & $50.96{\scriptstyle \pm 0.95}$ & $51.00{\scriptstyle \pm 0.95}$ & $51.07{\scriptstyle \pm 0.96}$ 
         \\
          & & ours & $67.60{\scriptstyle \pm 0.79}$ & $68.22{\scriptstyle \pm 0.78}$ & $68.56{\scriptstyle \pm 0.78}$ & $68.77{\scriptstyle \pm 0.78}$ & $68.92{\scriptstyle \pm 0.78}$ 
         & $50.87{\scriptstyle \pm 0.95}$ & $51.18{\scriptstyle \pm 0.95}$ & $51.42{\scriptstyle \pm 0.96}$ & $51.50{\scriptstyle \pm 0.96}$ & $51.62{\scriptstyle \pm 0.97}$ 
         \\
          \cmidrule{2-13}
        & \multirow{2}{*}{ResNet10} 
         & MAML & $73.03{\scriptstyle \pm 0.85}$ & $75.14{\scriptstyle \pm 0.79}$ & $75.22{\scriptstyle \pm 0.79}$ & $75.19{\scriptstyle \pm 0.79}$ & $75.20{\scriptstyle \pm 0.80}$ 
         & $57.80{\scriptstyle \pm 0.94}$ & $57.95{\scriptstyle \pm 0.96}$ & $58.07{\scriptstyle \pm 0.95}$ & $58.16{\scriptstyle \pm 0.96}$ & $58.22{\scriptstyle \pm 0.96}$ 
         \\ 
         & & ours & $75.77{\scriptstyle \pm 0.71}$ & $76.29{\scriptstyle \pm 0.70}$ & $76.45{\scriptstyle \pm 0.70}$ & $74.42{\scriptstyle \pm 0.70}$ & $76.45{\scriptstyle \pm 0.70}$ 
         & $59.65{\scriptstyle \pm 0.96}$ & $59.94{\scriptstyle \pm 0.95}$ & $60.22{\scriptstyle \pm 0.94}$ & $60.42{\scriptstyle \pm 0.93}$ & $60.53{\scriptstyle \pm 0.93}$ 
         \\
         \cmidrule{2-13}
        & \multirow{2}{*}{ResNet18} 
          & MAML & $73.90{\scriptstyle \pm 0.79}$ & $74.88{\scriptstyle \pm 0.76}$ & $74.98{\scriptstyle \pm 0.77}$ & $74.89{\scriptstyle \pm 0.77}$ & $74.94{\scriptstyle \pm 0.77}$ 
         & $57.71{\scriptstyle \pm 1.00}$ & $57.88{\scriptstyle \pm 1.00}$ & $57.90{\scriptstyle \pm 1.01}$ & $57.92{\scriptstyle \pm 1.01}$ & $57.99{\scriptstyle \pm 1.01}$  
         \\
          & & ours & $74.67{\scriptstyle \pm 0.74}$ & $74.86{\scriptstyle \pm 0.75}$ & $74.96{\scriptstyle \pm 0.74}$ & $75.02{\scriptstyle \pm 0.74}$ & $75.01{\scriptstyle \pm 0.74}$ 
         & $57.80{\scriptstyle \pm 0.98}$ & $57.97{\scriptstyle \pm 0.98}$ & $58.09{\scriptstyle \pm 0.99}$ & $58.14{\scriptstyle \pm 0.99}$ & $58.10{\scriptstyle \pm 0.98}$ 
         \\
          \midrule \midrule
          \multirow{8}{*}{\STAB{\rotatebox[origin=c]{90}{CIFAR-FS}}} 
         & \multirow{2}{*}{Conv4}
         & MAML & $69.97{\scriptstyle \pm 0.75}$ & $70.54{\scriptstyle \pm 0.76}$ & $71.04{\scriptstyle \pm 0.76}$ & $71.20{\scriptstyle \pm 0.76}$ & $71.32{\scriptstyle \pm 0.76}$ 
         & $51.84{\scriptstyle \pm 0.93}$ & $52.31{\scriptstyle \pm 0.93}$ & $52.52{\scriptstyle \pm 0.94}$ & $52.74{\scriptstyle \pm 0.94}$ & $52.89{\scriptstyle \pm 0.94}$ 
         \\
         & & ours & $70.32{\scriptstyle \pm 0.76}$ & $71.08{\scriptstyle \pm 0.77}$ & $71.52{\scriptstyle \pm 0.75}$ & $71.68{\scriptstyle \pm 0.75}$ & $71.86{\scriptstyle \pm 0.76}$ 
         & $52.01{\scriptstyle \pm 0.99}$ & $52.46{\scriptstyle \pm 0.98}$ & $52.68{\scriptstyle \pm 0.98}$ & $52.84{\scriptstyle \pm 0.99}$ & $53.02{\scriptstyle \pm 0.99}$ 
         \\
          \cmidrule{2-13}
         & \multirow{2}{*}{Conv6} 
          & MAML & $74.00{\scriptstyle \pm 0.79}$ & $74.56{\scriptstyle \pm 0.78}$ & $74.75{\scriptstyle \pm 0.77}$ & $74.78{\scriptstyle \pm 0.77}$ & $74.80{\scriptstyle \pm 0.77}$ 
         & $57.83{\scriptstyle \pm 0.95}$ & $58.10{\scriptstyle \pm 0.95}$ & $58.29{\scriptstyle \pm 0.96}$ & $58.38{\scriptstyle \pm 0.96}$ & $58.47{\scriptstyle \pm 0.96}$ 
         \\
          & & ours & $74.47{\scriptstyle \pm 0.78}$ & $74.99{\scriptstyle \pm 0.78}$ & $75.24{\scriptstyle \pm 0.78}$ & $75.38{\scriptstyle \pm 0.78}$ & $75.52{\scriptstyle \pm 0.77}$ 
         & $58.12{\scriptstyle \pm 0.96}$ & $58.43{\scriptstyle \pm 0.96}$ & $58.58{\scriptstyle \pm 0.97}$ & $58.70{\scriptstyle \pm 0.97}$ & $58.76{\scriptstyle \pm 0.97}$ 
         \\
          \cmidrule{2-13}
        & \multirow{2}{*}{ResNet10} 
         & MAML & $77.12{\scriptstyle \pm 0.77}$ & $78.75{\scriptstyle \pm 0.69}$ & $78.77{\scriptstyle \pm 0.69}$ & $78.72{\scriptstyle \pm 0.70}$ & $78.69{\scriptstyle \pm 0.71}$ 
         & $65.42{\scriptstyle \pm 0.98}$ & $65.82{\scriptstyle \pm 0.99}$ & $65.81{\scriptstyle \pm 0.99}$ & $65.98{\scriptstyle \pm 0.98}$ & $66.03{\scriptstyle \pm 0.97}$ 
         \\
          & & ours & $78.90{\scriptstyle \pm 0.70}$ & $79.30{\scriptstyle \pm 0.68}$ & $79.39{\scriptstyle \pm 0.68}$ & $79.34{\scriptstyle \pm 0.69}$ & $79.37{\scriptstyle \pm 0.69}$ 
         & $66.17{\scriptstyle \pm 0.95}$ & $66.04{\scriptstyle \pm 0.96}$ & $66.07{\scriptstyle \pm 0.97}$ & $66.16{\scriptstyle \pm 0.96}$ & $66.24{\scriptstyle \pm 0.96}$ 
         \\
         \cmidrule{2-13}
        & \multirow{2}{*}{ResNet18} 
          & MAML & $79.59{\scriptstyle \pm 0.70}$ & $80.63{\scriptstyle \pm 0.66}$ & $80.68{\scriptstyle \pm 0.65}$ & $80.60{\scriptstyle \pm 0.65}$ & $80.62{\scriptstyle \pm 0.65}$ 
         & $67.44{\scriptstyle \pm 0.99}$ & $67.73{\scriptstyle \pm 0.99}$ & $67.62{\scriptstyle \pm 1.01}$ & $67.67{\scriptstyle \pm 1.01}$ & $67.66{\scriptstyle \pm 1.02}$ 
         \\
         & & ours & $80.18{\scriptstyle \pm 0.65}$ & $80.37{\scriptstyle \pm 0.65}$ & $80.48{\scriptstyle \pm 0.64}$ & $80.60{\scriptstyle \pm 0.64}$ & $80.71{\scriptstyle \pm 0.64}$ 
         & $69.09{\scriptstyle \pm 1.03}$ & $69.35{\scriptstyle \pm 1.03}$ & $69.49{\scriptstyle \pm 1.03}$ & $69.52{\scriptstyle \pm 1.03}$ & $69.56{\scriptstyle \pm 1.03}$ 
         \\
          \midrule \midrule
          \multirow{8}{*}{\STAB{\rotatebox[origin=c]{90}{FC100}}} 
         & \multirow{2}{*}{Conv4}
          & MAML & $47.61{\scriptstyle \pm 0.72}$ & $48.40{\scriptstyle \pm 0.71}$ & $48.79{\scriptstyle \pm 0.73}$ & $49.14{\scriptstyle \pm 0.73}$ & $49.25{\scriptstyle \pm 0.73}$ 
         & $36.08{\scriptstyle \pm 0.76}$ & $36.45{\scriptstyle \pm 0.76}$ & $36.66{\scriptstyle \pm 0.77}$ & $36.82{\scriptstyle \pm 0.77}$ & $36.89{\scriptstyle \pm 0.76}$ 
         \\
          & & ours & $48.01{\scriptstyle \pm 0.73}$ & $48.59{\scriptstyle \pm 0.72}$ & $49.01{\scriptstyle \pm 0.72}$ & $49.30{\scriptstyle \pm 0.72}$ & $49.52{\scriptstyle \pm 0.72}$ 
         & $36.82{\scriptstyle \pm 0.76}$ & $37.08{\scriptstyle \pm 0.76}$ & $37.24{\scriptstyle \pm 0.76}$ & $37.31{\scriptstyle \pm 0.76}$ & $37.45{\scriptstyle \pm 0.75}$ 
         \\
          \cmidrule{2-13}
         & \multirow{2}{*}{Conv6} 
         & MAML & $46.48{\scriptstyle \pm 0.70}$ & $47.30{\scriptstyle \pm 0.70}$ & $47.48{\scriptstyle \pm 0.71}$ & $47.69{\scriptstyle \pm 0.71}$ & $47.82{\scriptstyle \pm 0.71}$ 
         & $35.64{\scriptstyle \pm 0.76}$ & $35.86{\scriptstyle \pm 0.76}$ & $35.99{\scriptstyle \pm 0.75}$ & $36.04{\scriptstyle \pm 0.76}$ & $36.05{\scriptstyle \pm 0.75}$ 
         \\
         & & ours & $47.45{\scriptstyle \pm 0.72}$ & $48.08{\scriptstyle \pm 0.73}$ & $48.56{\scriptstyle \pm 0.72}$ & $48.77{\scriptstyle \pm 0.72}$ & $48.90{\scriptstyle \pm 0.72}$ 
         & $36.40{\scriptstyle \pm 0.67}$ & $36.51{\scriptstyle \pm 0.67}$ & $36.72{\scriptstyle \pm 0.67}$ & $36.74{\scriptstyle \pm 0.67}$ & $36.80{\scriptstyle \pm 0.67}$ 
         \\
          \cmidrule{2-13}
        & \multirow{2}{*}{ResNet10} 
         & MAML & $47.03{\scriptstyle \pm 0.73}$ & $49.23{\scriptstyle \pm 0.71}$ & $49.11{\scriptstyle \pm 0.72}$ & $49.06{\scriptstyle \pm 0.72}$ & $49.10{\scriptstyle \pm 0.71}$ 
         & $36.33{\scriptstyle \pm 0.73}$ & $36.32{\scriptstyle \pm 0.73}$ & $36.47{\scriptstyle \pm 0.73}$ & $36.63{\scriptstyle \pm 0.74}$ & $36.76{\scriptstyle \pm 0.74}$ 
         \\
          & & ours & $50.86{\scriptstyle \pm 0.70}$ & $51.08{\scriptstyle \pm 0.70}$ & $51.15{\scriptstyle \pm 0.71}$ & $51.17{\scriptstyle \pm 0.70}$ & $51.16{\scriptstyle \pm 0.70}$ 
         & $37.50{\scriptstyle \pm 0.73}$ & $37.52{\scriptstyle \pm 0.74}$ & $37.36{\scriptstyle \pm 0.74}$ & $37.50{\scriptstyle \pm 0.74}$ & $37.55{\scriptstyle \pm 0.74}$ 
         \\
         \cmidrule{2-13}
        & \multirow{2}{*}{ResNet18} 
         & MAML & $48.56{\scriptstyle \pm 0.76}$ & $50.25{\scriptstyle \pm 0.76}$ & $50.29{\scriptstyle \pm 0.76}$ & $50.16{\scriptstyle \pm 0.75}$ & $50.18{\scriptstyle \pm 0.76}$ 
         & $35.19{\scriptstyle \pm 0.70}$ & $35.36{\scriptstyle \pm 0.72}$ & $35.44{\scriptstyle \pm 0.72}$ & $35.48{\scriptstyle \pm 0.72}$ & $35.57{\scriptstyle \pm 0.72}$ 
         \\
         & & ours & $51.22{\scriptstyle \pm 0.74}$ & $51.41{\scriptstyle \pm 0.74}$ & $51.55{\scriptstyle \pm 0.74}$ & $51.52{\scriptstyle \pm 0.74}$ & $51.63{\scriptstyle \pm 0.74}$ 
         & $37.02{\scriptstyle \pm 0.72}$ & $37.27{\scriptstyle \pm 0.72}$ & $37.26{\scriptstyle \pm 0.72}$ & $37.37{\scriptstyle \pm 0.72}$ & $37.40{\scriptstyle \pm 0.72}$ 
         \\
        \bottomrule
    \end{tabular}
    }
    }
    \label{tab:adapt_beyond_all_sup}
    \end{center}
    \vskip -0.1in
\end{table*}

\section{Ablating the proposed condition loss function}
We introduced in the main paper that computing the condition number as defined in Equation~(4) would ignore the distribution of all but two eigenvalues and thus unnecessarily weaken the training signal if directly used as conditioning objective. In this section, we back up this intuition with empirical insights. 
In detail, we contrast both versions 1) using our loss defined via the variance of the logarithmic eigenvalues of the approximated Hessian as proposed in the main paper in Equation~(9) to 2) simply using the logarithmic condition number computed via the maximum and minimum eigenvalues (\cref{tab:loss_ablation}).
We find that while using the logarithmic condition number does still lead to a significant improvement of adaptation performance especially during the first few steps when compared to its unconstrained counterpart (MAML), it is notably outperformed by our proposed loss using the variance of the eigenvalues. 
\begin{table*}[b]
\caption{\textbf{Ablating the condition loss function.} Reported are the step-wise classification accuracies on the validation set of \textit{mini}ImageNet~\cite{vinyals2016_matchingnet} for a 5-way 5-shot scenario (Conv6).}
\vspace{-8pt}
    \begin{center}
    \scalebox{0.83}
    {
    \hspace{-3.39mm}
    {
    \setlength{\tabcolsep}{6pt}
    \begin{tabular}{@{} l *5c @{}} 
    \specialrule{.2em}{.1em}{.1em}
          & \multicolumn{5}{c}{\textbf{5-shot}} \\
          \textbf{Loss~$\mathcal{L}_\kappa$} & \textbf{step 1$\uparrow$}  & \textbf{step 2$\uparrow$} & \textbf{step 3$\uparrow$}  & \textbf{step 4$\uparrow$} & \textbf{step 5$\uparrow$} \\    \midrule
          var(log(ev)) & $63.93{\scriptstyle \pm 1.76}$ & $68.44{\scriptstyle \pm 1.70}$ & $69.15{\scriptstyle \pm 1.70}$ & $69.78{\scriptstyle \pm 1.69}$ & $69.83{\scriptstyle \pm 1.73}$ \\
          log($\kappa$) & $55.30{\scriptstyle \pm 2.04}$ & $62.03{\scriptstyle \pm 1.87}$ & $63.95{\scriptstyle \pm 1.82}$ & $64.98{\scriptstyle \pm 1.87}$ & $65.36{\scriptstyle \pm 1.86}$      \\
        \bottomrule
    \end{tabular}
    }
    }
    \end{center}
    \label{tab:loss_ablation}
    \vskip -0.2in
\end{table*}


\section{Preconditioning -- Number of parameters and performance}
As discussed in the main paper, we compare our approach to other recently published methods that aim to achieve better convergence via preconditioning. \cref{tab:sup_precondupdt} outlines the different parameter update procedures and highlights the additionally introduced parameters of other methods (blue). Note that these parameters are required at both training and inference time, and lead to a significant increase in parameter count ranging from $96\%$ up to $2235\%$. In contrast, our proposed approach does not require any additional parameters to achieve preconditioning and thus allows to use more powerful backbones if increased parameter counts can be tolerated -- enabling our method to outperform others across the entire parameter-accuracy spectrum. While we show a visualization outlining the interplay between the total number of parameters and achieved accuracies within the main paper, we provide extended details regarding the explicit parameter counts and accuracy values in \cref{tab:sup_precondparams}.

\begin{table*}[h]
\caption{\textbf{Preconditioned parameter updates.} Detailed are the different ways of updating the parameters for recently published preconditioning methods. Additionally introduced parameters are highlighted in blue, and are required at both training and inference time (cf. \cref{tab:sup_precondparams}).}
\vspace{-8pt}
    \begin{center}
    \scalebox{0.9}
    {
    \hspace{-3.39mm}
    {
    \setlength{\tabcolsep}{6pt}
    \centering
        \begin{tabular}{l l}
        \specialrule{.2em}{.1em}{.1em}
        
        \textbf{Method}
        &\textbf{Inner Loop Param. Update}
        \\
        \toprule\toprule

        MAML~\cite{finn2017_maml}
        &$\boldsymbol{\theta}^{(k)}_{\tau} = \boldsymbol{\theta}^{(k-1)}_{\tau} - \alpha \nabla_{\boldsymbol{\theta}^{(k-1)}}\nicecal{L}(\boldsymbol{\theta}^{(k-1)}_{\tau})$
        \vspace{0.2em}\\
        
        Ours    
        &$\boldsymbol{\theta}^{(k)}_{\tau} = \boldsymbol{\theta}^{(k-1)}_{\tau} - \alpha \nabla_{\boldsymbol{\theta}^{(k-1)}}\nicecal{L}(\boldsymbol{\theta}^{(k-1)}_{\tau})$
        \vspace{0.2em}\\
        \midrule
        
        Meta-SGD~\cite{li2017_metasgd}
        &$\boldsymbol{\theta}^{(k)}_{\tau} = \boldsymbol{\theta}^{(k-1)}_{\tau} - \alpha \,\textcolor{blue}{\mathrm{diag}(\phi)} \nabla_{\boldsymbol{\theta}^{(k-1)}}\nicecal{L}(\boldsymbol{\theta}^{(k-1)}_{\tau})$
        \vspace{0.2em}\\
        
        MC~\cite{park2019_metacurvature}
        &$\boldsymbol{\theta}^{(k)}_{\tau} = \boldsymbol{\theta}^{(k-1)}_{\tau} - \alpha \,\textcolor{blue}{M(\boldsymbol{\theta}^{(k-1)}_{\tau},\psi)} \nabla_{\boldsymbol{\theta}^{(k-1)}}\nicecal{L}(\boldsymbol{\theta}^{(k-1)}_{\tau})$
        \vspace{0.2em}\\
        
        ModGrad~\cite{simon2020_gradientPrecon}
        &$\boldsymbol{\theta}^{(k)}_{\tau} = \boldsymbol{\theta}^{(k-1)}_{\tau} - \alpha \,\textcolor{blue}{M_{\tau}^{(k-1)}(\Psi)} \nabla_{\boldsymbol{\theta}^{(k-1)}}\nicecal{L}(\boldsymbol{\theta}^{(k-1)}_{\tau})$
        \vspace{0.2em}\\
        Warp-MAML~\cite{flennerhag2019_metawarpedgrad}
        &$\boldsymbol{\theta}^{(k)}_{\tau} = \boldsymbol{\theta}^{(k-1)}_{\tau} - \alpha \nabla_{\boldsymbol{\theta}^{(k-1)}}\nicecal{L}(\boldsymbol{\theta}^{(k-1)}_{\tau}, \textcolor{blue}{\zeta})$
        \\
        \end{tabular}
    }
    }
    \end{center}
    \label{tab:sup_precondupdt}
\end{table*}

\begin{table}[h]
    \caption{\textbf{Preconditioning methods, number of parameters and accuracies.} Obtained for $5$-way $5$-shot evaluated on the \textit{mini}ImageNet test set. Reported are results for MAML~\cite{finn2017_maml}, Meta-SGD~\cite{li2017_metasgd}, MC~\cite{park2019_metacurvature}, ModGrad~\cite{simon2020_gradientPrecon} and Warp-MAML~\cite{flennerhag2019_metawarpedgrad}. \textsuperscript{\textdagger}denotes reimplemented versions (cf. Table~1, main paper).}\label{tab:sup_precondparams}
\begin{center}
    \scalebox{0.9}
    {
    \hspace{-3.39mm}
    {
        \setlength{\tabcolsep}{6pt}
        \begin{tabular}{l l r c r r r}
        \specialrule{.2em}{.1em}{.1em}
         & \textbf{Backbone} & \textbf{Parameter}  & \textbf{Test} & \textbf{Rel. Acc.} & \textbf{\#Total} & \textbf{\#Backbone}\\
        \textbf{Method}& \textbf{Architecture} & \textbf{increase}\textdownarrow  & \textbf{Accuracy} & \textbf{increase\textuparrow} & \textbf{Parameters} & \textbf{Parameters}\\
        \toprule\toprule
        \rowcolor{Gray!20!White} MAML & Conv4 (32) & --\hphantom{$\boldsymbol{\,0\%}$}  & $63.11\scriptstyle\pm0.92$ & --\hphantom{$\boldsymbol{\;\,\%}$} & $32{,}901$ & $32{,}901$\\  
        ours & Conv4 (32) & $+$\hphantom{$\boldsymbol{10}$}$\boldsymbol{0\%}$  & $63.33\scriptstyle\pm0.72$ & \textcolor{Green}{$+\boldsymbol{0.3\%}$} & $32{,}901$ & $32{,}901$\\  
        Meta-SGD & Conv4 (32) & \textcolor{BrickRed}{$+\boldsymbol{100\%}$}  & $64.03\scriptstyle\pm0.94$ & \textcolor{Green}{$+\boldsymbol{1.5\%}$} & $65{,}802$ & $32{,}901$\\ 
        \textcolor{gray}{ours} & \textcolor{gray}{Conv6 (32)} & \textcolor{gray}{$+$\hphantom{$\boldsymbol{1}$}$\boldsymbol{57\%}$} & \textcolor{gray}{$64.47\scriptstyle\pm0.71$} & \textcolor{gray}{$+\boldsymbol{2.2\%}$} & \textcolor{gray}{$51{,}525$} & \textcolor{gray}{$51{,}525$} \\  
        \midrule
        \rowcolor{Gray!20!White}MAML\textsuperscript{\textdagger} & Conv4 (64) & --\hphantom{$\boldsymbol{\,0\%}$}   & $64.50\scriptstyle\pm0.69$ & --\hphantom{$\boldsymbol{\;\,\%}$} & $121{,}093$ & $121{,}093$\\  
        ours & Conv4 (64) & $+$\hphantom{$\boldsymbol{10}$}$\boldsymbol{0\%}$  & $65.26\scriptstyle\pm0.67$ & \textcolor{Green}{$+$\hphantom{$\boldsymbol{0}$}$\boldsymbol{1.2\%}$} & $121{,}093$ & $121{,}093$\\  
        ModGrad & Conv4 (64) & \textcolor{BrickRed}{$+\boldsymbol{873\%}$}  & $69.17\scriptstyle\pm0.69$ & \textcolor{Green}{$+$\hphantom{$\boldsymbol{0}$}$\boldsymbol{7.2\%}$} & $1{,}178{,}019$ & $121{,}093$\\ 
        \textcolor{gray}{ours} & \textcolor{gray}{Conv6 (64)} & \textcolor{gray}{$+$\hphantom{$\boldsymbol{0}$}$\boldsymbol{61\%}$}  & \textcolor{gray}{$68.43\scriptstyle\pm0.71$} & \textcolor{gray}{$+$\hphantom{$\boldsymbol{0}$}$\boldsymbol{6.1\%}$} & \textcolor{gray}{$195{,}205$} & \textcolor{gray}{$195{,}205$}\\ 
        \textcolor{gray}{ours} & \textcolor{gray}{Conv6 (128)} & \textcolor{gray}{$+\boldsymbol{527\%}$}  & \textcolor{gray}{$71.00\scriptstyle\pm0.68$} & \textcolor{gray}{$+\boldsymbol{10.1\%}$} & \textcolor{gray}{$759{,}045$} & \textcolor{gray}{$759{,}045$}\\ 
        \midrule
        \rowcolor{Gray!20!White}MAML\textsuperscript{\textdagger} & Conv4 (128) & --\hphantom{$\boldsymbol{\,0\%}$}  & $66.06\scriptstyle\pm0.71$ & --\hphantom{$\boldsymbol{\;\,\%}$} & $463{,}365$ & $463{,}365$\\  
        ours & Conv4 (128) & $+$\hphantom{$\boldsymbol{200}$}$\boldsymbol{0\%}$  & $68.07\scriptstyle\pm0.70$ & \textcolor{Green}{$+\boldsymbol{3.0\%}$} & $463{,}365$ & $463{,}365$\\  
        Warp-MAML & Conv4 (128) & \textcolor{BrickRed}{$+$\hphantom{$\boldsymbol{20}$}$\boldsymbol{96\%}$}  &  $68.4\:\;\scriptstyle\pm0.92$ & \textcolor{Green}{$+\boldsymbol{3.5\%}$} & $906{,}885$ & $463{,}365$ \\  
        MC & Conv4 (128) & \textcolor{BrickRed}{$+\boldsymbol{2235\%}$}   & $68.01\scriptstyle\pm0.73$ & \textcolor{Green}{$+\boldsymbol{3.0\%}$} & $10{,}818{,}928$ & $463{,}365$\\ 
        \textcolor{gray}{ours} & \textcolor{gray}{Conv6 (128)} & \textcolor{gray}{$+$\hphantom{$\boldsymbol{20}$}$\boldsymbol{64\%}$}  & \textcolor{gray}{$71.00\scriptstyle\pm0.68$} & \textcolor{gray}{$+\boldsymbol{7.5\%}$} & \textcolor{gray}{$759{,}045$} & \textcolor{gray}{$759{,}045$}\\ 
        \bottomrule
    \end{tabular}
    }
    }
    \end{center}
\end{table}

\section{Algorithm for better conditioned meta-learning}
\cref{alg:cond_alg} shows the concise form of how the conditioning loss presented in the main paper is used in the context of gradient-based few-shot meta-learning. The algorithm follows the concept introduced by~\citet{finn2017_maml} for MAML, with the addition of using our reformulated problem setting and in this way computing the condition information for each stage of the parameters updated during the inner loop (Lines~\ref{alg:line_J} and \ref{alg:line_EV}). The outer loop then incorporates the conditioning constraint (\cref{alg:line_Condloss}) as introduced in Equations~(9) and~(10) of the main paper and computes the overall task loss (\cref{alg:line_OverallTaskloss}). After completing all tasks in the current task batch, the network's parameters are then updated (\cref{alg:line_metaupdate}) by considering both the classification and condition objectives, encouraging the model to learn a well-conditioned parameter space while solving the classification challenge.

\begin{algorithm}

\caption{$\;\;$ Learning a Better Conditioned Parameter Space}\label{alg:train_cond}
\label{alg:cond_alg}
\begin{algorithmic}[1]
\Require $p(\nicecal{T}); \; \alpha, \beta, \gamma$  \Comment{Distribution over tasks; Hyperparameters}
\State $\boldsymbol{\theta}^*\leftarrow$ Random initialization
\While{not done}
    \LState $\{\tau_1,\dotso,\tau_B\} \sim p(\nicecal{T})$ \Comment{Sample a batch of tasks}
    \For{\textbf{all} $\tau_i$}
        \LState $\boldsymbol{\theta}^{0}_{\tau_i} \leftarrow \boldsymbol{\theta}^*$
        \LState $(\nicecal{D}^{\mathrm{train}}_{\tau_i} , \nicecal{D}^{\mathrm{val}}_{\tau_i}) \sim \tau_i$ \Comment{Sample train and validation set}
        \For{$k$ in $\{1,\dotso,K\}$ inner-loop update steps} \Comment{Inner-loop adaptation}
            \LState Compute $\mathbf{J}^{(k)}$ via $\nicecal{L}(\nicecal{D}^{\mathrm{train}}_{\tau_i},\boldsymbol{\theta}^{(k-1)}_{\tau_i})$ \Comment{Following Equations (5) - (8)} \label{alg:line_J}
            \LState Compute and temporarily store $\boldsymbol{\lambda}(\mathbf{J}^{(k)}\mathbf{J}^{(k)\top})$ \Comment{Eigenvalues of approx. Hessian}\label{alg:line_EV}
            \LState $\boldsymbol{\theta}^{(k)}_{\tau_i} \leftarrow \boldsymbol{\theta}^{(k-1)}_{\tau_i} - \alpha  \nabla_{\boldsymbol{\theta}^{(k-1)}_{\tau_i}}\nicecal{L}(\nicecal{D}^{\mathrm{train}}_{\tau_i},\boldsymbol{\theta}^{(k-1)}_{\tau_i})$ \Comment Inner-loop parameter update
        \EndFor
    \LState $\nicecal{L}_{\kappa}\left(\boldsymbol{\theta}^{(K)}_{\tau_i}\!\left(\nicecal{D}^{\mathrm{train}}_{\tau_i}, \boldsymbol{\theta}^{*}\right)\right) =  \frac{1}{K} \sum_{k=1}^{K}\mathrm{Var}\left(\log_{10}\left(\boldsymbol{\lambda}\left(\mathbf{J}^{(k)}\mathbf{J}^{(k)^{\top}}\right)\right)\right)$ \Comment{Cond. loss} \label{alg:line_Condloss}
    \LState $\nicecal{L}_{\tau_i} = \nicecal{L}\left(\nicecal{D}^{\mathrm{val}}_{\tau_i}, \;\; \boldsymbol{\theta}^{(K)}_{\tau_i}\!\left(\nicecal{D}^{\mathrm{train}}_{\tau_i}, \boldsymbol{\theta}^{*}\right)\right) \:\:  + \:\: \gamma \nicecal{L}_{\kappa}\left(\boldsymbol{\theta}^{(K)}_{\tau_i}\!\left(\nicecal{D}^{\mathrm{train}}_{\tau_i}, \boldsymbol{\theta}^{*}\right)\right)$ \Comment{Overall task loss} \label{alg:line_OverallTaskloss}
    \EndFor
    \LState $\boldsymbol{\theta}^{*} \leftarrow \boldsymbol{\theta}^{*} - \beta \nabla_{\boldsymbol{\theta}^{*}} \sum_{i=1}^{B}\nicecal{L}_{\tau_i}$ \Comment{Meta update overall parameter set} \label{alg:line_metaupdate}
\EndWhile%
\end{algorithmic}
\end{algorithm}

\section{Details on many-way multi-shot scenarios}
A detailed version of the results used for the visualization of different~5-way $K$-shot and $N$-way 5-shot scenarios depicted in the main paper are presented in \cref{tab:way_shot_sup}, including the 95\% confidence intervals. While enforcing a well-conditioned parameter space for the inner-loop optimization leads to significantly better first-step adaptation results, it can also be observed that the conditioning seems to additionally improve the overall results achieved after 5 updates. The results further indicate that the adaptation of the baseline parameters during the initial steps (mainly 1-3) differs dependent on the number of shots, and seems to be increasingly delayed to the last steps for settings with a higher number of shots (\eg, $42.10\%$ vs. $21.14\%$ vs. $20.32\%$ after 3 updates for $k=10$, $k=15$ and $k=20$, respectively). 
\begin{table}[h]
\caption{\textbf{Many-way multi-shot experiments.} Average test accuracy for various $5$-way $K$-shot and $N$-way $5$-shot scenarios evaluated on the \textit{mini}ImageNet~\cite{vinyals2016_matchingnet} test set using a Conv6 architecture.}
    \begin{center}
    \scalebox{0.76}
    {
    {
    \setlength{\tabcolsep}{6pt}
    \begin{tabular}{@{}c ll| *5c @{}} 
    \specialrule{.2em}{.1em}{.1em}
          & \textbf{Setting}  & \textbf{Method} & \textbf{step 1$\uparrow$}  & \textbf{step 2$\uparrow$} & \textbf{step 3$\uparrow$}  & \textbf{step 4$\uparrow$} & \textbf{step 5$\uparrow$} \\    \midrule 
         \multirow{10}{*}{\STAB{\rotatebox[origin=c]{90}{5-way}}} 
         & \multirow{2}{*}{1-shot} 
          & MAML & $21.40{\scriptstyle \pm 0.23}$ & $30.85{\scriptstyle \pm 0.66}$ & $45.91{\scriptstyle \pm 0.87}$ & $50.30{\scriptstyle \pm 0.88}$ & $51.22{\scriptstyle \pm 0.88}$ %
          \\
          & & ours  & $50.92{\scriptstyle \pm 0.85}$ & $52.98{\scriptstyle \pm 0.89}$ & $53.18{\scriptstyle \pm 0.89}$ & $53.28{\scriptstyle \pm 0.89}$ & $53.34{\scriptstyle \pm 0.89}$ %
          \\
          \cmidrule{2-8}
         & \multirow{2}{*}{5-shot} 
          & MAML & $20.19{\scriptstyle \pm 0.07}$ & $24.17{\scriptstyle \pm 0.39}$ & $57.25{\scriptstyle \pm 0.72}$ & $64.24{\scriptstyle \pm 0.73}$ & $65.96{\scriptstyle \pm 0.71}$ %
          \\
          & & ours  & $62.31{\scriptstyle \pm 0.72}$ & $66.66{\scriptstyle \pm 0.71}$ & $67.63{\scriptstyle \pm 0.72}$ & $68.21{\scriptstyle \pm 0.71}$ & $68.43{\scriptstyle \pm 0.71}$ %
          \\
          \cmidrule{2-8}
         & \multirow{2}{*}{10-shot} 
          & MAML & $20.00{\scriptstyle \pm 0.00}$ & $20.07{\scriptstyle \pm 0.04}$ & $42.10{\scriptstyle \pm 0.70}$ & $65.48{\scriptstyle \pm 0.74}$ & $70.66{\scriptstyle \pm 0.67}$ 
          \\
          & & ours  & $64.82{\scriptstyle \pm 0.72}$ & $70.35{\scriptstyle \pm 0.70}$ & $71.65{\scriptstyle \pm 0.67}$ & $72.68{\scriptstyle \pm 0.68}$ & $73.13{\scriptstyle \pm 0.67}$ 
          \\
          \cmidrule{2-8}
         & \multirow{2}{*}{15-shot} 
          & MAML & $20.00{\scriptstyle \pm 0.00}$ & $20.00{\scriptstyle \pm 0.00}$ & $21.14{\scriptstyle \pm 0.18}$ & $68.00{\scriptstyle \pm 0.67}$ & $71.22{\scriptstyle \pm 0.63}$ 
          \\
          & & ours  & $63.91{\scriptstyle \pm 0.69}$ & $70.18{\scriptstyle \pm 0.65}$ & $72.30{\scriptstyle \pm 0.65}$ & $73.58{\scriptstyle \pm 0.63}$ & $74.01{\scriptstyle \pm 0.63}$ 
          \\
          \cmidrule{2-8}
         & \multirow{2}{*}{20-shot} 
          & MAML & $20.00{\scriptstyle \pm 0.00}$ & $20.00{\scriptstyle \pm 0.00}$ & $20.32{\scriptstyle \pm 0.08}$ & $69.41{\scriptstyle \pm 0.65}$ & $72.98{\scriptstyle \pm 0.61}$  
          \\
          & & ours  & $64.83{\scriptstyle \pm 0.66}$ & $71.56{\scriptstyle \pm 0.63}$ & $73.56{\scriptstyle \pm 0.60}$ & $74.85{\scriptstyle \pm 0.60}$ & $75.55{\scriptstyle \pm 0.58}$ 
          \\
          \midrule \midrule
          \multirow{6}{*}{\STAB{\rotatebox[origin=c]{90}{5-shot}}} 
         & \multirow{2}{*}{5-way} 
          & MAML & $20.19{\scriptstyle \pm 0.07}$ & $24.17{\scriptstyle \pm 0.39}$ & $57.25{\scriptstyle \pm 0.72}$ & $64.24{\scriptstyle \pm 0.73}$ & $65.96{\scriptstyle \pm 0.71}$  %
          \\
          & & ours & $62.31{\scriptstyle \pm 0.72}$ & $66.66{\scriptstyle \pm 0.71}$ & $67.63{\scriptstyle \pm 0.72}$ & $68.21{\scriptstyle \pm 0.71}$ & $68.43{\scriptstyle \pm 0.71}$ %
          \\
          \cmidrule{2-8}
         & \multirow{2}{*}{10-way} 
          & MAML & $10.00{\scriptstyle \pm 0.00}$ & $14.11{\scriptstyle \pm 0.30}$ & $41.50{\scriptstyle \pm 0.43}$ & $47.38{\scriptstyle \pm 0.47}$ & $49.59{\scriptstyle \pm 0.46}$  
          \\
          & & ours & $42.20{\scriptstyle \pm 0.42}$ & $48.94{\scriptstyle \pm 0.45}$ & $50.79{\scriptstyle \pm 0.46}$ & $51.67{\scriptstyle \pm 0.46}$ & $52.02{\scriptstyle \pm 0.47}$ 
          \\
          \cmidrule{2-8}
          & \multirow{2}{*}{15-way} 
          & MAML & $6.67{\scriptstyle \pm 0.00}$ & $7.66{\scriptstyle \pm 0.27}$ & $35.32{\scriptstyle \pm 0.32}$ & $38.29{\scriptstyle \pm 0.31}$ & $41.08{\scriptstyle \pm 0.31}$ 
          \\
          & & ours  & $32.55{\scriptstyle \pm 0.30}$ & $39.81{\scriptstyle \pm 0.32}$ & $41.83{\scriptstyle \pm 0.32}$ & $42.84{\scriptstyle \pm 0.32}$ & $43.30{\scriptstyle \pm 0.21}$ 
          \\
        \bottomrule
    \end{tabular}
    }
    }
    \end{center}
    \label{tab:way_shot_sup}
\end{table}

\section{Constraining parameter subsets}
\begin{figure}[t]
\centering
\begin{subfigure}[b]{0.48\textwidth}
    \includegraphics[width=\textwidth]{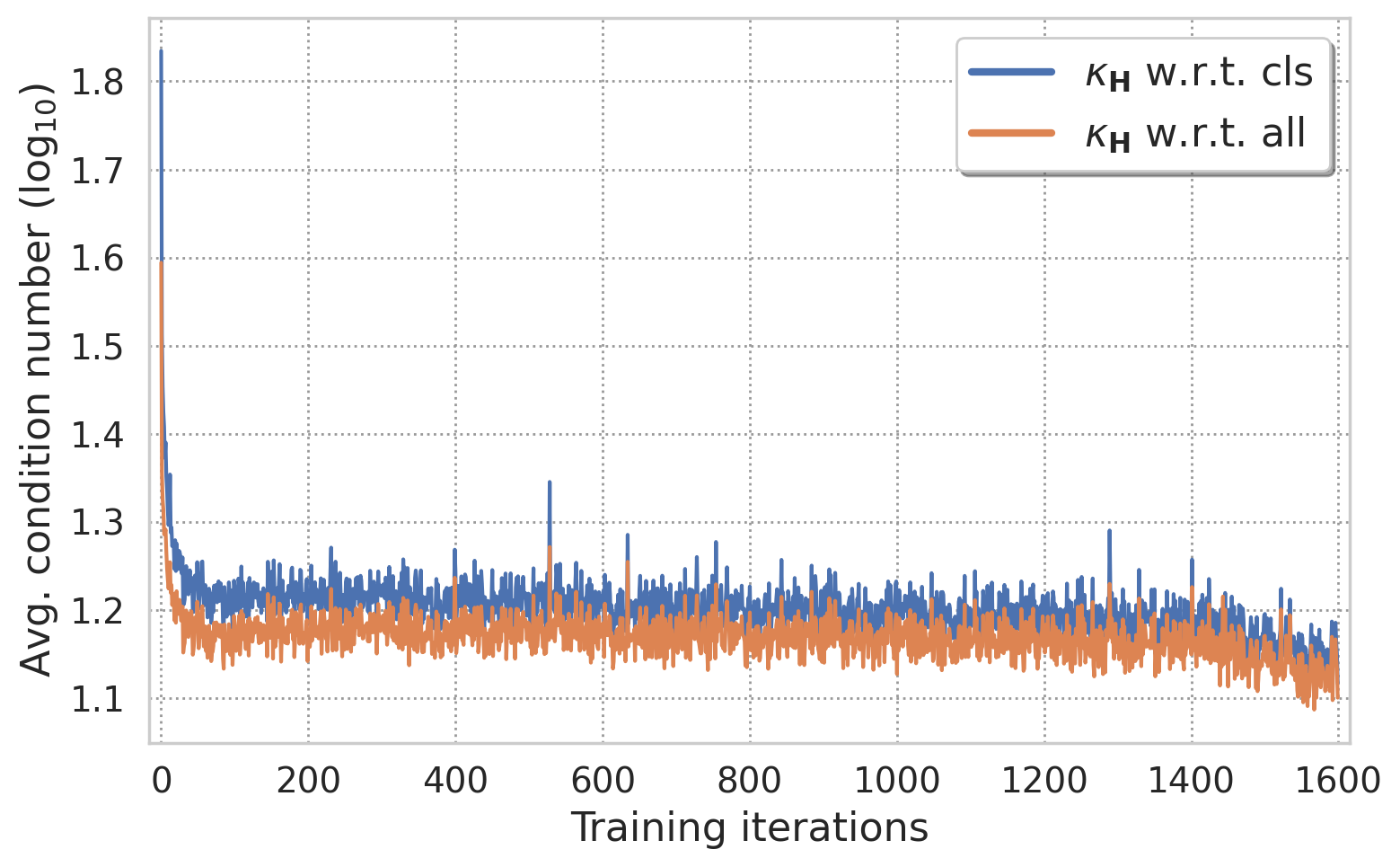}
    \caption{}
    \label{subfig:sup_parsel_cls}
\end{subfigure}
\hfill
\begin{subfigure}[b]{0.48\textwidth}
    \includegraphics[width=\textwidth]{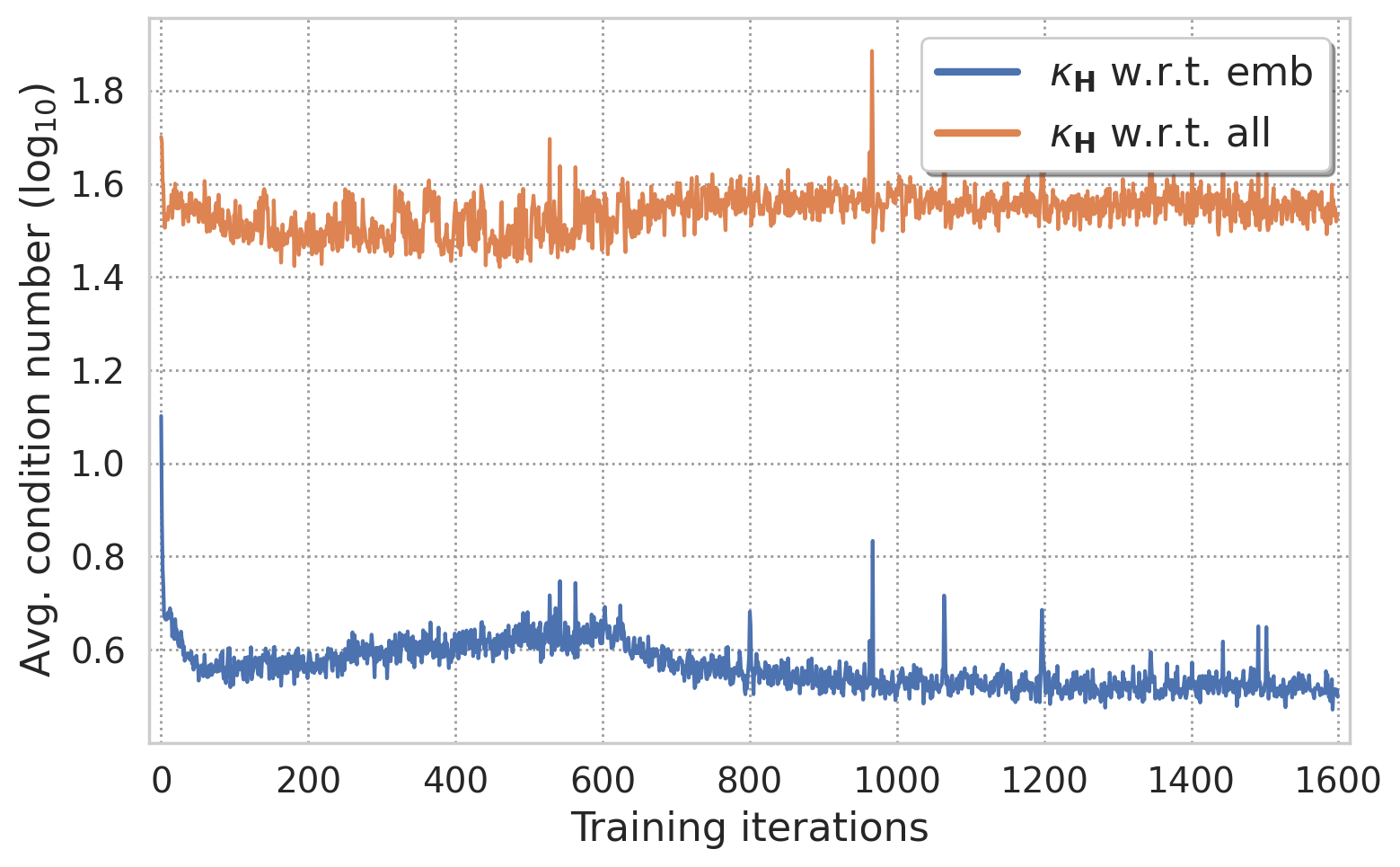}
    \caption{}
    \label{subfig:sup_parsel_emb}
\end{subfigure}
\hfill
\begin{subfigure}[b]{0.48\textwidth}
    \includegraphics[width=\textwidth]{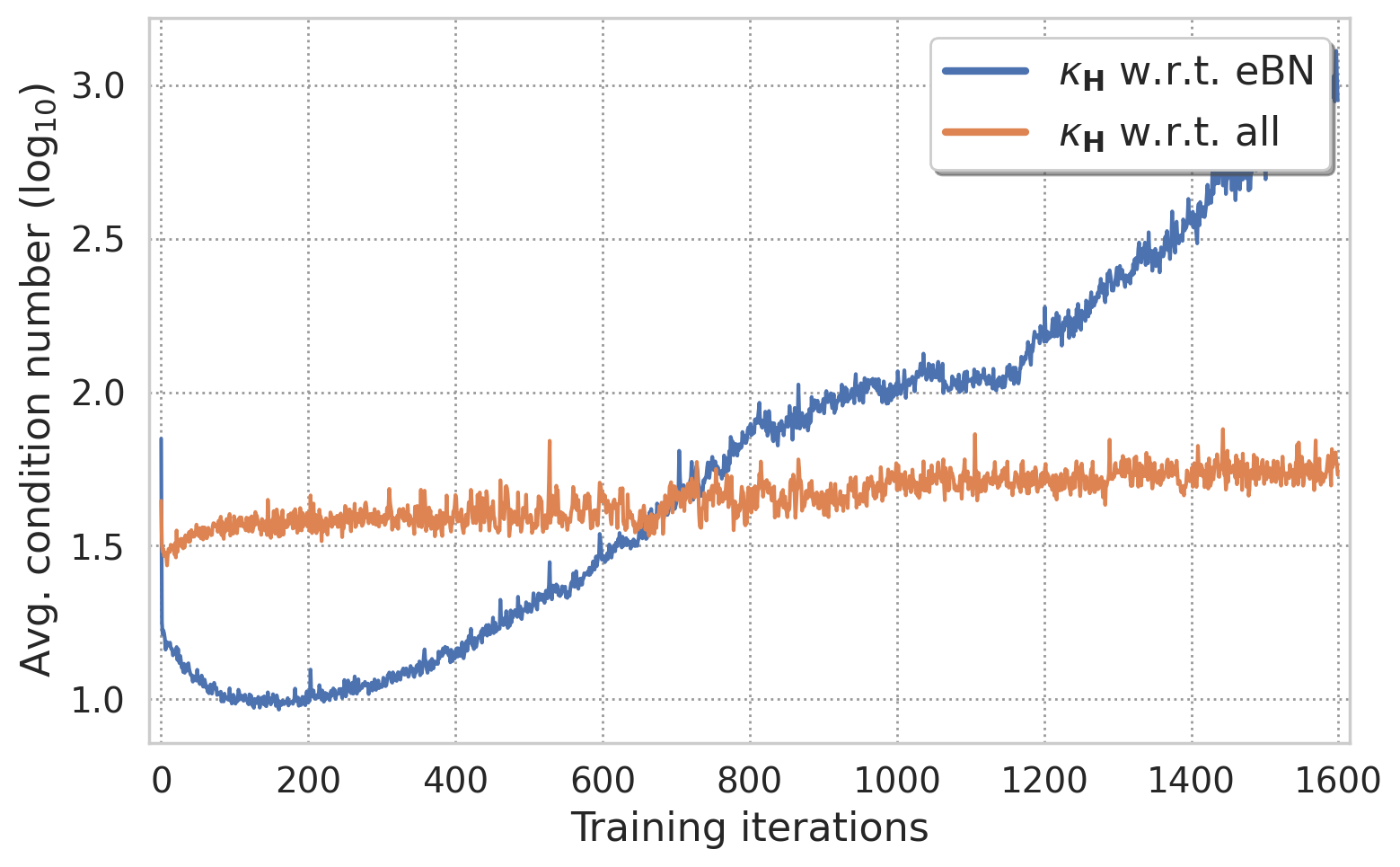}
    \caption{}
    \label{subfig:sup_parsel_eBN}
\end{subfigure}
\hfill
\begin{subfigure}[b]{0.48\textwidth}
    \includegraphics[width=\textwidth]{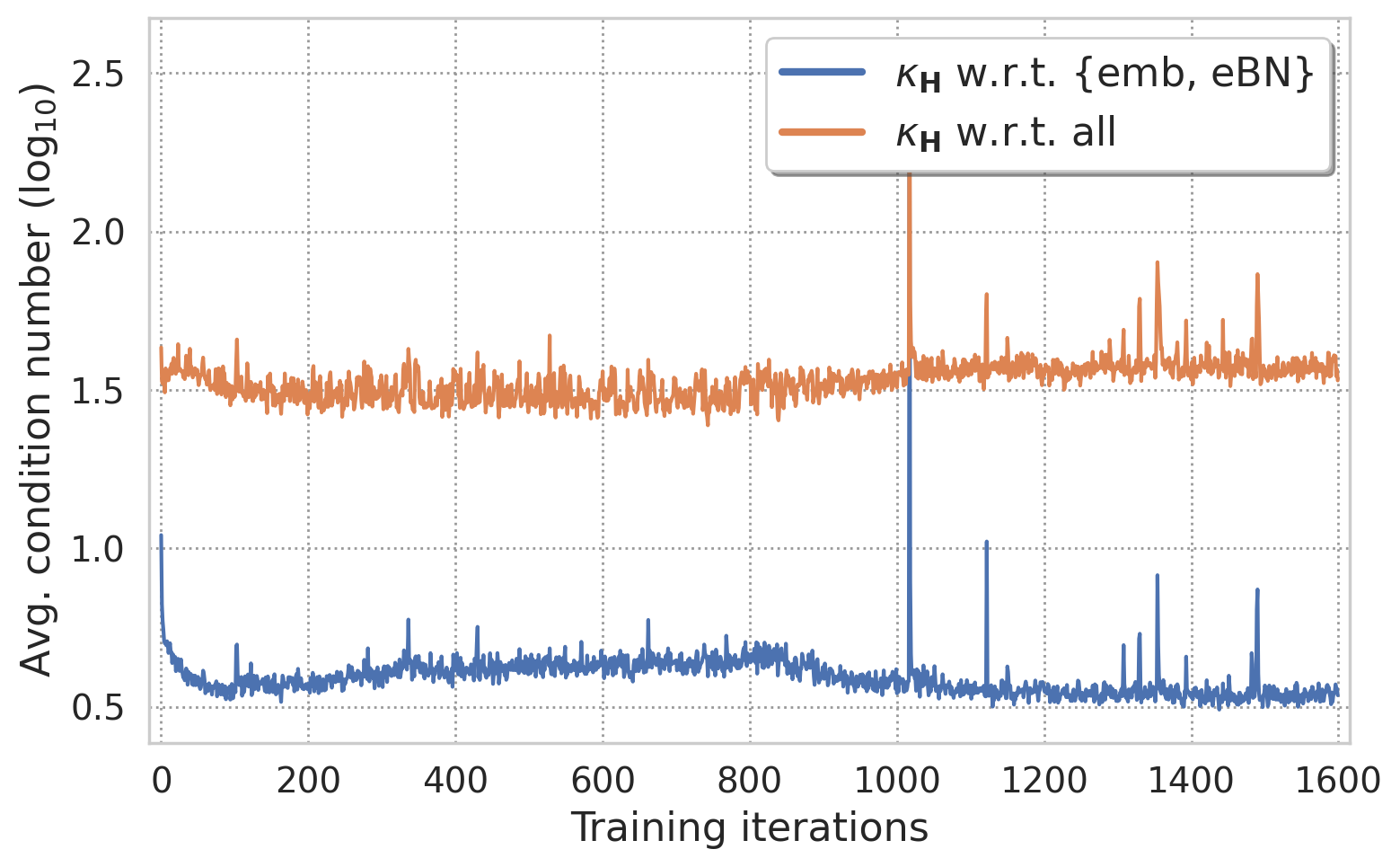}
    \caption{}
    \label{subfig:sup_parsel_emb_eBN}
\end{subfigure}
\hfill
\begin{subfigure}[b]{0.48\textwidth}
    \includegraphics[width=\textwidth]{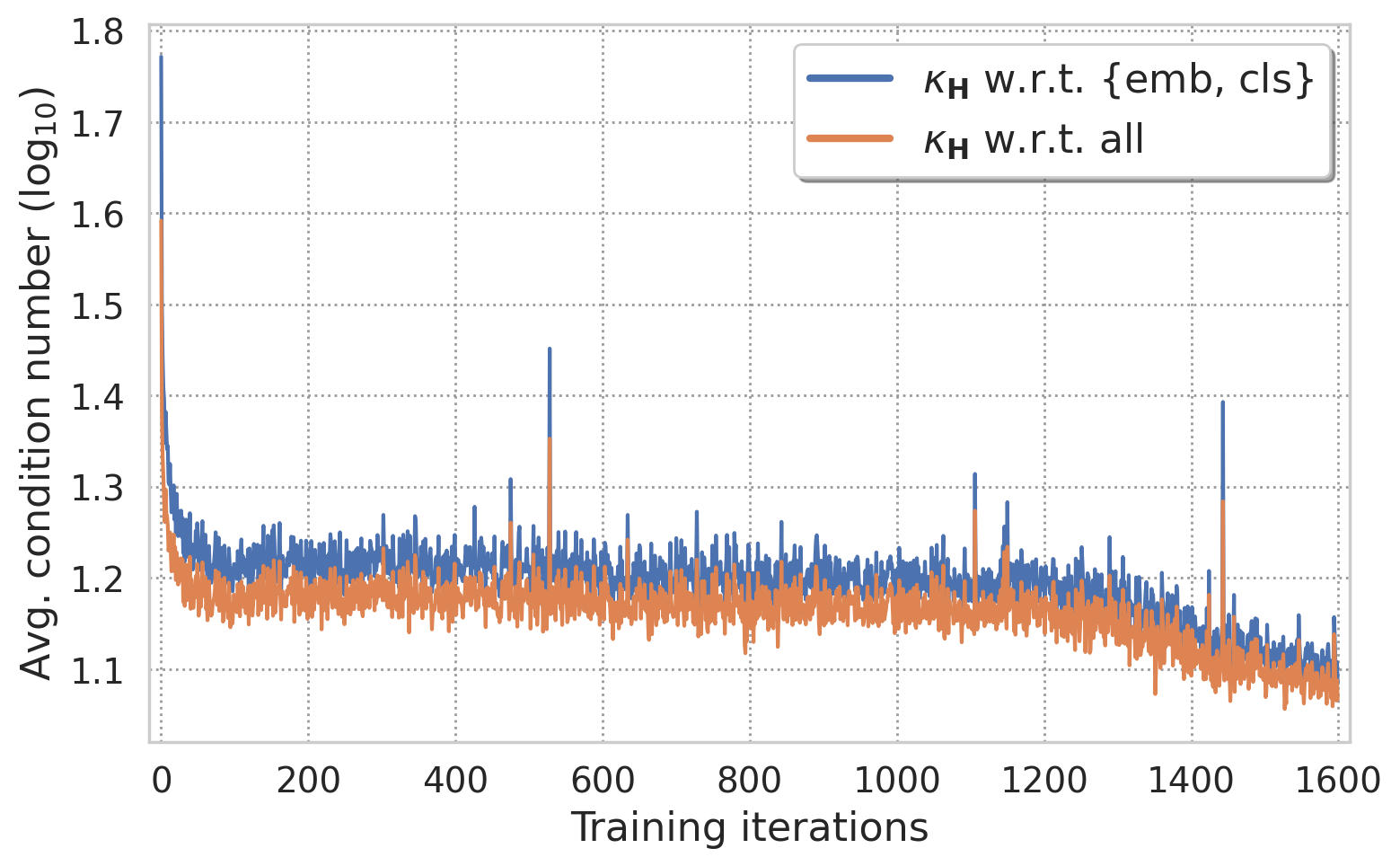}
    \caption{}
    \label{subfig:sup_parsel_emb_cls}
\end{subfigure}
\hfill
\begin{subfigure}[b]{0.48\textwidth}
    \includegraphics[width=\textwidth]{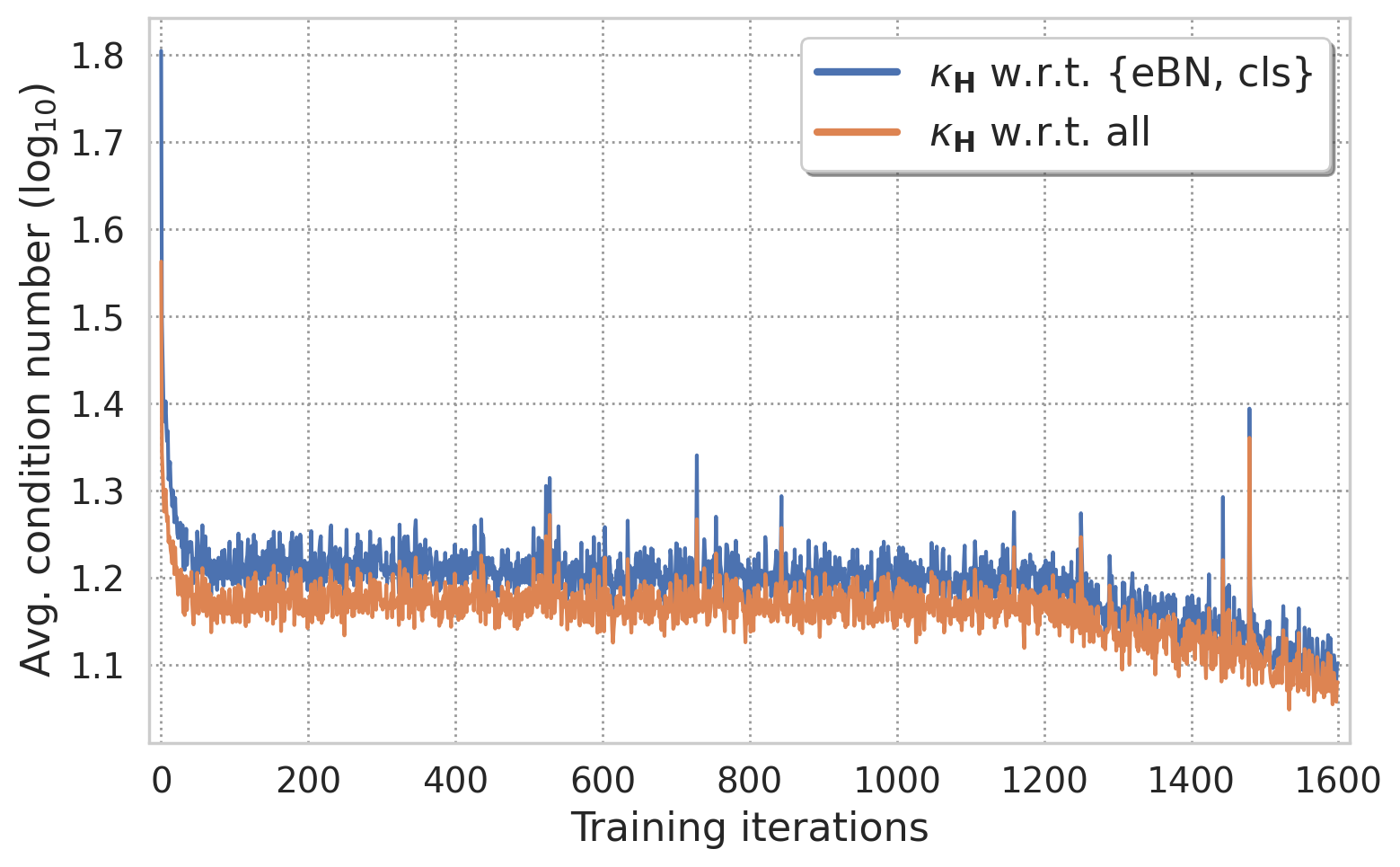}
    \caption{}
    \label{subfig:sup_parsel_eBN_cls}
\end{subfigure}
\hfill
\begin{subfigure}[b]{0.48\textwidth}
    \includegraphics[width=\textwidth]{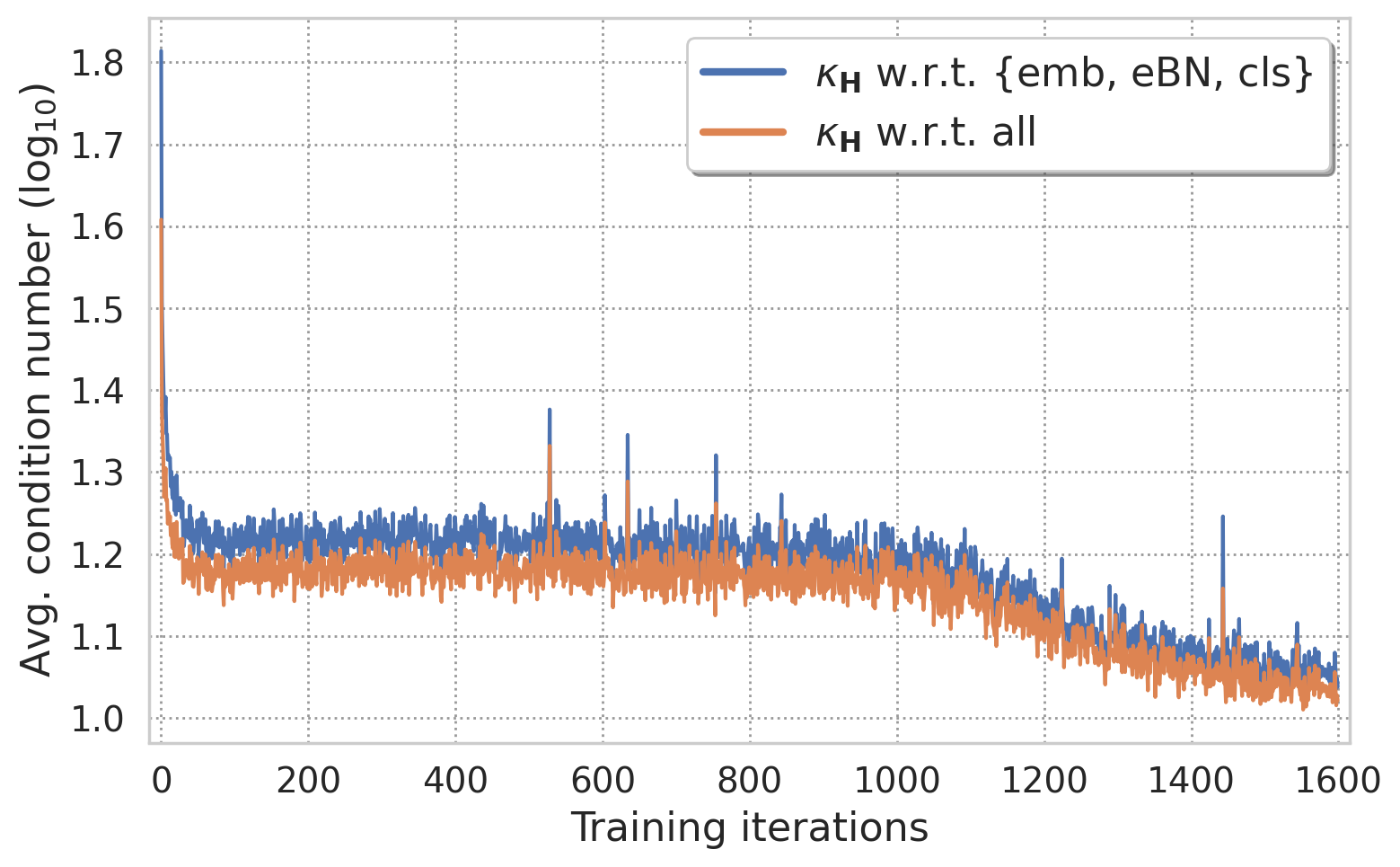}
    \caption{}
    \label{subfig:sup_parsel_emb_eBN_cls}
\end{subfigure}
\hfill
\begin{subfigure}[b]{0.48\textwidth}
    \includegraphics[width=\textwidth]{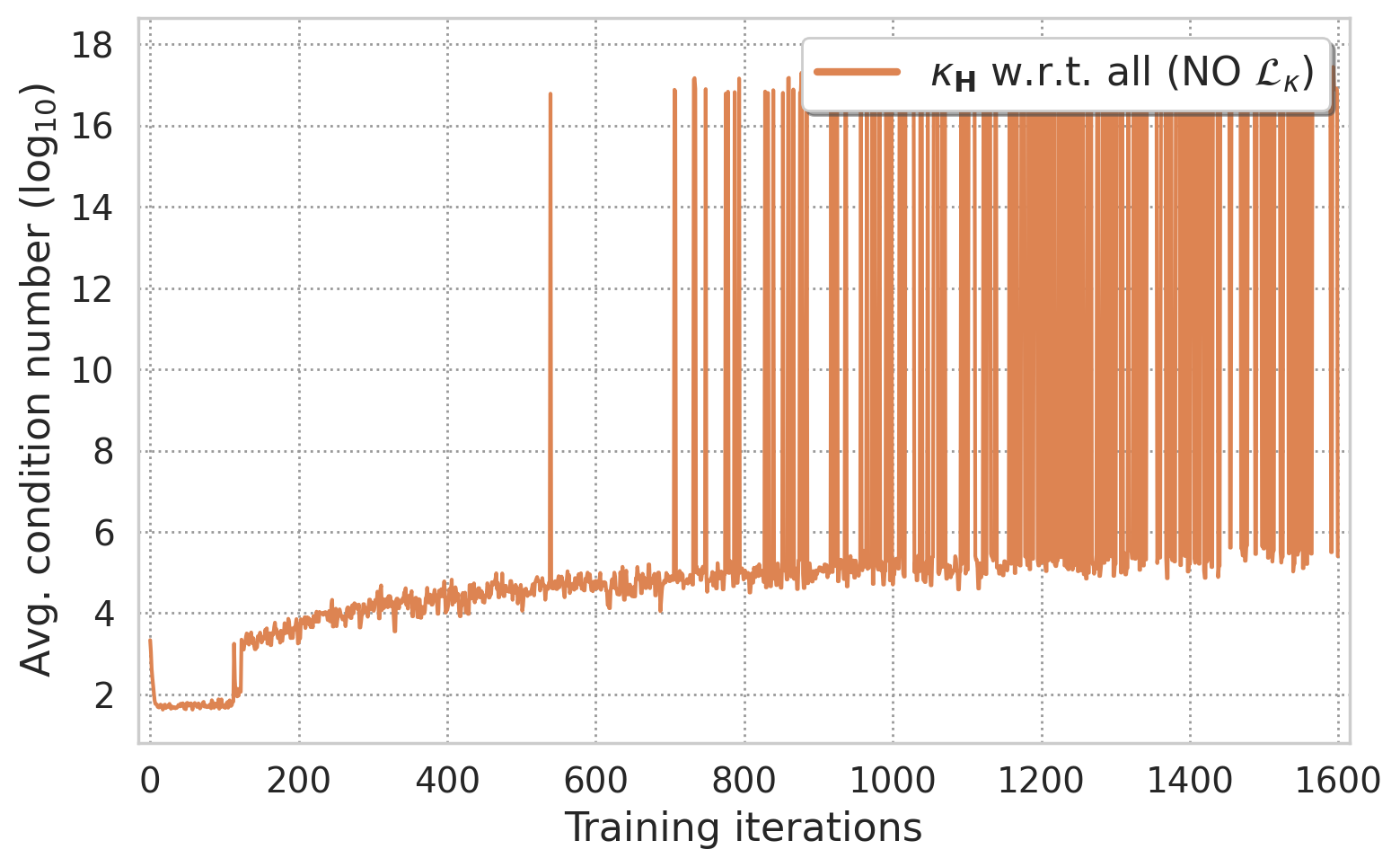}
    \caption{}
    \label{subfig:sup_parsel_noco}
\end{subfigure}
\caption{\textbf{Constraining a reduced parameter set.}  Condition number with respect to the reduced parameter subset denoted in the respective legend, and to all parameters of the model over 1600 iterations on the \textit{mini}ImageNet dataset with a Conv4 architecture for a 5-way 5-shot scenario. Models in (\subref{subfig:sup_parsel_cls}) -  (\subref{subfig:sup_parsel_emb_eBN_cls}) are trained with $\nicecal{L}_{\kappa}$ \textit{w.r.t.} the respective subset, while (\subref{subfig:sup_parsel_noco}) shows the development for the model trained without the use of the proposed~$\nicecal{L}_{\kappa}$.}
\label{fig:parsel_sup}
\end{figure}

As discussed in the main paper, we choose to apply our proposed conditioning constraint only to a subset of the network's parameters to increase efficiency and scalability. We demonstrated that the development of the condition number calculated with respect to only the parameters of the classifier is representative for the condition number calculated with respect to the full set of network parameters. In this section, we provide the visualisations of the development of all evaluated subsets. It is to be noted that for all depicted results, the condition constraint is enforced to the parameter subset denoted in the respective legend. As can be observed in \cref{fig:parsel_sup}, all subsets except for the batchnorm of the embedding layer `\textit{eBN}' demonstrate a development of the condition number that is very similar to the one of the condition number \textit{w.r.t.} the full parameter set. For completeness, we additionally provide the development of the condition number \textit{w.r.t.} the full parameter set if the model is trained \textit{without} our proposed conditioning loss in \cref{fig:parsel_sup}\,(\subref{subfig:sup_parsel_noco}) (\ie, conventional MAML baseline like proposed by~\citet{finn2017_maml}) -- demonstrating the significantly higher condition number and thus worse-conditioned parameter space that is learned by the unconstrained method. 
In stark contrast, it can further be observed that the trajectories of the methods actively enforcing conditioning are very close for all subsets where the parameters of the classifier `\textit{cls}' are involved in the conditioning constraint, and that the condition numbers of the actual network (`\textit{all}') is particularly low for all these cases, justifying our choice of using the `\textit{cls}' subset throughout all major experiments in the main paper.

\section{Condition number and few-step performance}
\begin{figure}[h]
\centering
\begin{subfigure}[b]{0.48\textwidth}
    \includegraphics[width=\textwidth]{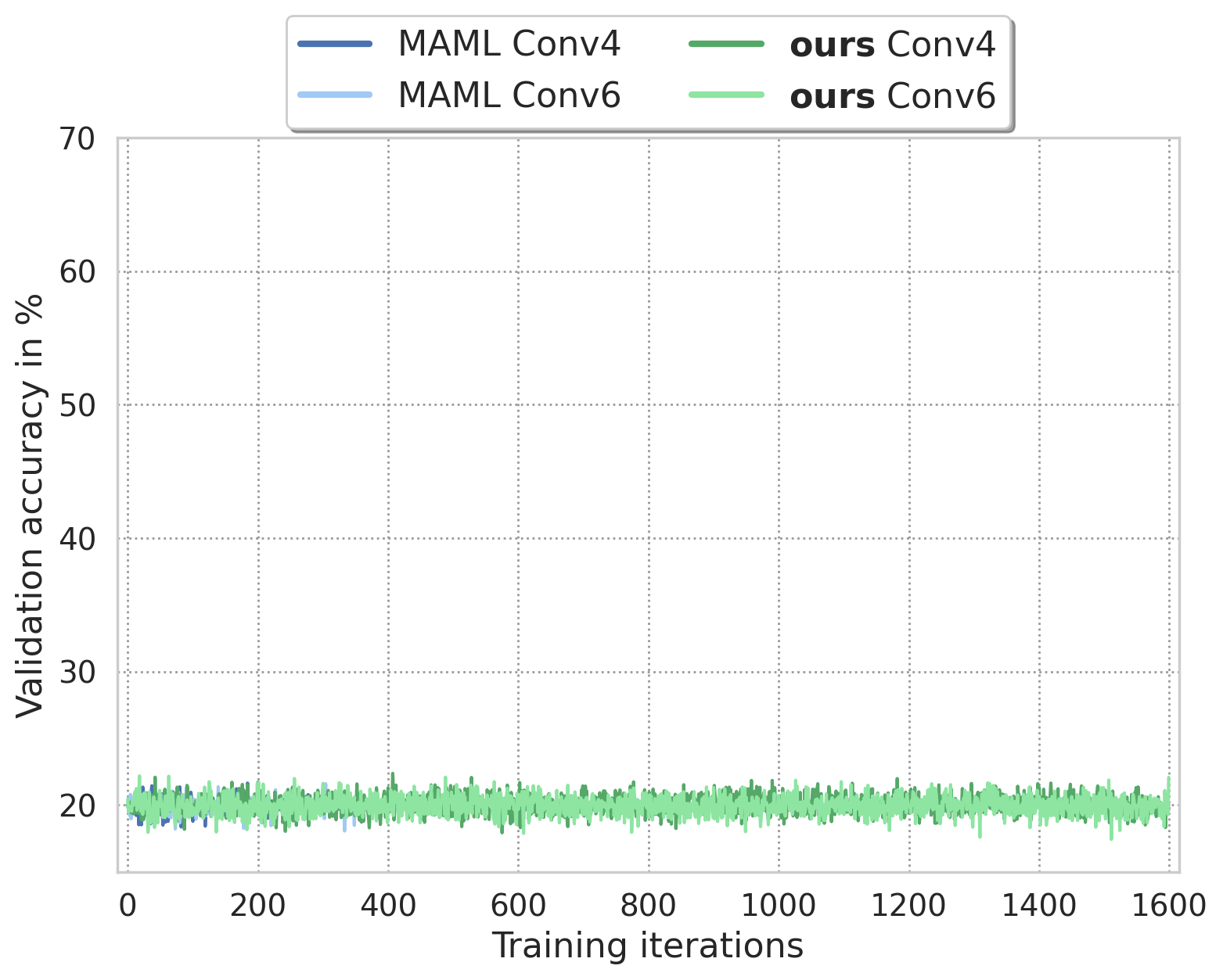}
    \caption{Step 0: $\boldsymbol{\theta}^{(0)}$}
    \label{subfig:sup_valacc_s0}
    \vspace*{0.15in}
\end{subfigure}
\hfill
\begin{subfigure}[b]{0.48\textwidth}
    \includegraphics[width=\textwidth]{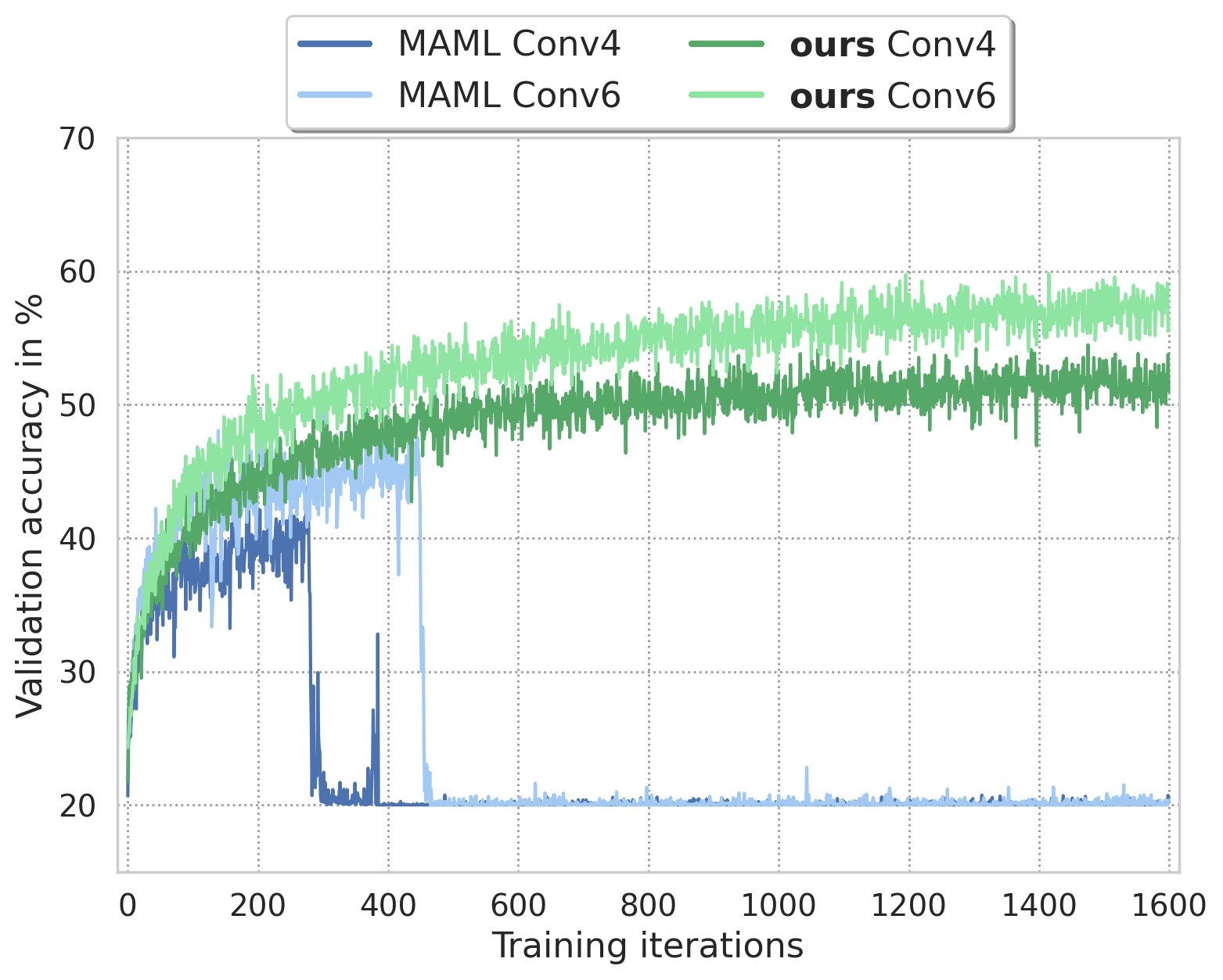}
    \caption{Step 1: $\boldsymbol{\theta}^{(1)}$}
    \label{subfig:sup_valacc_s1}
    \vspace*{0.15in}
\end{subfigure}
\hfill
\begin{subfigure}[b]{0.48\textwidth}
    \includegraphics[width=\textwidth]{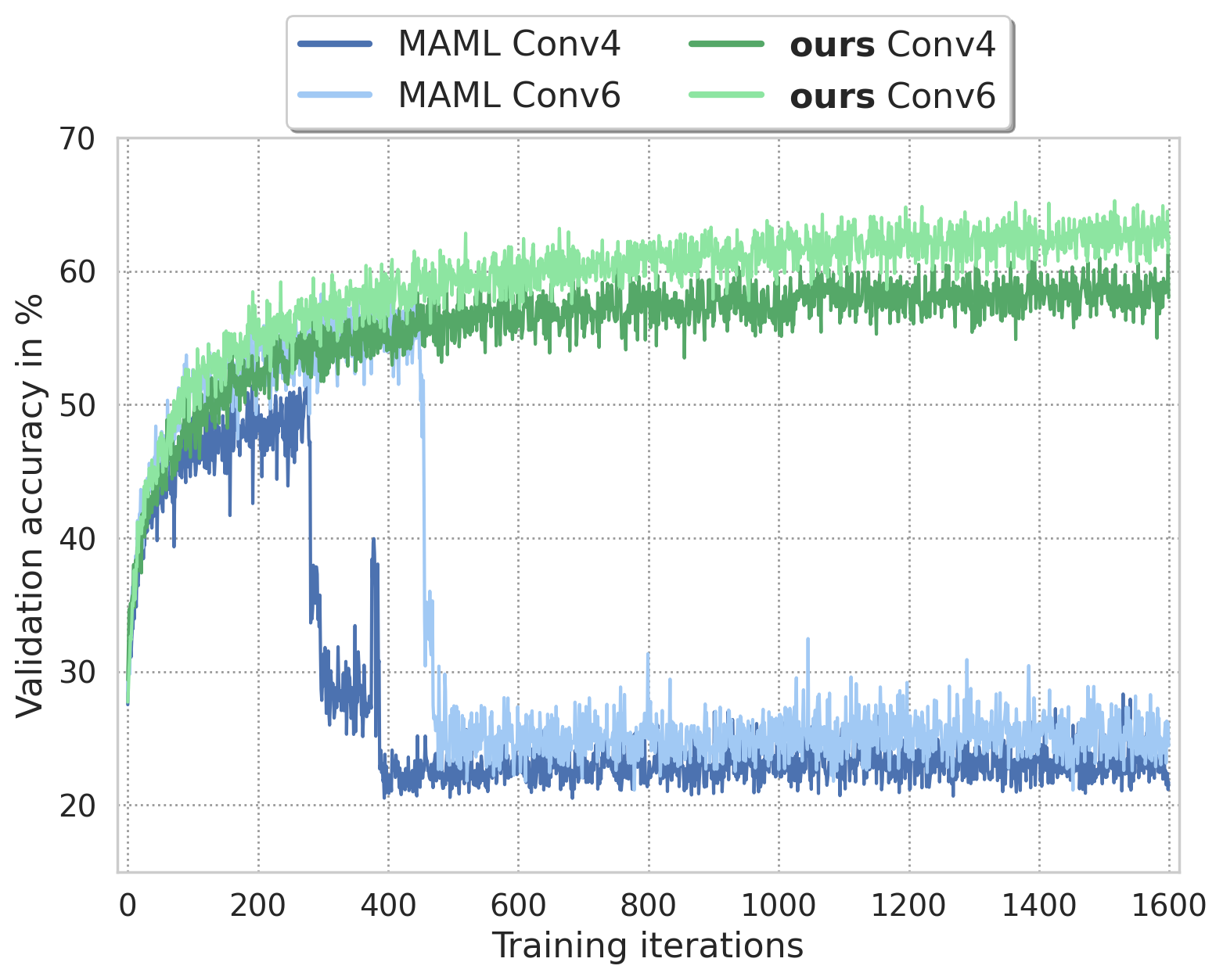}
    \caption{Step 2: $\boldsymbol{\theta}^{(2)}$}
    \label{subfig:sup_valacc_s2}
    \vspace*{0.15in}
\end{subfigure}
\hfill
\begin{subfigure}[b]{0.48\textwidth}
    \includegraphics[width=\textwidth]{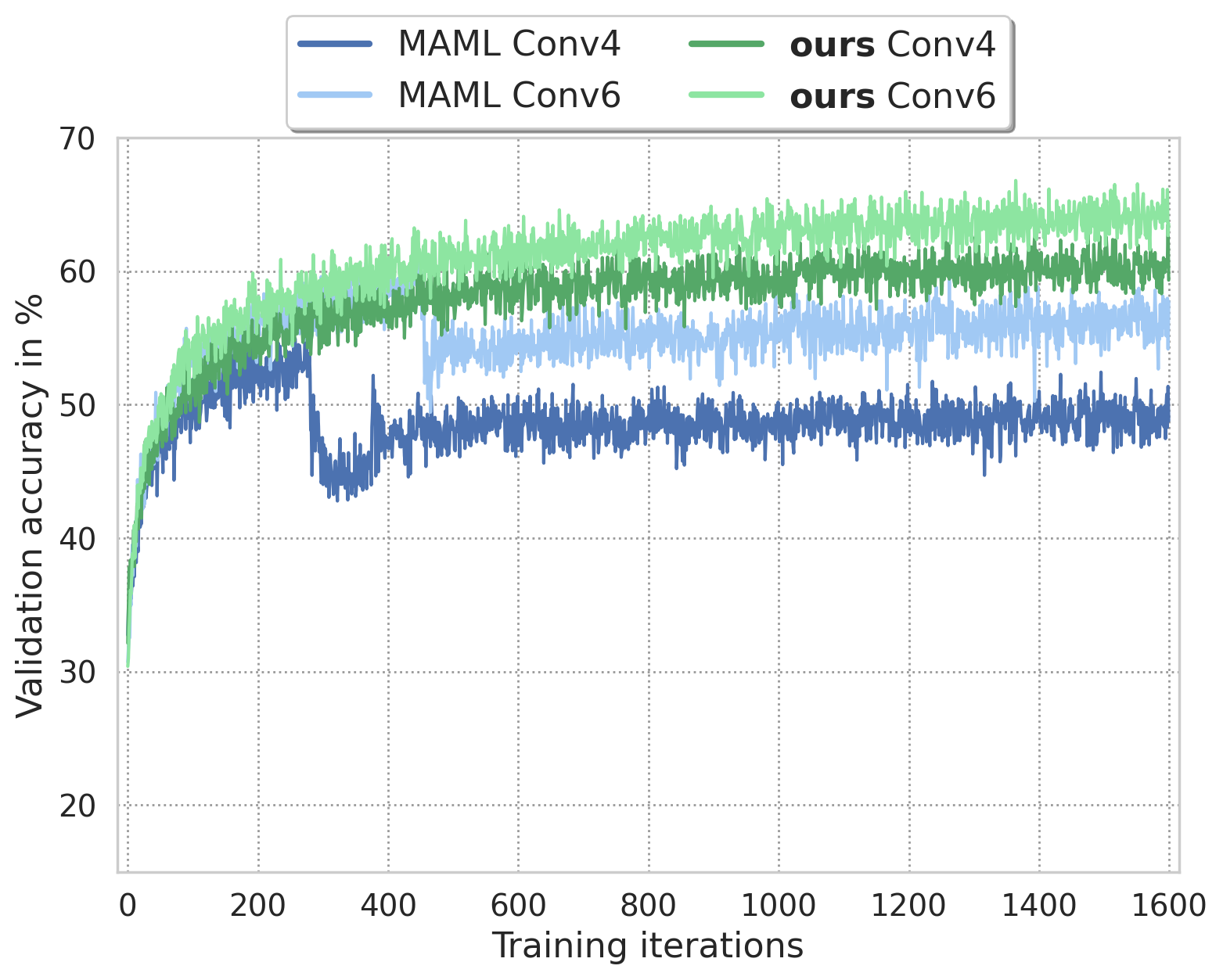}
    \caption{Step 3: $\boldsymbol{\theta}^{(3)}$}
    \label{subfig:sup_valacc_s3}
    \vspace*{0.15in}
\end{subfigure}
\hfill
\begin{subfigure}[b]{0.48\textwidth}
    \includegraphics[width=\textwidth]{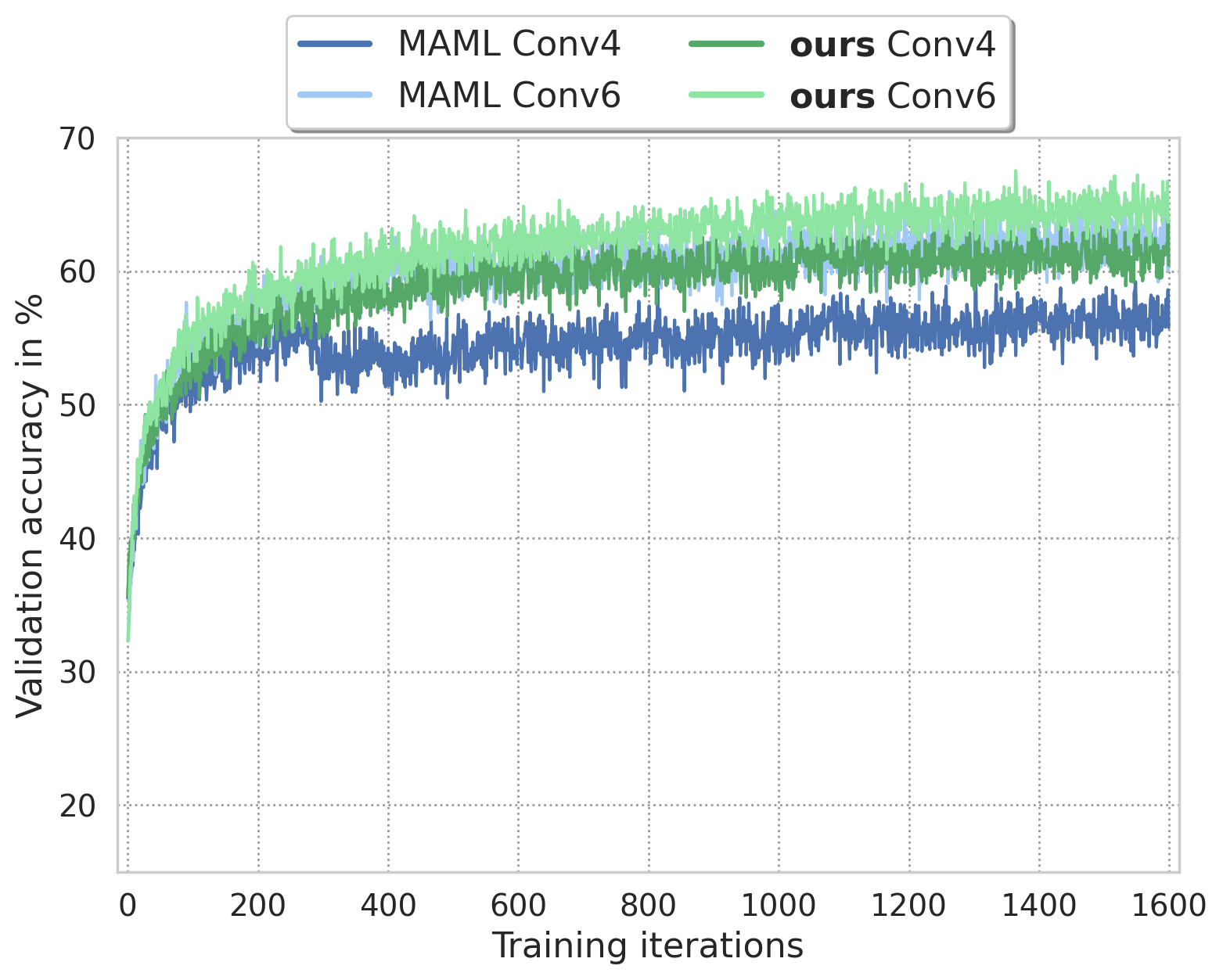}
    \caption{Step 4: $\boldsymbol{\theta}^{(4)}$}
    \label{subfig:sup_valacc_s4}
    \vspace*{0.15in}
\end{subfigure}
\hfill
\begin{subfigure}[b]{0.48\textwidth}
    \includegraphics[width=\textwidth]{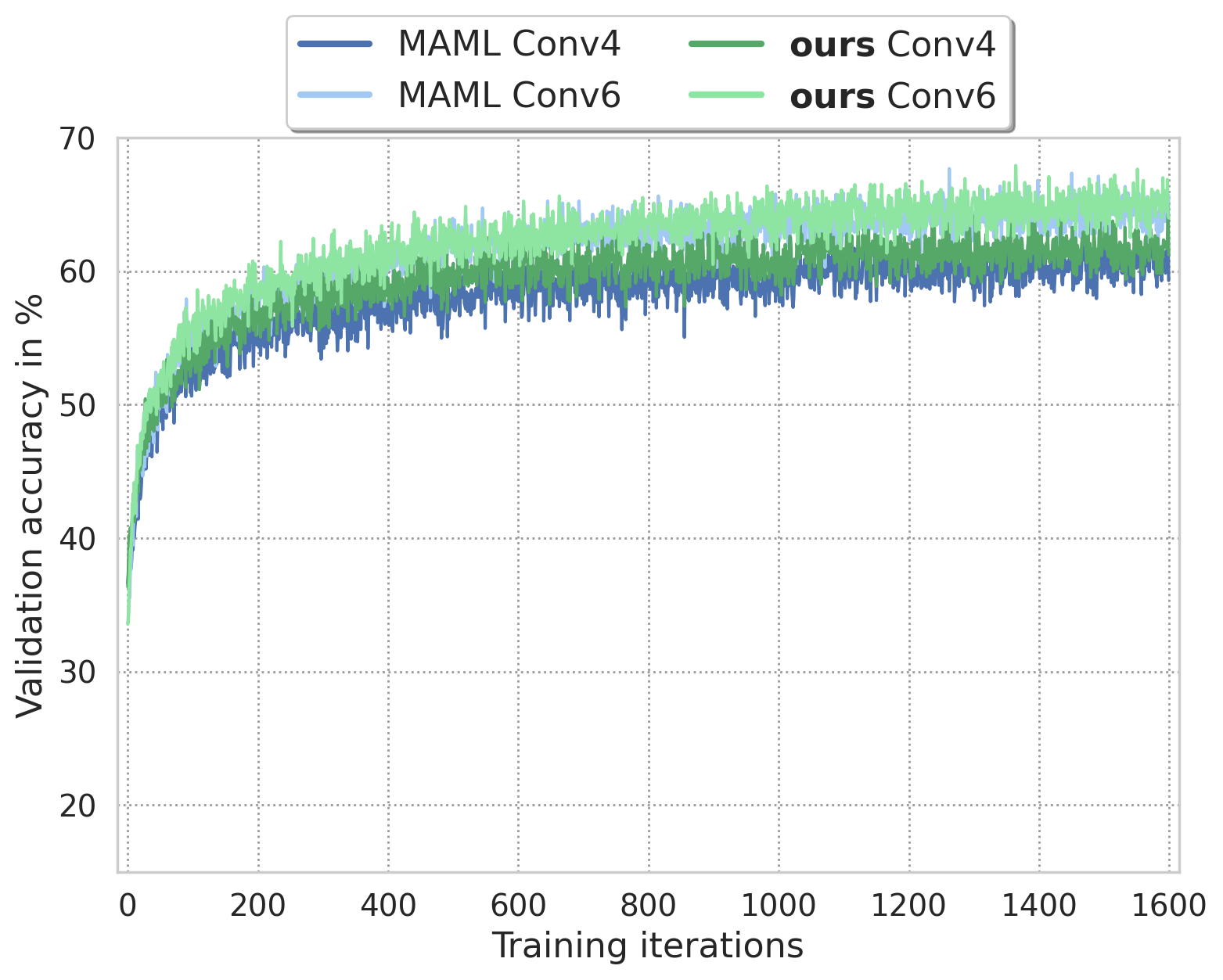}
    \caption{Step 5: $\boldsymbol{\theta}^{(5)}$}
    \label{subfig:sup_valacc_s5}
    \vspace*{0.15in}
\end{subfigure}
\caption{\textbf{Validation accuracy over inner-loop update steps.} Reported results obtained by training the baseline \textit{without} (`MAML') and \textit{with} our proposed conditioning constraint (`ours') with respect to the parameters of the model's classifier. Training has been conducted over 1600 iterations in a 5-way 5-shot scenario on the \textit{tiered}ImageNet dataset with a Conv4 and Conv6 architecture.}
\label{fig:sup_valcond_valacc}
\end{figure}

As discussed in the main paper, the development of the condition number and the validation accuracy are directly related. While we presented the validation accuracies for a Conv4 and Conv6 architecture together with the condition number of inner-loop update step~$1$ in the main paper, we herein show the detailed development of all five inner-loop update steps. The corresponding visualisations of the classification accuracy achieved on the validation set during training are presented in \cref{fig:sup_valcond_valacc} for a Conv4 and Conv6 architecture trained \textit{without} (`MAML') and \textit{with} (`ours') the proposed conditioning constraint enforced via~$\nicecal{L}_\kappa$. We further show the development of the condition number with respect to the parameters of the classifier using the support sets of the training data $(\text{left column,} \;\kappa({\boldsymbol{\theta}^{(k)}_{\mathrm{train}}}))$ and the validation data~$(\text{right column,}\; \kappa({\boldsymbol{\theta}^{(k)}_{\mathrm{valid}}}))$ in \cref{fig:sup_valcond_cnmbr} for steps $k=0$ up to $k=4$, \ie, all parameter sets that will be updated during the course of the 5 inner-loop update steps. Note that the condition property of the initial parameter space at step~$0$ is important to perform the first inner-loop update (step~1), which is why we investigate the condition numbers of the parameter sets before each update (\ie, sets at stages 0 - 4 for update steps 1 - 5). Both architectures have been trained on the \textit{tiered}ImageNet dataset~\cite{ren2018_metasemisup}. 

\begin{figure}[h]
\centering
\begin{subfigure}[b]{0.48\textwidth}
    \includegraphics[width=\textwidth]{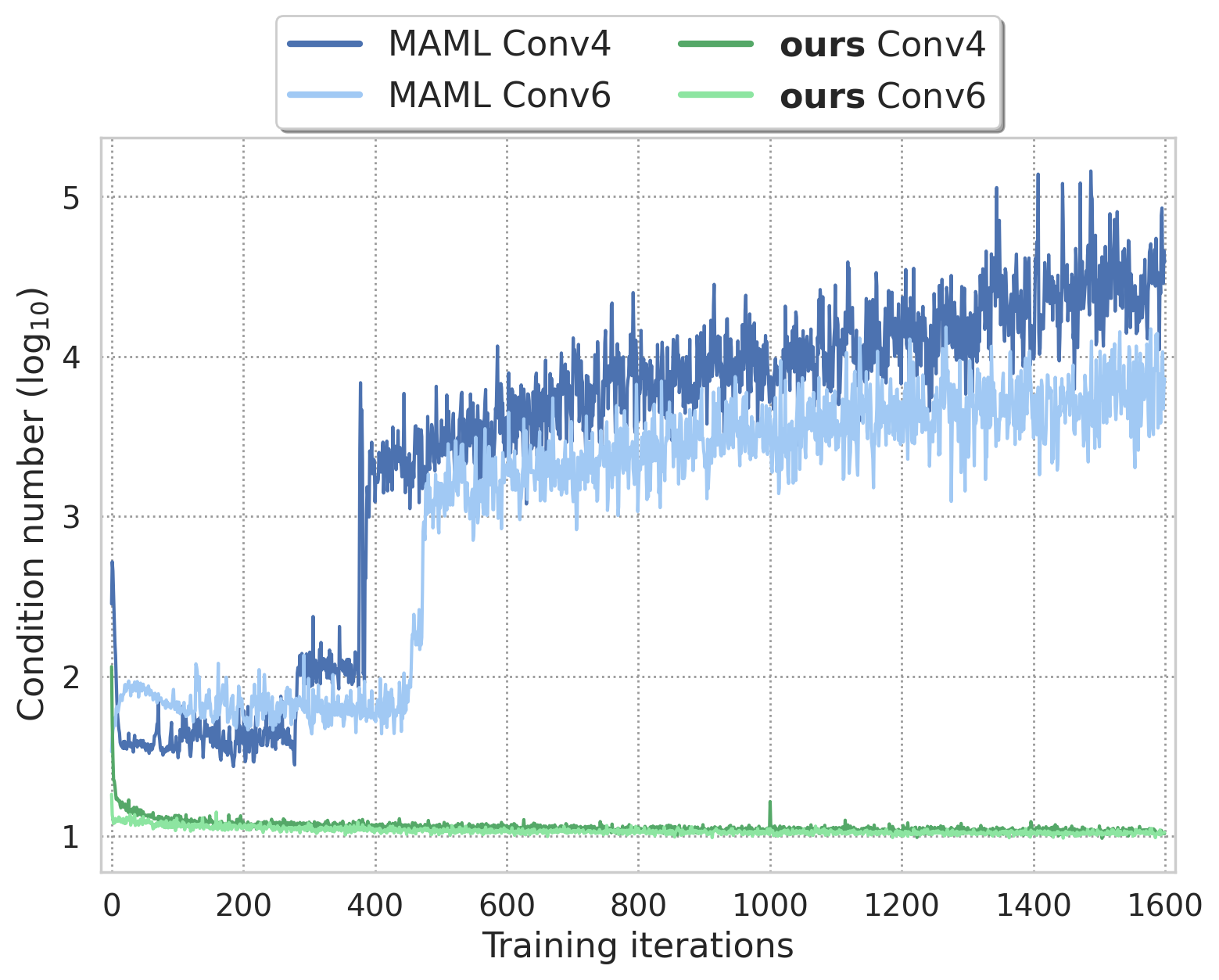}
    \caption{Step 0: $\kappa\left(\boldsymbol{\theta}^{(0)}_{\mathrm{train}}\right)$}
    \label{subfig:sup_cnmbr_train_s0}
    \vspace*{0.2in}
\end{subfigure}
\hfill
\begin{subfigure}[b]{0.48\textwidth}
    \includegraphics[width=\textwidth]{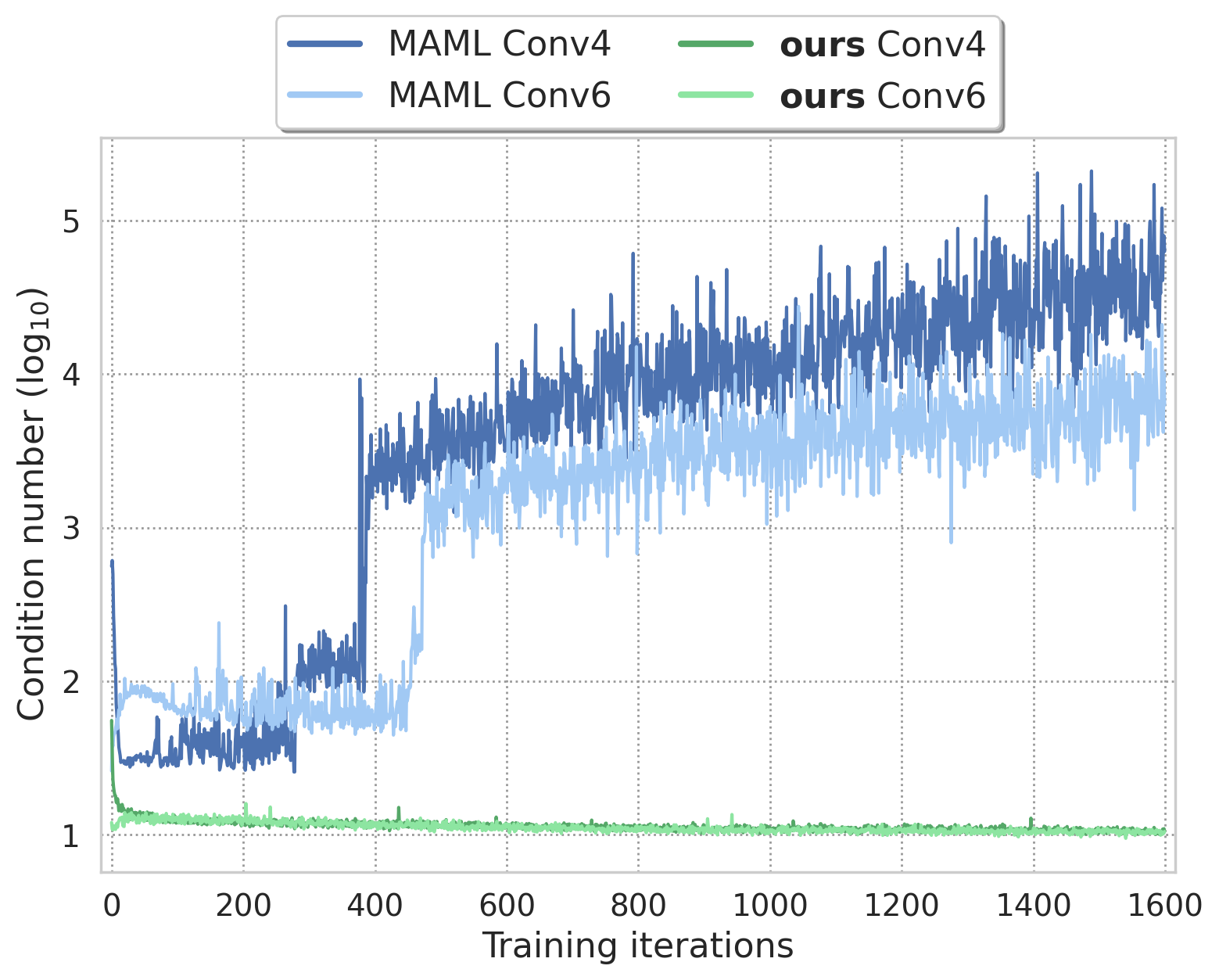}
    \caption{Step 0: $\kappa\left(\boldsymbol{\theta}^{(0)}_{\mathrm{valid}}\right)$}
    \label{subfig:sup_cnmbr_val_s0}
    \vspace*{0.2in}
\end{subfigure}
\hfill
\begin{subfigure}[b]{0.48\textwidth}
    \includegraphics[width=\textwidth]{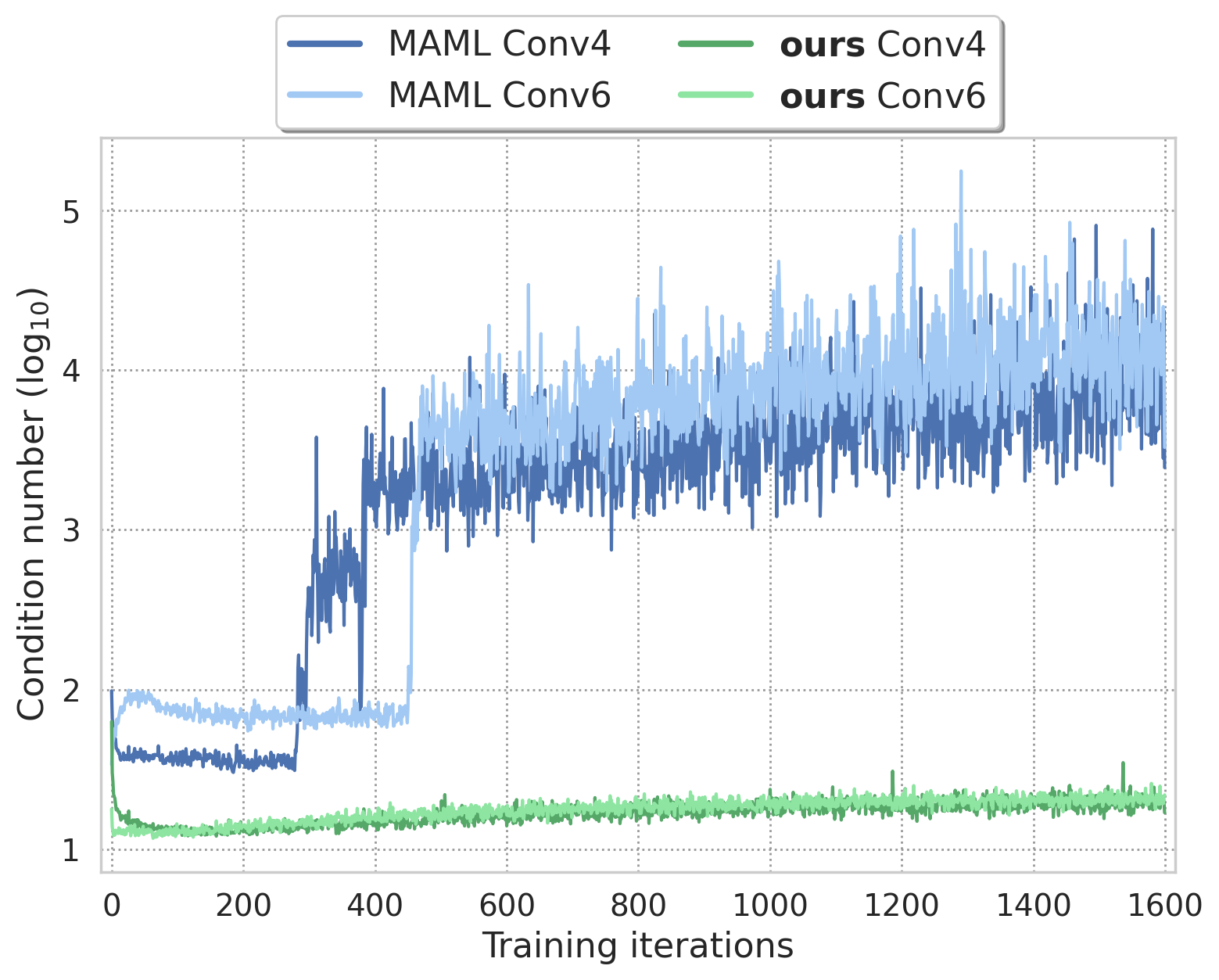}
    \caption{Step 1: $\kappa\left(\boldsymbol{\theta}^{(1)}_{\mathrm{train}}\right)$}
    \label{subfig:sup_cnmbr_train_s1}
    \vspace*{0.2in}
\end{subfigure}
\hfill
\begin{subfigure}[b]{0.48\textwidth}
    \includegraphics[width=\textwidth]{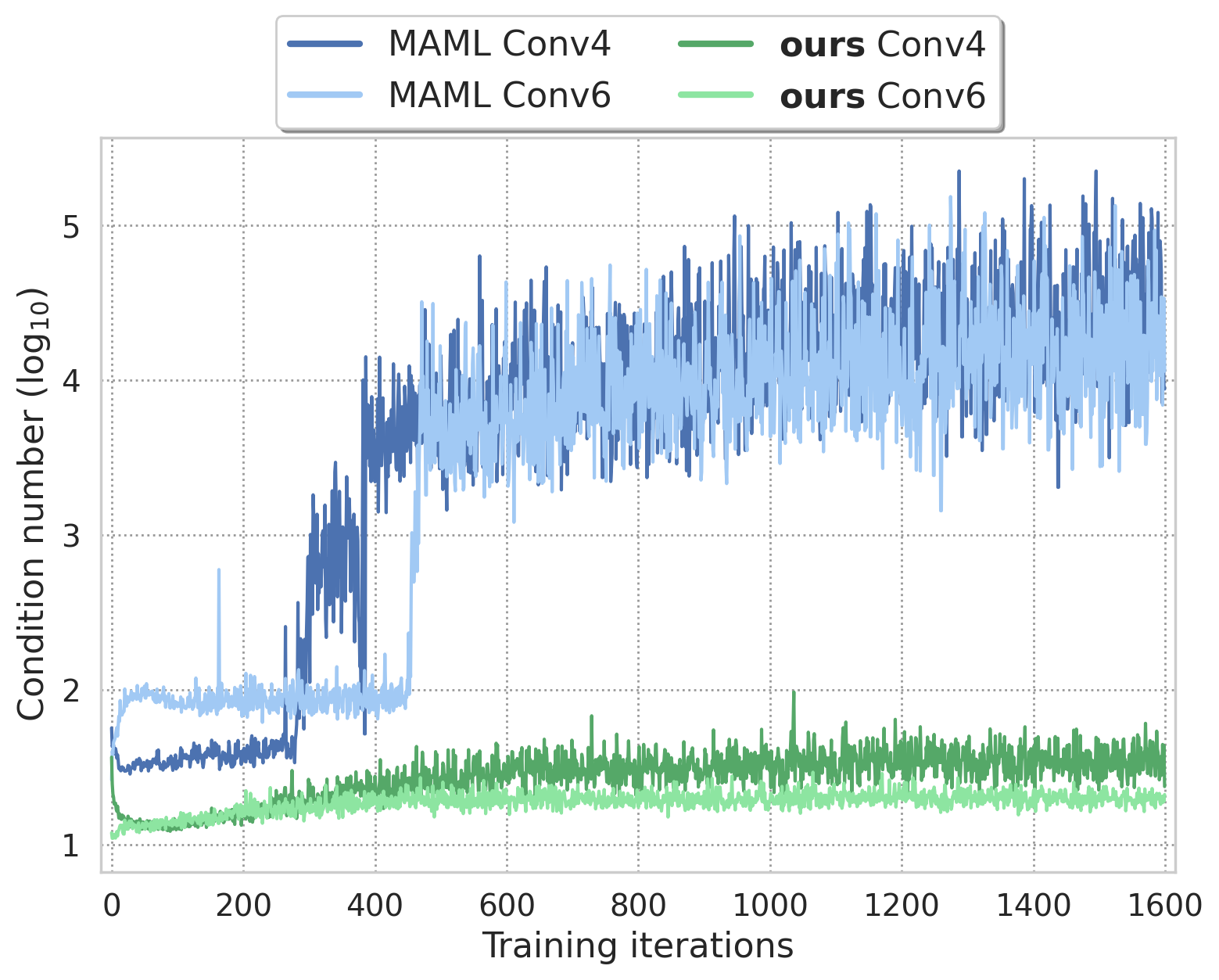}
    \caption{Step 1: $\kappa\left(\boldsymbol{\theta}^{(1)}_{\mathrm{valid}}\right)$}
    \label{subfig:sup_cnmbr_val_s1}
    \vspace*{0.2in}
\end{subfigure}
\hfill
\begin{subfigure}[b]{0.48\textwidth}
    \includegraphics[width=\textwidth]{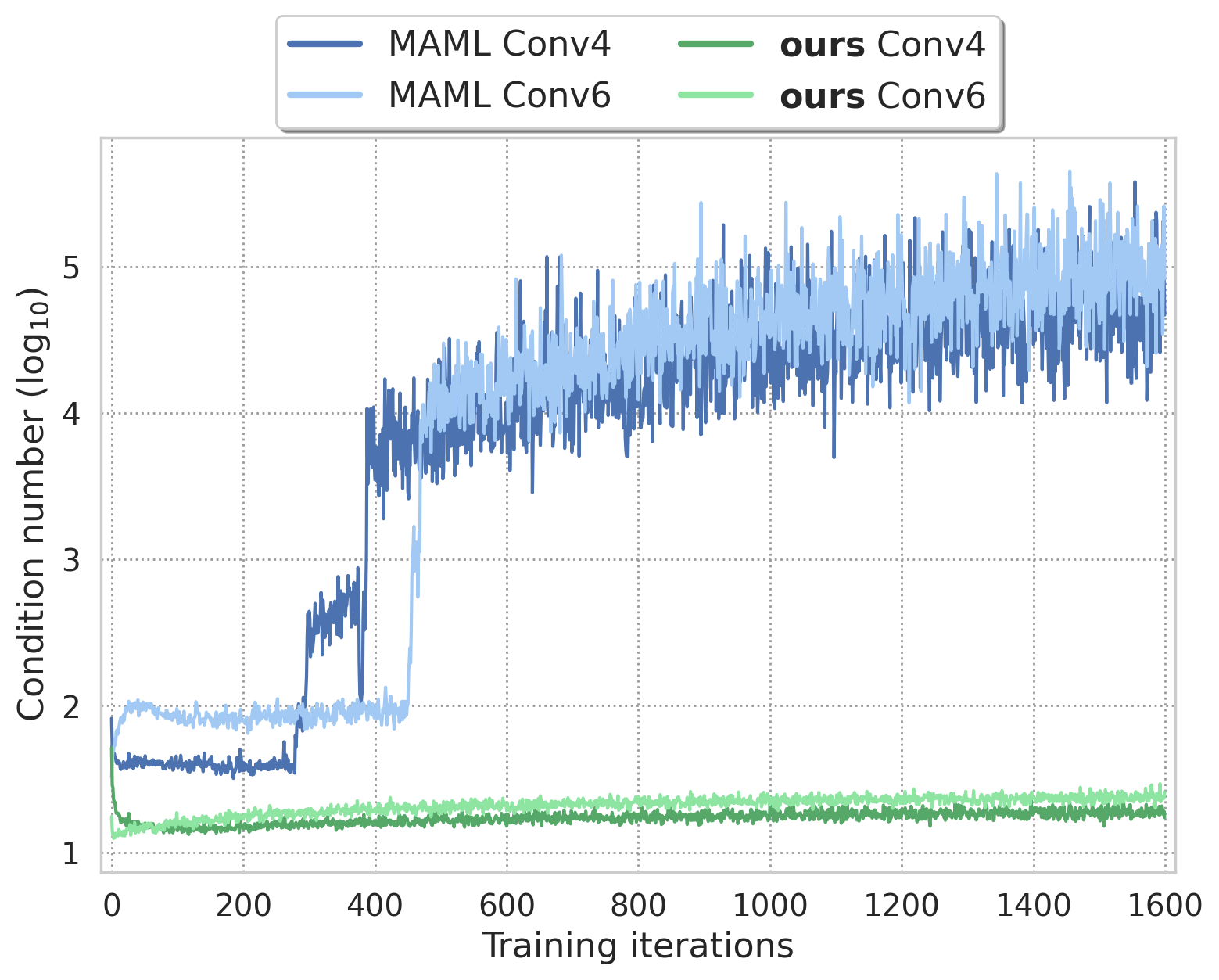}
    \caption{Step 2: $\kappa\left(\boldsymbol{\theta}^{(2)}_{\mathrm{train}}\right)$}
    \label{subfig:sup_cnmbr_train_s2}
    \vspace*{0.2in}
\end{subfigure}
\hfill
\begin{subfigure}[b]{0.48\textwidth}
    \includegraphics[width=\textwidth]{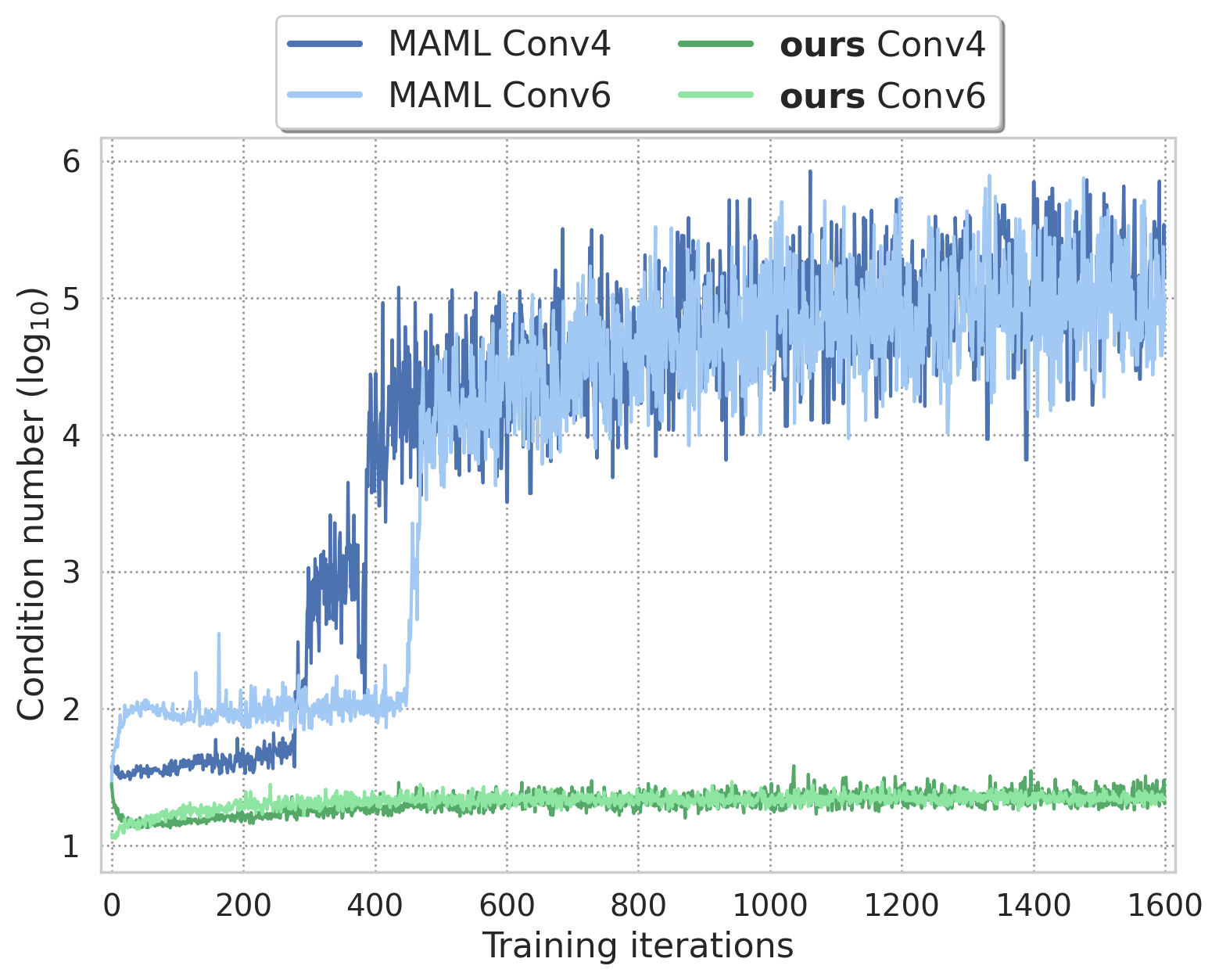}
    \caption{Step 2: $\kappa\left(\boldsymbol{\theta}^{(2)}_{\mathrm{valid}}\right)$}
    \label{subfig:sup_cnmbr_val_s2}
    \vspace*{0.2in}
\end{subfigure}
\end{figure}
\begin{figure}[h]\ContinuedFloat
\centering
\begin{subfigure}[b]{0.48\textwidth}
    \includegraphics[width=\textwidth]{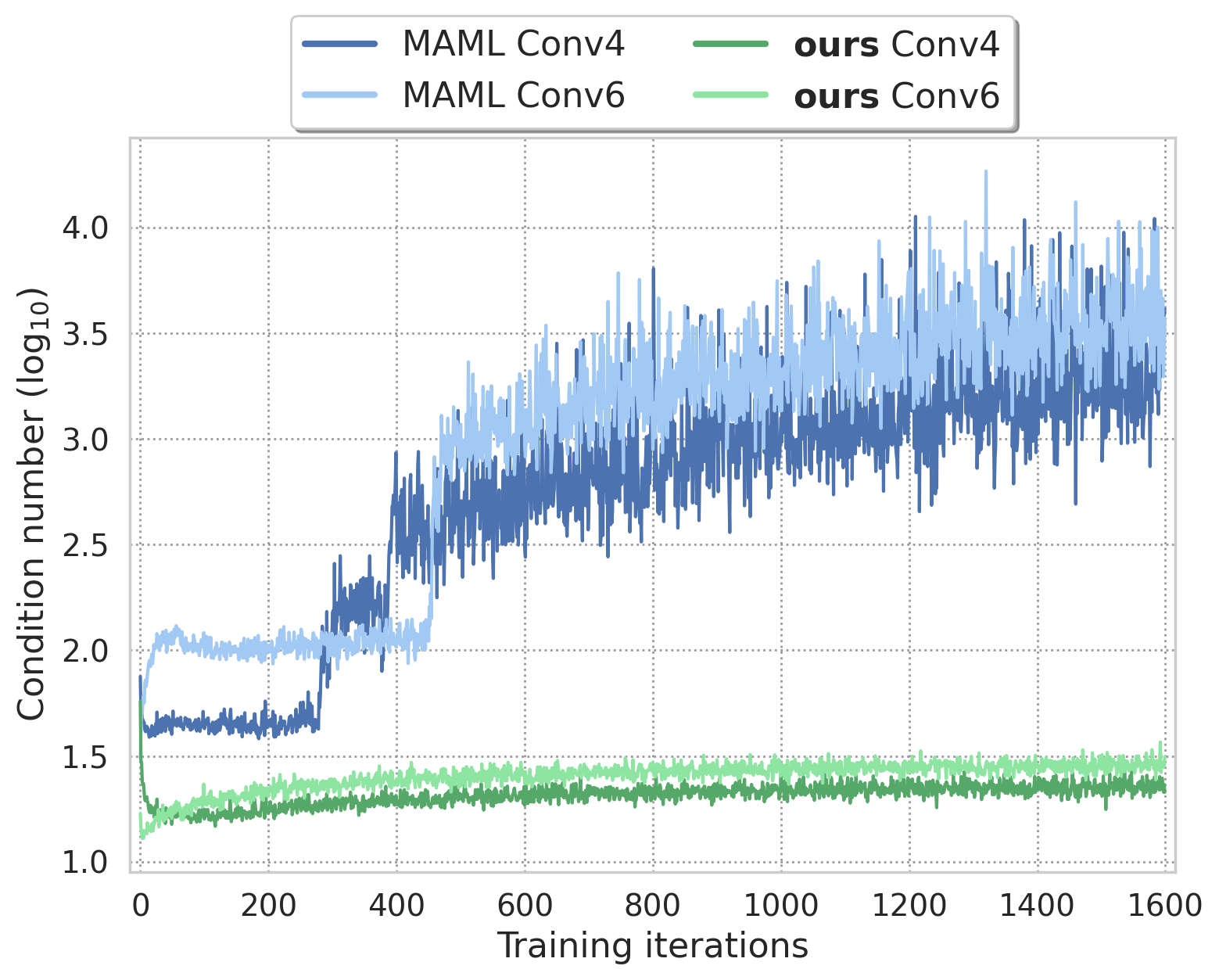}
    \caption{Step 3: $\kappa\left(\boldsymbol{\theta}^{(3)}_{\mathrm{train}}\right)$}
    \label{subfig:sup_cnmbr_train_s3}
    \vspace*{0.2in}
\end{subfigure}
\hfill
\begin{subfigure}[b]{0.48\textwidth}
    \includegraphics[width=\textwidth]{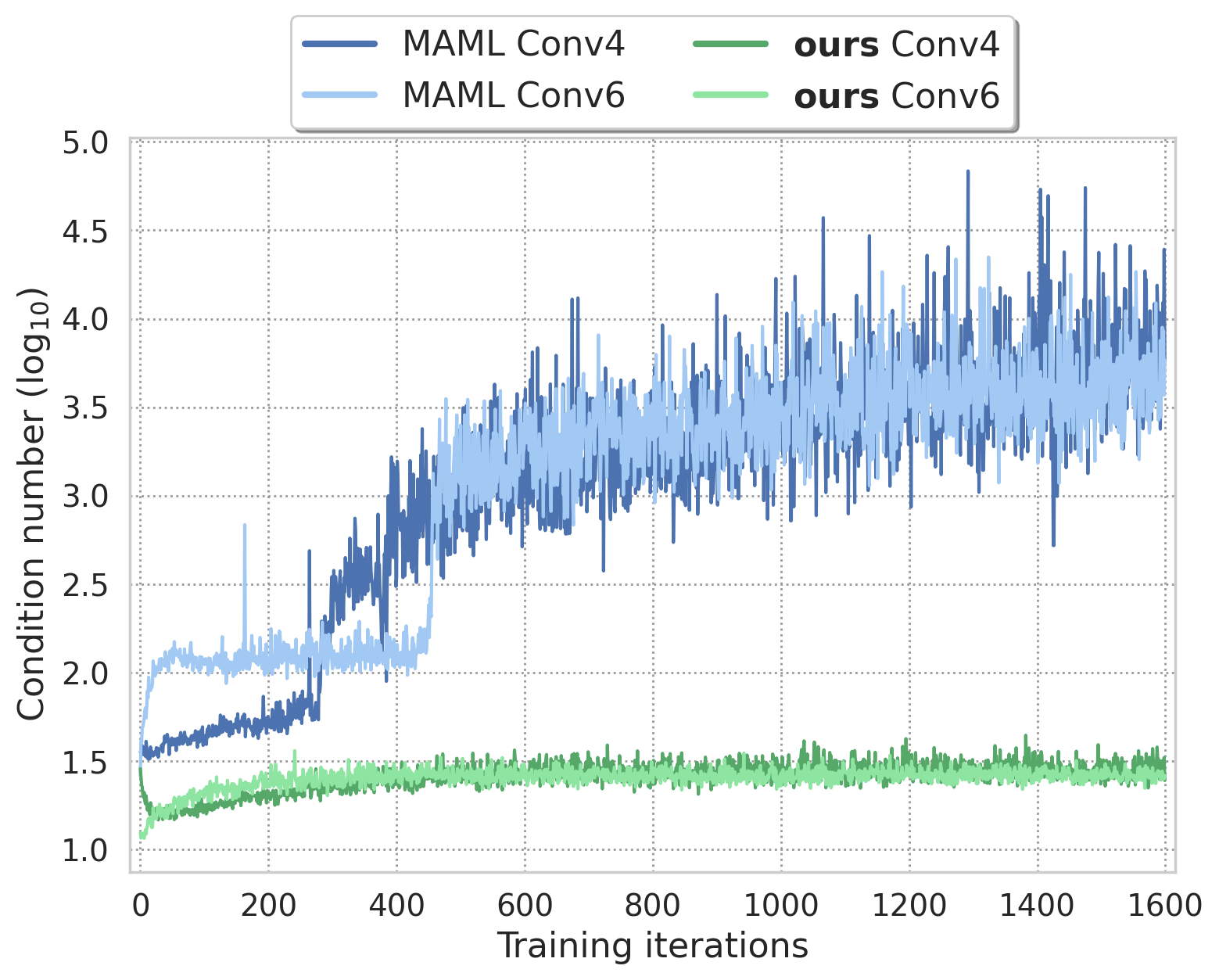}
    \caption{Step 3: $\kappa\left(\boldsymbol{\theta}^{(3)}_{\mathrm{valid}}\right)$}
    \label{subfig:sup_cnmbr_val_s3}
    \vspace*{0.2in}
\end{subfigure}
\hfill
\begin{subfigure}[b]{0.48\textwidth}
    \includegraphics[width=\textwidth]{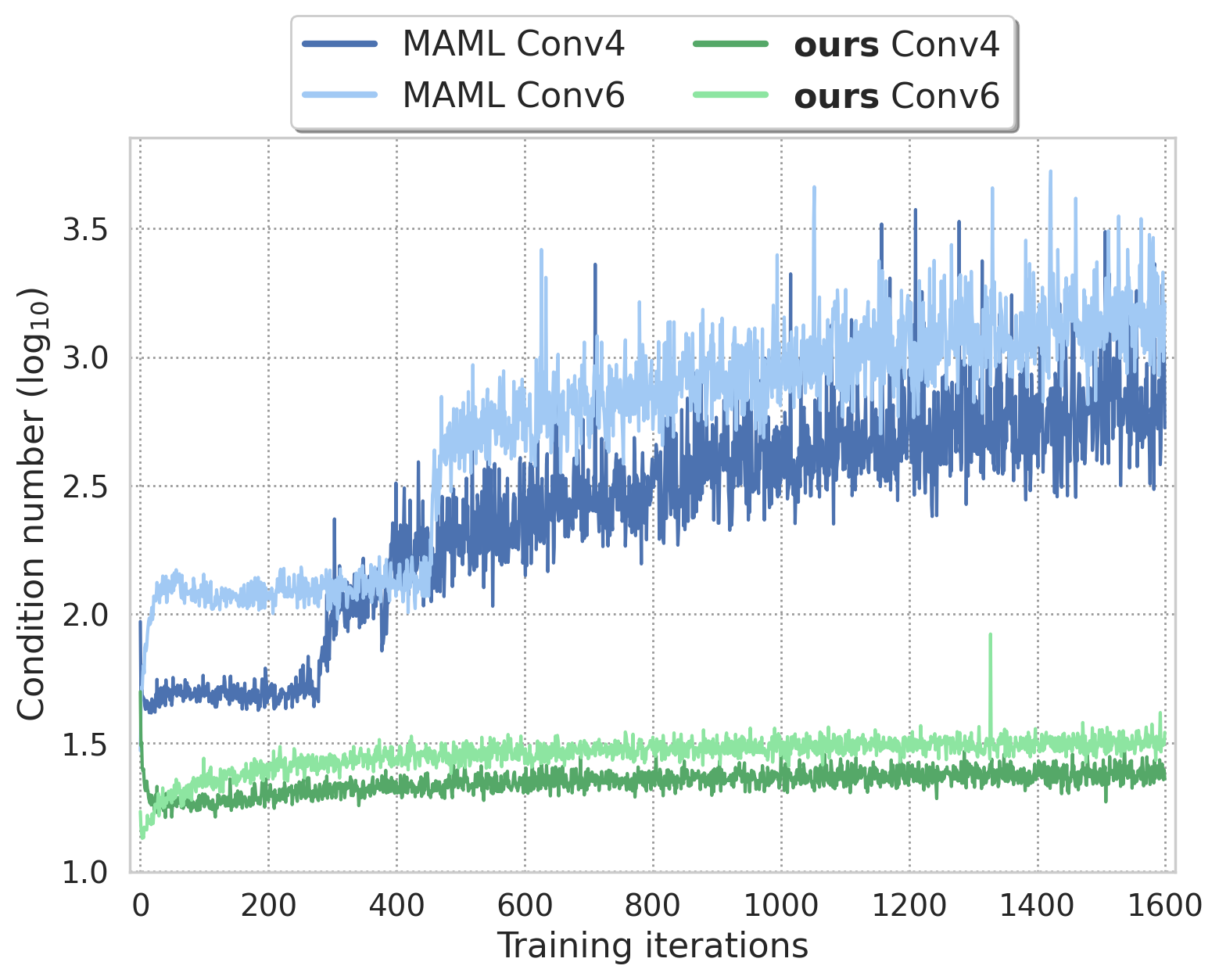}
    \caption{Step 4: $\kappa\left(\boldsymbol{\theta}^{(4)}_{\mathrm{train}}\right)$}
    \label{subfig:sup_cnmbr_train_s4}
    \vspace*{0.2in}
\end{subfigure}
\hfill
\begin{subfigure}[b]{0.48\textwidth}
    \includegraphics[width=\textwidth]{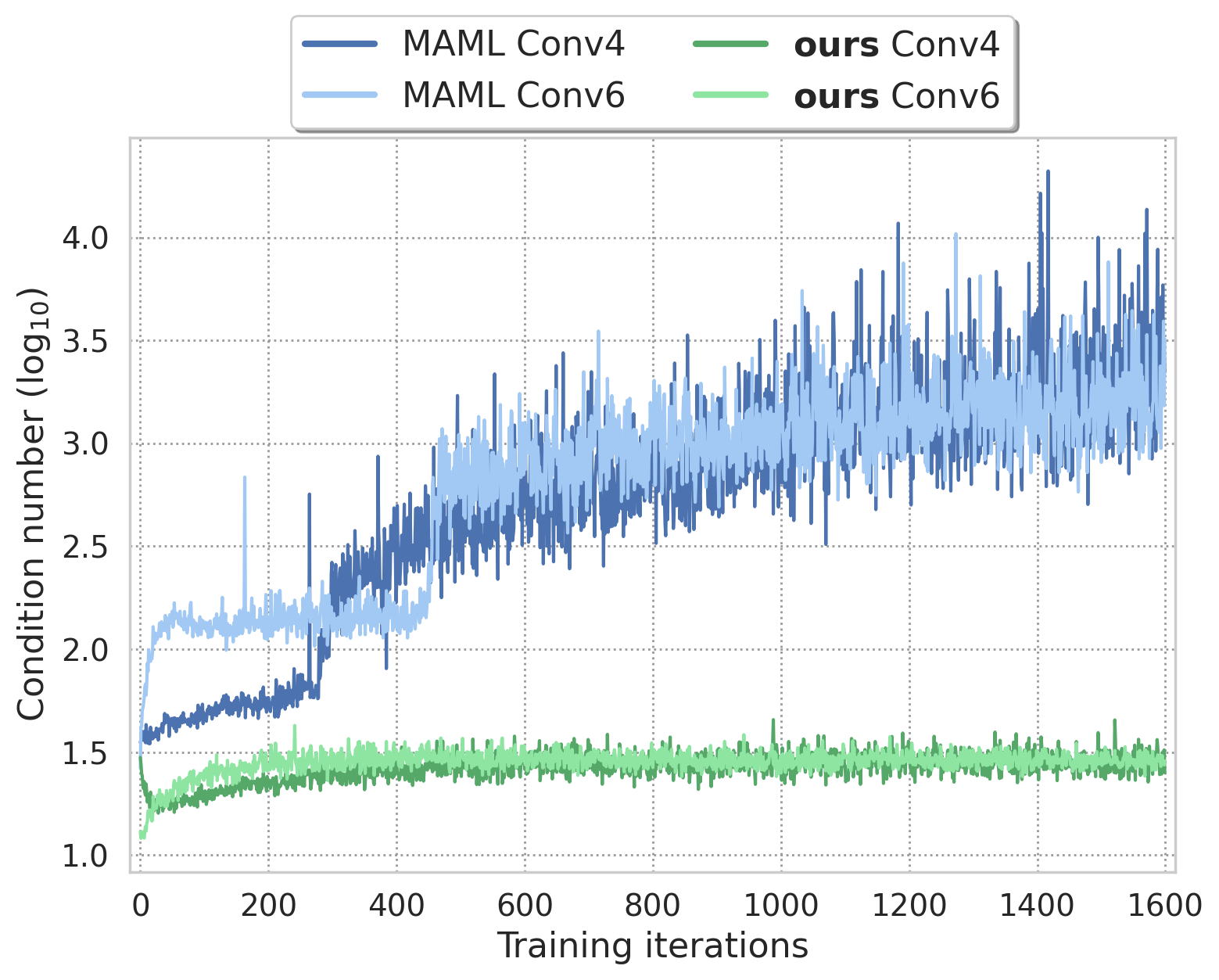}
    \caption{Step 4: $\kappa\left(\boldsymbol{\theta}^{(4)}_{\mathrm{valid}}\right)$}
    \label{subfig:sup_cnmbr_val_s4}
    \vspace*{0.2in}
\end{subfigure}
\caption{\textbf{Condition numbers over inner-loop update steps.} Reported results were obtained by training the baseline \textit{without} (`MAML') and \textit{with} our proposed conditioning constraint (`ours') with respect to the parameters of the model's classifier. Training has been conducted over 1600 iterations in a 5-way 5-shot scenario on the \textit{tiered}ImageNet dataset with a Conv4 and Conv6 architecture. For each update step, we report the condition number computed via either the support set of the training data $\kappa(\boldsymbol{\theta}^{(k)}_{\mathrm{train}})$ or validation data $\kappa(\boldsymbol{\theta}^{(k)}_{\mathrm{valid}})$.}
\label{fig:sup_valcond_cnmbr}
\end{figure}


\clearpage
{
\small
\bibliography{biblio_conditioning}
\bibliographystyle{icml2022}
}